\documentclass[conference]{IEEEtran}
\IEEEoverridecommandlockouts

\usepackage{amsmath,amssymb,amsfonts}
\usepackage{algorithmic}
\usepackage{graphicx}
\usepackage{textcomp}
\usepackage{xcolor}
\usepackage{makecell}
\usepackage{hyperref}
\usepackage[flushleft]{threeparttable}
\usepackage{multirow} 
\usepackage{tabularx} 
\usepackage[caption=false,font=normalsize,labelfont=sf,textfont=sf]{subfig}
\def\BibTeX{{\rm B\kern-.05em{\sc i\kern-.025em b}\kern-.08em
    T\kern-.1667em\lower.7ex\hbox{E}\kern-.125emX}}

\usepackage[
backend=biber,
style=ieee,
]{biblatex}
\DeclareMathOperator*{\argmax}{argmax}

\begin{document}

\title{An Empirical Bayes Analysis of Object Trajectory Representation Models\\
}

\author{Yue Yao, Daniel Goehring, Joerg Reichardt
\thanks{Yue Yao is with Continental AG and Ph.D. candidate at Freie Universität Berlin.}%
\thanks{Daniel Goering is Juniorprofessor for Autonomous Vehicles at the Dahlem Center for Machine Learning and Robotics of the Freie Universität Berlin}%
\thanks{Joerg Reichardt is with Continental AG.}%
}

\maketitle

\begin{abstract}
Linear trajectory models provide mathematical advantages to autonomous driving applications such as motion prediction.
However, linear models' expressive power and bias for real-world trajectories have not been thoroughly analyzed.
We present an in-depth empirical analysis of the trade-off between model complexity and fit error in modelling object trajectories. We analyze vehicle, cyclist, and pedestrian trajectories. Our methodology estimates observation noise and prior distributions over model parameters from several large-scale datasets. Incorporating these priors can then regularize prediction models. Our results show that linear models do represent real-world trajectories with high fidelity at very moderate model complexity. This suggests the feasibility of using linear trajectory models in future motion prediction systems with inherent mathematical advantages.

\end{abstract}

\section{Introduction}


The prediction of other traffic participants' future trajectories is an important input to the motion planning systems of autonomous agents. Recently, a number of large datasets of real-world traffic scenarios, e.g. Argoverse Motion Forecasting v1.1 (A1) \cite{chang_argoverse_2019}, Argoverse 2 Motion Forecasting (A2) \cite{wilson_argoverse2_2021} and Waymo Open Motion (WO) \cite{ettinger_waymo_2021}, have been made available and corresponding prediction challenges have sparked a large interest in designing trajectory prediction systems.

As one of the most widely used criteria of prediction accuracy in these challenges, the displacement error measures the difference between the predicted object positions and the \emph{noisy observations} of the object positions over the prediction horizon.

\begin{figure}[!h]
\centering
\includegraphics[height=1.3in]{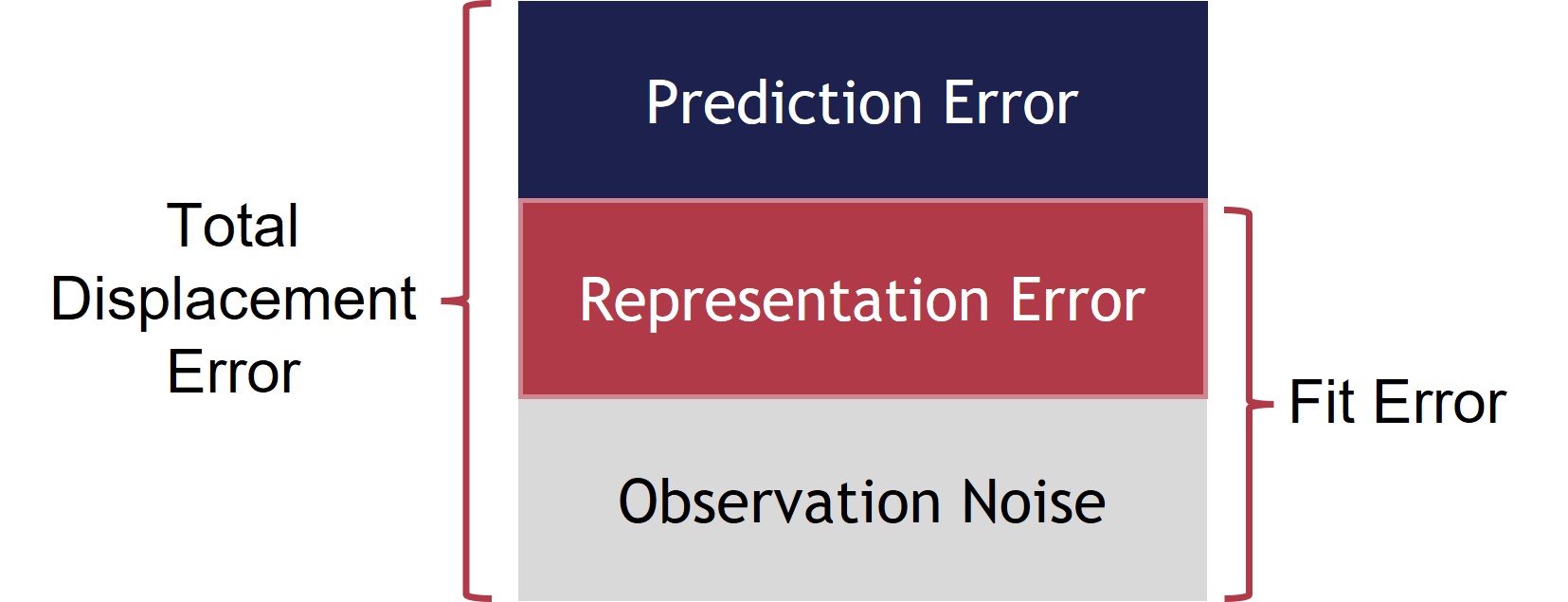}
\caption{The decomposition of the total displacement error from prediction system.}
\label{fig:error_decompostion}
\vspace{-1.5em}
\end{figure}

Figure \ref{fig:error_decompostion} shows how we can decompose the total displacement error produced by any prediction system into three independent components:
\begin{itemize}
\item \emph{Observation noise}: Deviation between ground truth and observations that is intrinsic to the data(set) and its collection process.
\item \emph{Representation error}: Deviation between ground truth and predictions due to a trajectory modelling choice.
\item \emph{Prediction error}: Deviation between ground truth and predictions due to the inadequacies of the prediction system and missing input information such as agents' intentions.
\end{itemize}
We denote \emph{fit error} as the sum of representation error and observation noise to describe how much a trajectory representation model deviates from observations.

Linear trajectory models, i.e. linear combination of basis fuctions, such as polynomials, are one of the model-based approaches for representing predicted trajectories in recent works \cite{buhet_plop_2020, su_temporally_2021}. Linear models bring many advantages (cf. Section \ref{section:linear models of trajectories}) into the prediction task and real applications \cite{reichardt_trajectories_2022}. As an example, instead of predicting future states sequentially, the prediction task can be simplified as predicting the polynomial parameters of trajectories. However, linear models have limited expressive power depending on model complexity and introduce representation error (bias) into the total displacement error.

Though linear trajectory models are used in recent works, the following questions have not been studied extensively:

\begin{itemize}
\item What is the optimal model complexity for representing trajectories with different timescales?
\item How much representation error is introduced by linear models of trajectories?
\end{itemize}
The public datasets provide noisy observations rather than ground truth, thus we will answer the second question by analyzing the fit error as the upper bound of representation error as visualized in Figure \ref{fig:error_decompostion}.

In this work, we study the general class of linear models with polynomial basis functions and investigate the trade-off between model complexity, i.e., flexibility, and model fit in a principled manner and on a large scale. Our focus is the general characterization of linear trajectory models \emph{independent} of any particular prediction method. 
Hence, we deliberately do not analyze the performance difference of any prediction models integrating linear trajectory models. This avoids conflating representation error and prediction error in our work. Our key contributions are as follows:
\begin{itemize}
\item An extensive empirical investigation of the fit error of polynomial trajectory representations for different classes of traffic participants in public datasets.
\item Estimation of prior distributions over observation noise and model parameters using the Empirical Bayes method. 
\item Estimation of the optimal model complexity from noisy trajectory data.
\end{itemize}
To the best of our knowledge, this work is the first that estimates noise levels in trajectory prediction datasets in a grounded manner.

Our discussion will proceed as follows:
We will first summarize some of the theoretical advantages of polynomial representations. We introduce the Empirical Bayes method to estimate prior distributions over model parameters and observation noise. We characterize the trade-off between model complexity, i.e.\ the number of basis functions used, and data-fit. For this, we employ information theoretic measures and average fit error between the trajectory representation and the data. In three large public datasets and for three different classes of traffic participants, we investigate this trade-off for trajectories of various lengths and determine the optimal model complexities. Compared to the estimated total displacement error reported for state-of-the art prediction systems, we show that the fit error is indeed small. From this, we conclude that model-based representations currently do not limit prediction performance and argue for the feasibility of their use in prediction systems.

\section{Related Work}


\begin{figure}[!h]
\centering
\includegraphics[width=3.4in]{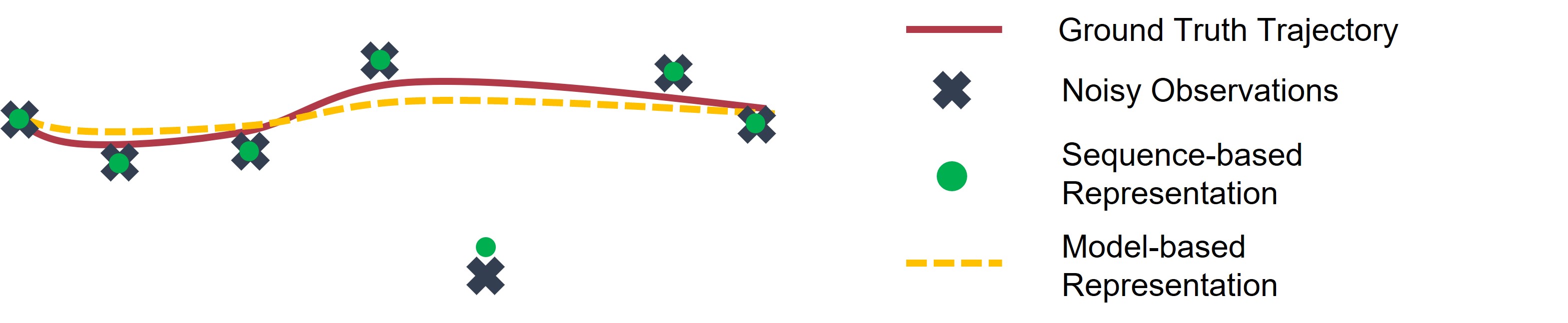}
\caption{An example of expression flexibility of sequence-based representations and model-based representations.}
\label{fig:sequence_vs_model}
\end{figure}

Many prediction systems employ sequence-based representation for the output, e.g., as a sequence of positions \cite{liang_learning_2020, phan_covernet_2020, nayakanti_wayformer_2022} or control states (acceleration and turn rate) \cite{salzmann_trajectron_2020, varadarajan_multipath++_2022} at fixed time-points over the prediction horizon.
Sequence-based approaches appear, at first glance, as very rational trajectory representations, having the same format as observations in datasets. Sequence-based approaches are examples of unbiased representations that have no representation error by design. They can express even random, erratic trajectories as in Figure \ref{fig:sequence_vs_model}. The downside of unbiasedness for any estimator is the high variance which typically requires more training data and the fact that sequence-based representations amalgamate the observation uncertainty and the actual motion of the objects being modelled. Additionally, the computational requirements of these approaches scale with the length and temporal resolution of the prediction horizon.

On the other hand, parametric, or model-based, trajectory representations are used less as output representations in recent works. Scholler et al.\ \cite{scholler_constant_2020} for example argue that predicting pedestrian trajectories as straight lines, i.e. polynomials of degree 1, yields competitive prediction performance. Vehicle trajectories are represented as polynomials of at most degree 4 in \cite{buhet_plop_2020, su_temporally_2021}. Su et al.\ \cite{ su_temporally_2021} highlight the temporal continuity, i.e. the ability to provide arbitrary temporal resolution, of this representation. Reichardt \cite{reichardt_trajectories_2022} argues for the use of polynomial representations to integrate trajectory tracking and prediction into a filtering problem. Model-based trajectory representations place restrictions on the kind of trajectories that can be predicted and introduce bias into prediction systems. This limited flexibility is generally associated with lower demands for training data and greater computational efficiency. The deliberate choice of a model, its parameters and associated prior distributions can directly encode knowledge about the system behavior that serves as a regularizer for prediction models.

\section{Linear Models of Trajectories}
\label{section:linear models of trajectories}

In contrast to paths, i.e. curves in space, trajectories are curves in space \emph{and} time. Over any finite time horizon $[t_0, t_0+T]$, the position of an object $\mathbf{c}(\tau)\in \mathbb{R}^d$ at (rescaled) time $\tau=\frac{t-t_0}{T}$  can be expressed as a linear combination of $n+1$ \emph{fixed} basis functions of time $\phi_k(\tau) : \mathbb{R}\rightarrow\mathbb{R}$ and parameters $\boldsymbol{\omega}_k\in \mathbb{R}^d$:
\begin{equation}
\label{eqn:linear_combination}
\mathbf{c}(\tau) = \sum_{k=0}^n\phi_k(\tau)\boldsymbol{w}_k.
\end{equation}

It allows us to incorporate prior knowledge about object trajectories via the basis functions $\phi_k(\tau)$. For example, since traffic participants carry mass and underlie physical constraints on how much force they can exert as well as having strong preferences for smooth motion \cite{macadam_understanding_2003, hayati_jerk_2020, bae_self_2020}, we can choose a small number $n+1$ of smooth basis functions and model deviations from smooth behavior as random noise.



We now discuss the benefits of linear models for representing trajectories. 

First, parameters $\boldsymbol{\omega}_k$ have spatial semantics. Taking the \emph{Bernstein} polynomial as an example, the curve $\mathbf{c}$ is the Bezier curve and the parameters $\boldsymbol{\omega}_k$ are control points. This allows the formulation of prior distributions based on spatial information, which is advantageous to regularize and inform predictions of traffic participants' future motion \cite{bahari_vehicle_2022}. Without loss of generality, we can constrain the range of $\phi_k(\tau)$ to the interval $[0, 1]$ and this automatically constrains $\mathbf{c}(\tau)$ to the convex hull of the $\boldsymbol{\omega}_k$.

Second, the transformation of $\mathbf{c}(\tau)$ is equivalent to applying the same transformation to the $\boldsymbol{\omega}_k$ and vice versa, which significantly simplifies applications with moving sensors.

Third, temporal derivatives only affect the basis functions 
$
\dot{\mathbf{c}}(\tau)=\sum_{k=0}^n\dot{\phi}_k(\tau)\boldsymbol{\omega}_k
$,
i.e.\ positions, velocities, accelerations and higher derivatives share the same parameterization!

Fourth, spatio-temporal distributions over the kinematic state of an object at any point in time can be readily derived from distributions over the parameters $\boldsymbol{\omega}_k$.

Fifth, we may estimate the parameters $\boldsymbol{\omega}_k$ via the solution of a linear system from either measurements of position at different time-points or measurements of different derivatives at the same time-point, or combinations thereof. For example, with $n=5$ we only need position, velocity and acceleration at two time-points $\tau$ and $\tau+\Delta \tau$ as $6$ constraints in order to fully determine all $6$ trajectory parameters $\boldsymbol{\omega}_k$.


Lastly, the linear combination of basis functions enables a common trajectory representation in tracking, filtering and prediction applications, e.g. parameters of the past trajectory can be tracked and parameters of the future trajectory will be predicted. It even allows the formulation of the trajectory prediction problem as a filtering problem \cite{reichardt_trajectories_2022}. In these applications, the trajectory parameters $\boldsymbol{\omega}_k$ replace the conventionally used kinematic state variables for tracking, filtering and prediction. The motion models in such applications can then be informed and regularized by prior distributions over $\boldsymbol{\omega}_k$.




\section{Estimation of Fit Error}
\begin{figure*}[!ht]
\centering
\hfill
\includegraphics[height=2.0in]{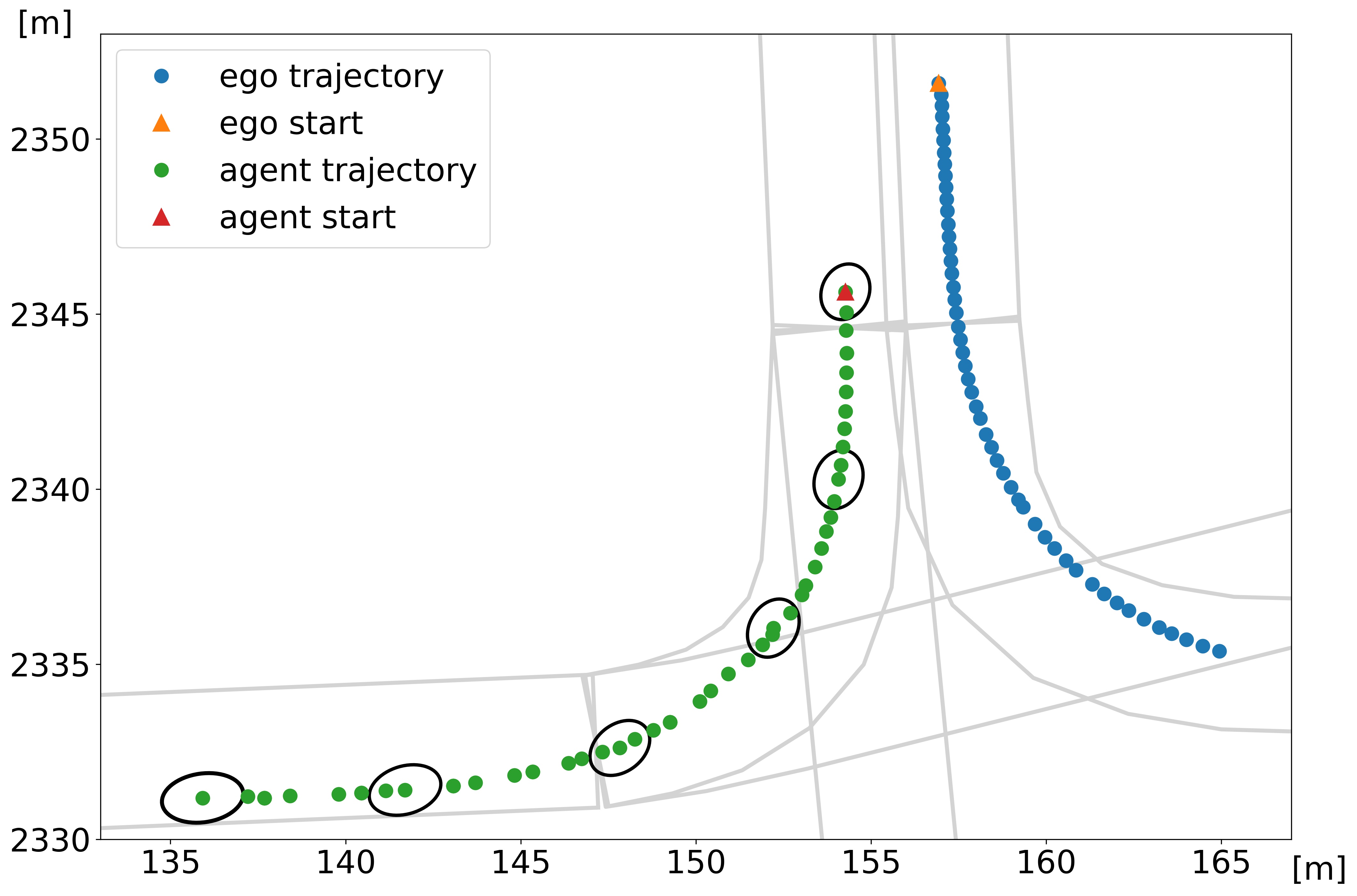}\label{fig:observation cov}
\hfill
\includegraphics[height=2.0in]{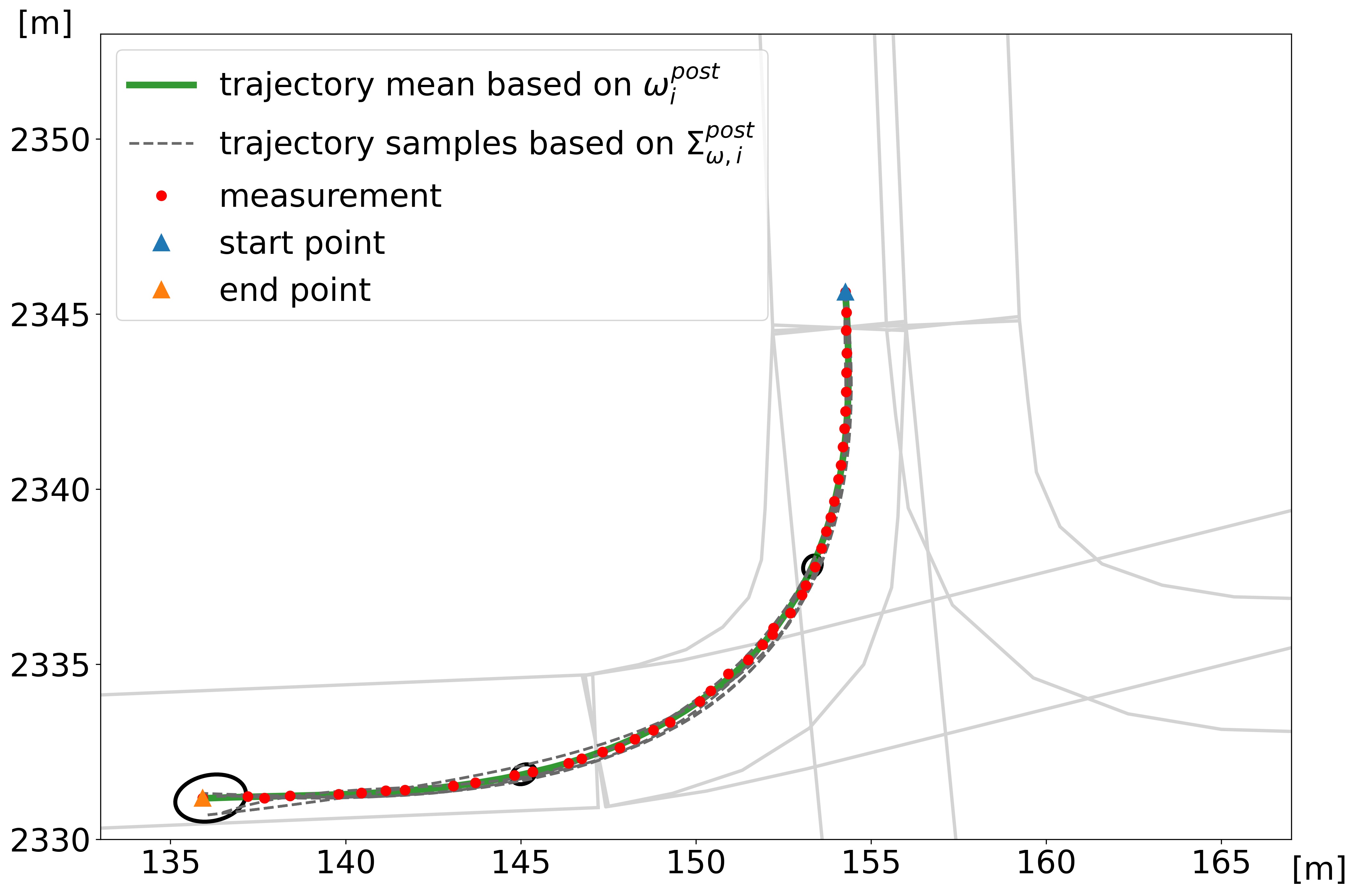}\label{fig:parameter cov}
\hfill
\hfill
\caption{\textbf{Left:} A typical scene from a trajectory prediction dataset, here Argoverse Motion Forecasting v1.1 \cite{chang_argoverse_2019}. Data is gathered by a moving sensor platform (ego vehicle) and subsequently transformed into a fixed world coordinate frame. As distance and angle between sensor and agent change during recording, the observation covariance of agent locations stretches and rotates over time. We show all sample points and a few $95\%$ confidence ellipses for agent position, enlarged by a factor of $4$ for better visibility. \textbf{Right:} The same agent trajectory as in the \textbf{left} figure but fitted with a $5$-degree polynomial trajectory representation estimated via eq.\ (\ref{eqn:posterior_mean}). The resulting posterior covariances for agent positions are also shown, enlarged by a factor of $8$ for better visibility.}
\label{agt_ob_cov}
\vspace{-1.0em}
\end{figure*}

We summarized the characteristics of 3 large-scale datasets in Table \ref{tab: dataset characteristics}. Datasets for training and evaluating trajectory prediction methods are obtained from a moving sensor platform (\emph{ego} vehicle) during measurement campaigns. Object detections are tracked and transformed into a \emph{world} coordinate system and reported in the dataset. Each dataset selects one or multiple objects of interest in one scenario and refers to them as \emph{agents}. Rather than ground truth, object positions and kinematics in the data represent noisy estimates. Hence, when fitting trajectory models to the data in order to measure the fit error, we need to regularize and take the observation noise into account explicitly, i.e.\ we need to perform a Bayesian regression. The kinematic variables provided in datasets vary, but all provide \emph{position measurements} for object centroids, which is what we focus on here. Figure \ref{agt_ob_cov} shows a typical example.

\begin{table}[!h]
\vspace{-0.0em}
\caption{Datasets characteristics}
\centering
\begin{tabularx}{3.2in}{c c >{\centering\arraybackslash}X >{\centering\arraybackslash}X >{\centering\arraybackslash}X}
\Xhline{3\arrayrulewidth}
\multicolumn{2}{c}{Dataset (training split)} & A1  \cite{chang_argoverse_2019} & A2 \cite{wilson_argoverse2_2021} & WO \cite{ettinger_waymo_2021}\\
\Xhline{3\arrayrulewidth}
\multicolumn{2}{l}{\#scenarios, \#ego trajectories} & 206K & 200K  & 487K \\
\hline
\multicolumn{2}{l}{\#agent (vehicle) trajectories}  & 206K & 176K & 1.84M \\
\hline
\multicolumn{2}{l}{\#agent (cyclist) trajectories}  & - & 3K & 63K \\
\hline
\multicolumn{2}{l}{\#agent (pedestrian) trajectories}  & - & 14K & 232K \\
\hline
\multicolumn{2}{l}{maximal time horizon [s]} & 5 & 11 & 9\\
\hline
\multicolumn{2}{l}{\#cities} & 2 & 6 & 6\\
\hline
\multicolumn{2}{l}{sampling rate} & \multicolumn{3}{c}{10 Hz}\\
\hline
\multirow{4}{*}{\thead{trajectory \\ information}} & position & 2D & 2D & 3D\\
\cline{2-5}
&velocity & - & 2D & 2D\\
\cline{2-5}
&orientation & - & \checkmark & \checkmark\\
\cline{2-5}
&timestamp & \checkmark & \checkmark & \checkmark\\
\Xhline{3\arrayrulewidth}
\label{tab: dataset characteristics}
\vspace{-1em}
\end{tabularx}
\end{table}

In order to formulate the regression, we will introduce some definitions for notational convenience. We form a parameter vector $\boldsymbol{\omega}\in \mathbb{R}^{(n+1)d}$ as $\boldsymbol{\omega}=[\boldsymbol{\omega}_0,\cdots,\boldsymbol{\omega}_n]^{\top}$. Then we can express a Gaussian prior over parameters $p(\boldsymbol{\omega}) = \mathcal{N}(\boldsymbol{\omega}|\boldmath{0}, \boldsymbol{\Sigma}_{\boldsymbol{\omega}})$. The prior has zero mean due to symmetry and a full covariance $\boldsymbol{\Sigma}_{\boldsymbol{\omega}}$ that allows expressing correlations between spatial dimensions. We form a vector of basis functions $\boldsymbol{\phi}(\tau)\in \mathbb{R}^{n+1}$ by concatenation $\boldsymbol{\phi}(\tau)=[\phi_0(\tau),\cdots,\phi_n(\tau)]^{\top}$. In this notation, equation (\ref{eqn:linear_combination}) is then written as $\mathbf{c}(\tau)=(\boldsymbol{\phi}^{\top}(\tau)\otimes\mathbf{I}_d)\boldsymbol{\omega}$, where $\mathbf{I}_d$ is a $d\times d$ identity matrix and the Kronecker product $\otimes$ distributes the basis functions over the $d$ spatial dimensions. Data for a single object $i$ consists of $m$ sample points at discrete time-points $\tau_{i,1}, ...,\tau_{i,m}$. We can arrange all modelled trajectory points of the object $i$ into a vector $\mathbf{c}_i\in\mathbb{R}^{md}$ as  $\mathbf{c}_i=[\mathbf{c}_i(\tau_{i,1}),\cdots,\mathbf{c}_i(\tau_{i,m})]^{\top}$ noting that the sample times, in general, are different for different objects. By introducing the $(n+1)d\times md$ matrix $\boldsymbol{\Phi}_i=[\boldsymbol{\phi}(\tau_{i,1})\otimes\mathbf{I}_d, \cdots,\boldsymbol{\phi}(\tau_{i,m})\otimes\mathbf{I}_d]$ we can write most compactly $\mathbf{c}_i=\boldsymbol{\Phi}_i^{\top}\boldsymbol{\omega}$.

We denote the observed trajectory as $\mathbf{c}^\mathrm{ob}_i$ for the $i^{\mathrm{th}}$ object and $\mathbf{c}^\mathrm{ob}_{i, j}$ for $j^{\mathrm{th}}$ sample point to distinguish them from our model notation. We assume additive zero mean Gaussian observation noise with $d\times d$ covariance matrix $\boldsymbol{\Sigma}_{o,i,j}$ depending on the object index $i$ and the sample point $j$:
\begin{equation}
\begin{aligned}
\mathbf{c}^{\mathrm{ob}}_{i,j} & = \mathbf{c}_{i,j} + \eta, \:\:
\eta \sim \mathcal{N}(0, \boldsymbol{\Sigma}_{o, i, j}) 
\end{aligned}
\end{equation}

Assuming statistically independent noise along a single trajectory, we form the observation noise covariance for $\mathbf{c}_i$ as an $md\times md$ block diagonal matrix $\boldsymbol{\Sigma}_{o, i}$ where the $j^\mathrm{th}$ diagonal block is given by $\boldsymbol{\Sigma}_{o,i,j}$. The left part of Figure \ref{agt_ob_cov} illustrates the time and trajectory dependence of the observation noise. Note how it rotates and scales due to the angular and distance dependence of the observation noise to the sensors on the ego vehicle.

Using this notation, the posterior estimate of model parameters for a single trajectory $\mathbf{c}_i$ is given in closed form \cite[232--234]{murphy_machine_2012}:
\vspace{-0.5em}
\begin{equation}
\label{eqn:posterior_mean}
\begin{aligned}
\boldsymbol{\Sigma}^{\mathrm{post}}_{\omega, i} = & (\boldsymbol{\Sigma}_\omega^{-1} + \boldsymbol{\Phi}_i  \boldsymbol{\Sigma}_{o, i}^{-1} \boldsymbol{\Phi}_i^{\top})^{-1}\\
\boldsymbol{\omega}^{\mathrm{post}}_i = & \boldsymbol{\Sigma}^\mathrm{post}_{\omega, i}\boldsymbol{\Phi}_i \boldsymbol{\Sigma}_{o,i}^{-1} \mathbf{c}^\mathrm{ob}_i
\end{aligned}
\end{equation}

With this improved parameter estimate from (\ref{eqn:posterior_mean}), we can then compute the \emph{average fit error} (AFE) along each trajectory in an entire dataset:
\begin{equation}
\mathrm{AFE} = \frac{1}{Nm}\sum_i^N\sum_j^m ||(\boldsymbol{\phi}^{\top}(\tau_{i,j}) \otimes I_d)\boldsymbol{\omega}_i^{\mathrm{post}} - \mathbf{c}^\mathrm{ob}_{i,j}||_2
\end{equation}

For vehicles, the projection of the AFE onto the longitudinal and lateral direction of motion are of particular interest for driving applications. We denote these projections as AFE$_{\mathrm{lon}}$ and AFE$_{\mathrm{lat}}$. The object heading is either provided directly in the data (A2 and WO) or is inferred from a Rauch-Tung-Striebel (RTS) smoothing of the data (A1, cf.\ Section \ref{section:preprocessing}).

The above estimation of fit error requires the specification of observation covariance $\boldsymbol{\Sigma}_{o, i}$, prior covariance $\boldsymbol{\Sigma}_\omega$ and model complexity $n$ - neither of which is given. We now employ the Empirical Bayes Method to estimate all three quantities.

\section{Empirical Bayes Method}  
\label{section:method}%
The Empirical Bayes approach \cite{efron_large_2012} allows us to bootstrap prior distributions over model parameters if many independent samples of the same phenomenon are observed, such as the object trajectories in our datasets. The idea is to formulate the likelihood of all observed trajectories $\mathbf{C}$ as a function of the prior parameters alone. This can be achieved by marginalizing the actual model parameters. Optimal prior parameters maximize the resulting, so-called, type-\RN{2} likelihood.

In an ideal world, the only prior parameter to be estimated would be the covariance matrix $\boldsymbol{\Sigma}_{\boldsymbol{\omega}}$. The observation noise covariance matrices $\boldsymbol{\Sigma}_{o,i,j}$ would be derived from the ego vehicle's sensor setup and given with the dataset. Unfortunately, the $\boldsymbol{\Sigma}_{o,i,j}$ are not provided in any dataset and so we have to reverse engineer, i.e.\ estimate, them from the data.

Clearly, we cannot estimate individual $\boldsymbol{\Sigma}_{o,i,j}$ for every trajectory and every time-point. Instead, we provide a structured parameterization in the form $\boldsymbol{\Sigma}_{o,i,j}=\boldsymbol{\Sigma}_o(\boldsymbol{\theta})$.

Since ego trajectories and agent trajectories result from different sensor setups, we also differentiate the parameterization of their noise models.

\subsection{Observation Noise Covariance for Ego Trajectories}
We model the observation noise covariance for ego trajectories in \emph{world} coordinates, assuming constant observation noise in $x$ and $y$ direction for all sample points. The covariance for one observation point is expressed as:
\begin{equation}
\label{eqn_ego_cov}
\begin{aligned}
\boldsymbol{\Sigma}_{o,i,j}^{\mathrm{ego, world}} & = \boldsymbol{\Sigma}_{o}^{\mathrm{ego, world}} = 
\begin{bmatrix}
\sigma^2_{\mathrm{x}} & \sigma_{\mathrm{xy}} \\
\sigma_{\mathrm{xy}} & \sigma^2_{\mathrm{y}} \\
\end{bmatrix} 
\end{aligned}
\end{equation}
with $\sigma^2_{\mathrm{x}} = \sigma^2_{\mathrm{y}} = \sigma^2_{\mathrm{diag}}$ and $\sigma_{\mathrm{xy}} = \sigma_{\mathrm{cov}}$. We can extend it to the covariance of one complete ego trajectory via $\boldsymbol{\Sigma}_{o,i}(\boldsymbol{\theta}_{\mathrm{ego}}) = \boldsymbol{I}_m \otimes \boldsymbol{\Sigma}_o^{\mathrm{ego, world}}$ and thus for all ego trajectories, we only have two parameters to estimate  $\boldsymbol{\theta}_{\mathrm{ego}}=[\sigma_{\mathrm{diag}}, \sigma_{\mathrm{cov}}]$. 

\subsection{Observation Noise Covariance for Agent Trajectories}
Since the ego vehicles used in the datasets feature a suite of LIDAR sensors, the observation uncertainty for agents is best expressed in \emph{polar} coordinates. We assume a constant angular resolution ($\sigma^2_{\alpha}$) and variable distance resolution ($\sigma^2_{r, i,j}$):
\begin{equation}
\label{eqn_cov_polar}
\begin{aligned}
\boldsymbol{\Sigma}_{o, i,j}^{\mathrm{agent, polar}} &= 
\begin{bmatrix}
\sigma^2_{r, i,j} & 0 \\
0 & \sigma^2_{\alpha} \\
\end{bmatrix}
\end{aligned}
\end{equation}
We model $\sigma_{r,i,j}^2$ as an increasing function of the measured distance $r_{i,j}$ between agent and ego at the $j^{\mathrm{th}}$ sample point ($j \in [1, 2, \hdots, m]$) with parameters $[\beta_0, \beta_1, \beta_2] \in \mathbb{R}^+$:
\begin{equation}
\label{eqn_cov_r}
\sigma^2_{r,i,j} = \beta_{0}+ \beta_{1} r_{i,j} + \beta_{2}r_{i,j}^2
\end{equation}
Clearly, this is not perfect since we are using observations to parameterize the uncertainty of these very observations. But it is the best approach we have and it is safe to assume $\sigma^2_{r,i,j}$ varies only slightly within the uncertainty of $r_{i,j}$.
Next, we transform the observation covariance from the $polar$ frame to the Cartesian \emph{ego} frame $\boldsymbol{\Sigma}_{o, i,j}^{\mathrm{agent, ego}}$  based on \cite[77]{kampchen_feature_2007}.

Finally, we rotate $\boldsymbol{\Sigma}_{o, i,j}^{\mathrm{agent, ego}}$ from the \emph{ego} frame to the \emph{world} frame with:
\begin{equation}
\label{eqn_cov_ego}
\begin{aligned}
\boldsymbol{\Sigma}_{o, i,j}^{\mathrm{agent, world}} = \boldsymbol{R}_{i,j}(\boldsymbol{\Sigma}_{o, i,j}^{\mathrm{agent, ego}} +  \sigma_{c}^2 \boldsymbol{I}_d)\boldsymbol{R}^{\top}_{i,j}
\end{aligned}
\end{equation}
where $\boldsymbol{R}_{i,j}$ denotes the rotation matrix at $j^{\mathrm{th}}$ timestamp in the $i^{\mathrm{th}}$ agent trajectory. The additional diagonal term $\sigma_{c}^2$ is necessary to model errors resulting from timing instability of the tracking system especially at close ranges.

The observation covariance $\boldsymbol{\Sigma}_{o,i}(\boldsymbol{\theta}_{\mathrm{agent}})$ for one complete agent trajectory is then an $md\times md$ block diagonal matrix where the $j^{\mathrm{th}}$ block is given by $\boldsymbol{\Sigma}_{o, i,j}^{\mathrm{agent, world}}$. Hence, we have only 5 parameters to estimate: $\boldsymbol{\theta}_{\mathrm{agent}}=[\sigma_\alpha, \beta_0, \beta_1, \beta_2, \sigma_c]$. We illustrate one example with our observation covariance model in Figure \ref{agt_ob_cov} with covariance marked as black ellipses.

\subsection{Estimating Prior Parameters via Empirical Bayes}

Finally, we are in the position to formulate the type-\RN{2} likelihood. Following \cite[172--176]{murphy_machine_2012}, we obtain:
\begin{equation}
\begin{aligned}
p(\mathbf{C} | \boldsymbol{\Sigma}_o(\boldsymbol{\theta}), \boldsymbol{\Sigma}_\omega) = & \prod_{i=1}^N\int \mathcal{N}(\mathbf{c}^\mathrm{ob}_i|\boldsymbol{\Phi}^{\top}_{i} \boldsymbol{\omega}, \boldsymbol{\Sigma}_{o,i}(\boldsymbol{\theta})) \\ & \times \mathcal{N}(\boldsymbol{\omega}|\boldsymbol{0}, \boldsymbol{\Sigma}_\omega)\mathrm{d}\boldsymbol{\omega} \\
= &\prod_{i=1}^N \mathcal{N}(\mathbf{c}^\mathrm{ob}_i|\boldsymbol{0}, \boldsymbol{\Sigma}_{o,i}(\boldsymbol{\theta}) + \boldsymbol{\Phi}^{\top}_{i} \boldsymbol{\Sigma}_\omega \boldsymbol{\Phi}_{i})\\ 
\end{aligned}
\label{eqn:type 2 likelihood}
\end{equation}
where $\mathbf{C}$ denotes all $N$ trajectories in the dataset. We maximize the log of (\ref{eqn:type 2 likelihood}) with respect to $\boldsymbol{\Sigma}_\omega$ and $\boldsymbol{\theta}$ for ego and agent trajectories separately using gradient descent method \cite{dillon_tensorflow_2017}. These optima represent the prior parameters estimated from the dataset:
\begin{equation}
\label{eqn_prior_optimization}
\hat{\boldsymbol{\theta}}, \hat{\boldsymbol{\Sigma}}_\omega =  \argmax_{\boldsymbol{\theta}, \boldsymbol{\Sigma}_\omega} \log(p(\mathbf{C} | \boldsymbol{\Sigma}_o(\boldsymbol{\theta}), \boldsymbol{\Sigma}_\omega))
\end{equation}

For any model complexity $n$, we can thus estimate Bayesian optimal trajectory representations from given data by plugging the estimated prior parameters $\hat{\boldsymbol{\Sigma}}_\omega$ and ${\boldsymbol{\Sigma}}_{o, i}(\hat{\boldsymbol{\theta}})$ into (\ref{eqn:posterior_mean}).

With increasing $n$ we will be able to achieve smaller AFE and estimate lower observation noise at the expense of an increasing number of parameters, i.e. we will start to overfit. We next find the optimal trade-off between data fit and model complexity. 

\subsection{Estimating Optimal Model Complexity}
The Akaike Information Criterion (AIC) \cite{akaike_AIC_1973} and Bayesian Information Criterion (BIC) \cite{schwarz_BIC_1978} characterize the score for a model in terms of how well it fits the data, minus how complex the model is to define. AIC and BIC are defined as:
\begin{equation}
\begin{aligned}
\label{eqn:AIC}
\mathrm{AIC} &= \frac{\log(p(\mathbf{C}| \boldsymbol{\Sigma}_o(\boldsymbol{\theta}),\boldsymbol{\Sigma}_\omega))}{N}  - \mathrm{dof}(\boldsymbol{\theta}, \boldsymbol{\Sigma}_\omega)\\
\mathrm{BIC} & = \frac{\log(p(\mathbf{C} | \boldsymbol{\Sigma}_o(\boldsymbol{\theta}), \boldsymbol{\Sigma}_\omega))}{N} - \frac{\mathrm{dof}(\boldsymbol{\theta}, \boldsymbol{\Sigma}_\omega)}{2} \log(m)
\end{aligned}
\end{equation}
where $\mathrm{dof}(\boldsymbol{\theta}, \boldsymbol{\Sigma}_\omega)=\mathrm{dof}(\boldsymbol{\theta})+d(n+1)(d(n+1)+1)/2$ denotes the degrees of freedom in the observation covariance and model parameter covariance. For ego and agent trajectories, $\mathrm{dof}(\boldsymbol{\theta}_{\mathrm{ego}})=2$ and $\mathrm{dof}(\boldsymbol{\theta}_{\mathrm{agent}})=5$, respectively. A maximum of either criterion as a function of $n$ indicates optimal trade-off between data-fit and model complexity. In general, BIC penalizes model complexity higher and tends to pick a simpler model.


\section{Experiments} 


\subsection{Preprocessing: Data Selection and Outlier Detection}
\label{section:preprocessing}
For comparability, we only analyze the training set and limit our analysis to vehicle trajectories in A1 and additionally consider cyclist and pedestrian trajectories in A2 and WO. We analyze trajectories for time horizons $T\in\{3\mathrm{s}, 5\mathrm{s}, 8\mathrm{s}\}$. For $T$ smaller than the maximal observation horizon in the dataset, we select time windows of size $T$ in strides of $1$s for A1 and randomly, one from each trajectory, for A2 and WO. From the much larger set of vehicle trajectories in WO, we limit our analysis to a random sample of 300k (without outlier) from 1.84 million vehicle trajectories to reduce computational cost.

Timestamps do not exactly follow the sampling rate. We consider all samples that are within sensor range of the ego vehicle for $T-0.5$s to $T+0.5$s in the analysis of time horizon $T$, but discard all static trajectories with lengths $\leq 0.5$m as these can be trivially represented with small error. 

We notice a number of outliers in the data due to tracking loss and inconsistent timing, i.e.\ objects are reported at physically impossible locations for the given timestamps. To automatically detect and discard such trajectories, we employ a Rauch-Tung-Striebel (RTS) smoothing of the data with a double integrator based on \cite[48]{philipp_perception_2021} and adjust the parameters according to \cite[59]{philipp_perception_2021}. We discard trajectories for which the RTS-Smoother estimates positions more than $2$m away from observation or the estimated longitudinal acceleration (deceleration) exceeds a $6\frac{\mathrm{m}}{\mathrm{s}^2}$ ($-10\frac{\mathrm{m}}{\mathrm{s}^2}$) threshold for vehicles \cite{bokare_acceleration_2017}, $2\frac{\mathrm{m}}{\mathrm{s}^2}$ ($-4\frac{\mathrm{m}}{\mathrm{s}^2}$) threshold for cyclists \cite{famiglietti_bicycle_2020}, and $2\frac{\mathrm{m}}{\mathrm{s}^2}$ ($-3\frac{\mathrm{m}}{\mathrm{s}^2}$) threshold for pedestrians. Table \ref{tab: outlier} gives an overview of the percentage of outliers in datasets discarded for trajectories of time horizon $T=5$s.

We notice the timing issue in A1, where $T$ varies from $4.81$s to $25.64$s for 50 sample points with $10$Hz \cite{argo1_time_issue}. 
The RTS-Smoother detects ego outliers in A2 due to the unstable velocity estimation at the trajectory's start or end. 
In WO, we find quantities of outliers detected by RTS-Smoother primarily because the agents leave the sensor range and their positions reset to $(0, 0)$. WO flags these trajectory points as non-valid.

\begin{table}[h]
\caption{Percentage of Outliers for $T=5$s Trajectory}
\centering
\begin{tabularx}{3.2in}{c c !{\vrule width 1.5pt} >{\centering\arraybackslash}X |  >{\centering\arraybackslash}X | >{\centering\arraybackslash}X}
\Xhline{3\arrayrulewidth}
\multicolumn{2}{c}{Datasets}&\multicolumn{1}{c}{A1} & \multicolumn{1}{c}{A2} & \multicolumn{1}{c}{WO}\\
\Xhline{3\arrayrulewidth}
\multirow{5}{*}{ego} & time &22.81& 0 & 0.02\\
\cline{2-5}
&static &23.95&20.66& 25.41\\
\cline{2-5}
&out of view &0&0& 0\\
\cline{2-5}
&RTS &0&1.18& 0\\
\cline{2-5}
&total &42.95&21.84& 25.42\\
\hline
\hline
\multirow{5}{*}{\thead{agent \\ vehicle}} & time &22.81& 0 & 0.02\\
\cline{2-5}
&static &0&4.95& 1.70\\
\cline{2-5}
&out of view &0&0& 19.40\\
\cline{2-5}
&RTS &6.81&0.86& 19.45\\
\cline{2-5}
&total &28.11&5.81& 20.83\\
\hline
\hline
\multirow{5}{*}{\thead{agent \\ cyclist}} & time &-& 0 & 0.02\\
\cline{2-5}
&static &-&1.04& 1.43\\
\cline{2-5}
&out of view &-&0& 24.12\\
\cline{2-5}
&RTS &-&3.91& 24.58\\
\cline{2-5}
&total &-&4.94& 25.62\\
\hline
\hline
\multirow{5}{*}{\thead{agent \\ pedestrian}} & time &-& 0 & 0.02\\
\cline{2-5}
&static &-&0.19& 1.32\\
\cline{2-5}
&out of view &-&0& 32.20\\
\cline{2-5}
&RTS &-&0.76& 32.25\\
\cline{2-5}
&total &-&0.93& 33.12\\
\Xhline{3\arrayrulewidth}
\label{tab: outlier}
\vspace{-2em}
\end{tabularx}
\end{table}


\subsection{Results}
\subsubsection{Estimation of Observation Noise}
\label{section:Noise Estimation}

Table \ref{tab: observation noise} reports results for $T=5$s at the value of $n=\hat{n}$ that maximizes AIC. As expected, ego trajectories exhibit much lower observation noise than vehicle trajectories in all datasets. The agent trajectories in A1 are significantly noisier than in A2 and WO. WO provides data with the least estimated observation noise for vehicle trajectories due to its offline tracking algorithm \cite{ettinger_waymo_2021}. 


\subsubsection{Model Complexity and Fit Error}
\begin{figure*}[!t]
\centering
\hfill
\includegraphics[width=2.3in, height=3.1in]{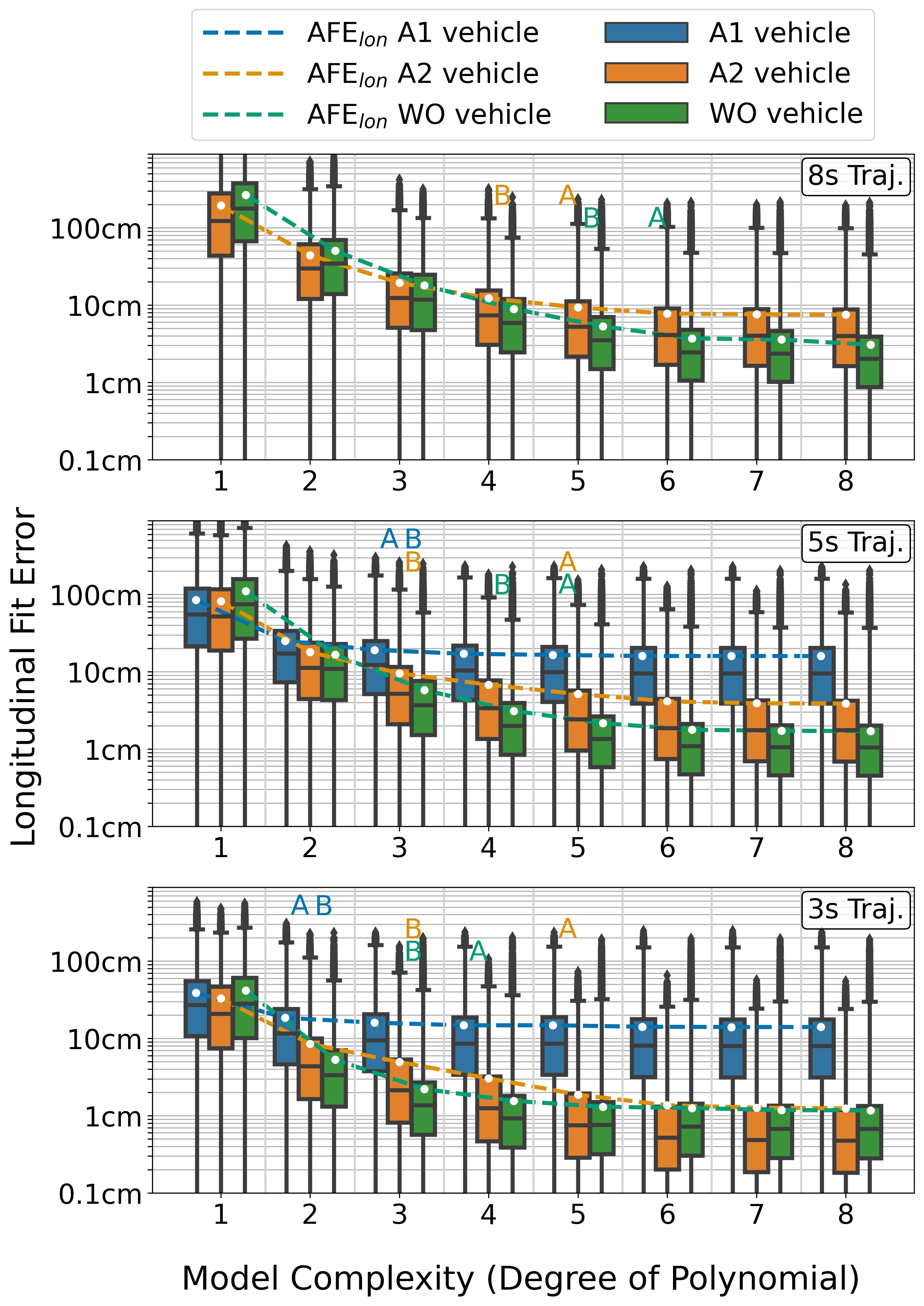} 
\hfill
\includegraphics[width=2.3in, height=3.1in]{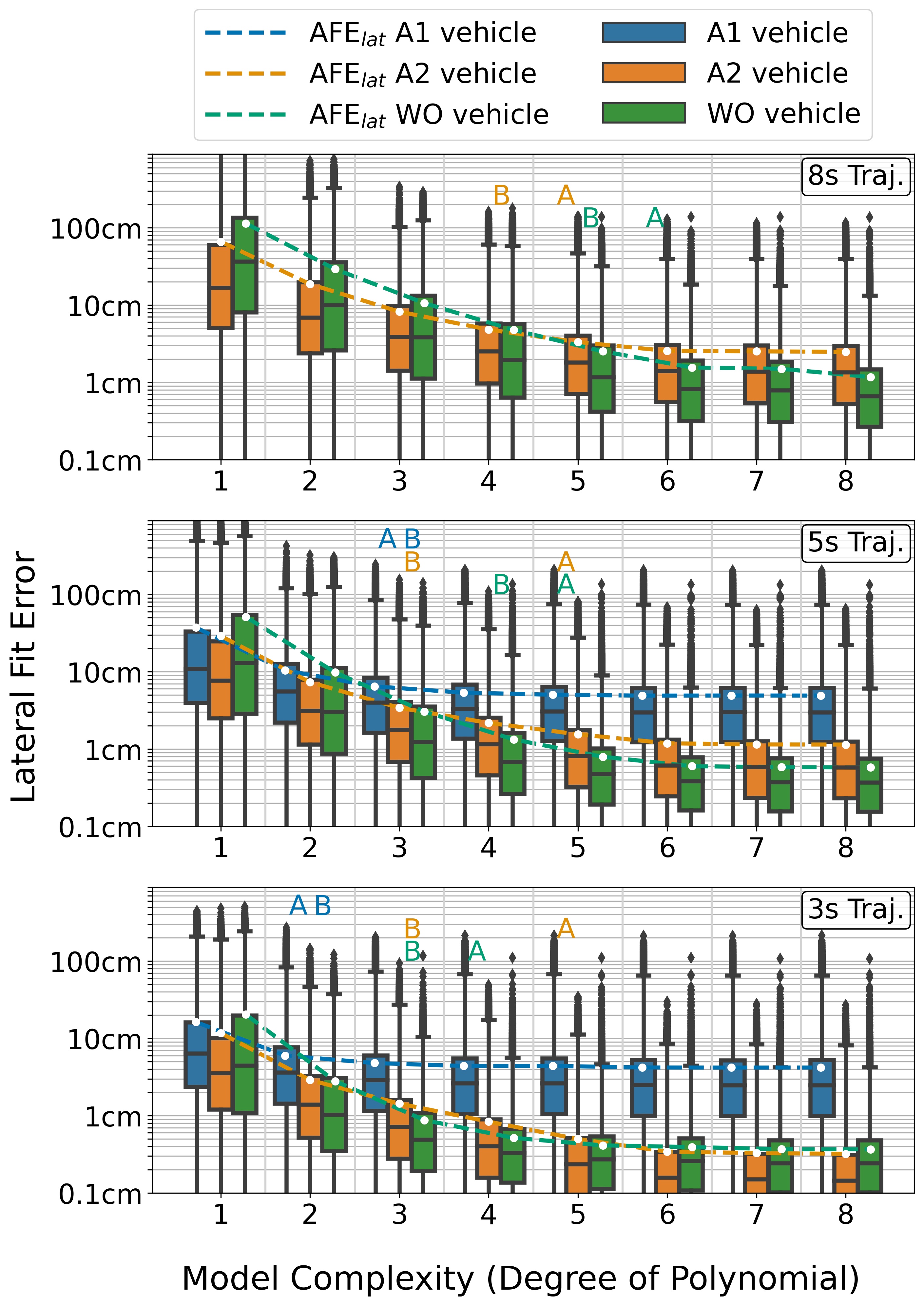} 
\hfill
\hfill
\caption{The longitudinal (left) and lateral (right) fit error of models for vehicle trajectories in A1, A2 and WO with $T \in [3\mathrm{s}, 5\mathrm{s}, 8\mathrm{s}]$. "A, B" denote the model complexity $n=\hat{n}$ that maximizes AIC and BIC, respectively. The upper whisker denotes the 99.9\% percentile.}
\label{fig:trade_off_agt_vehicle}
\vspace{-0.5em}
\end{figure*}

\begin{figure*}[!h]
\centering
\hfill
\includegraphics[width=2.3in, height=3.1in]{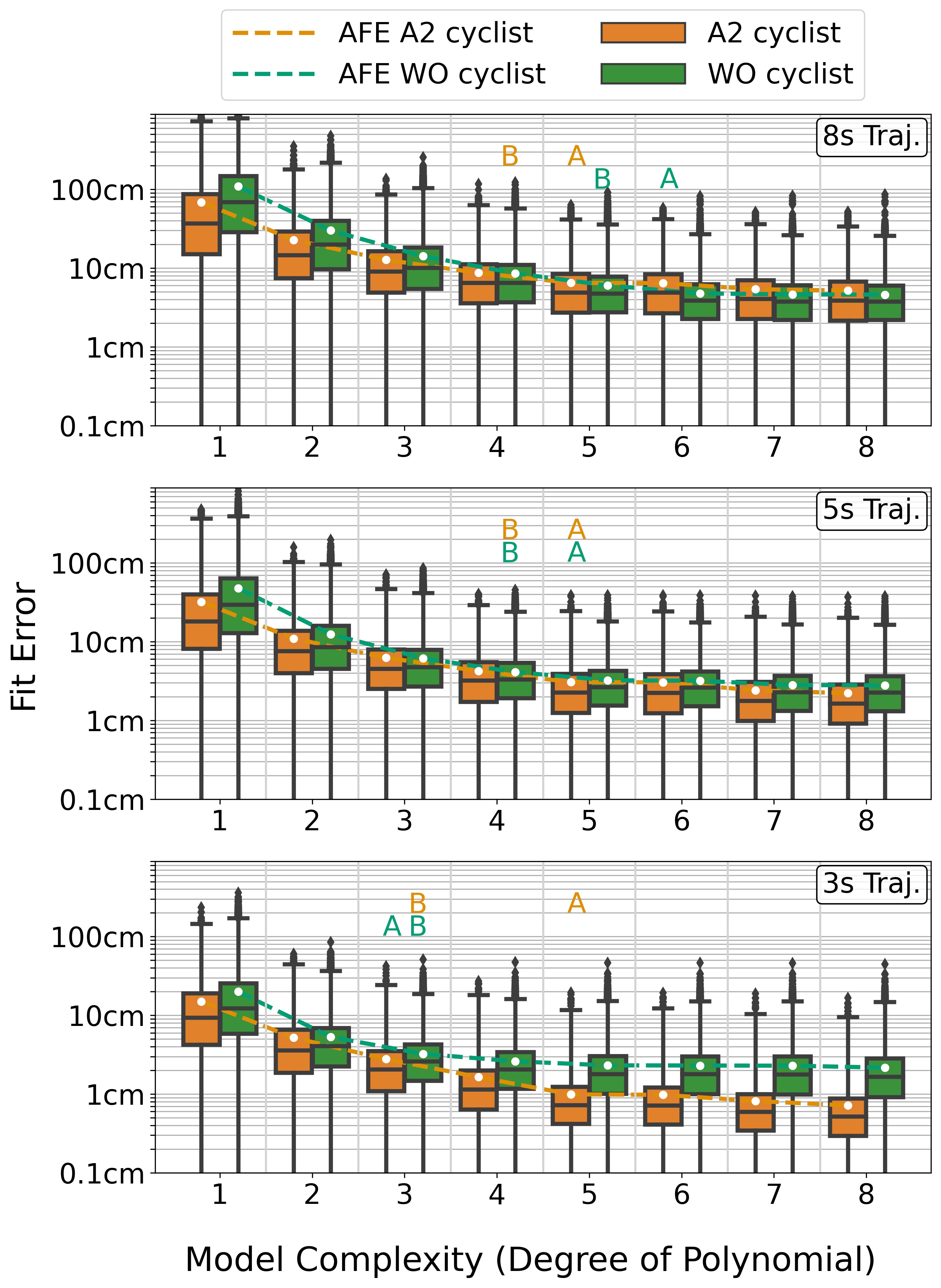} 
\hfill
\includegraphics[width=2.3in, height=3.1in]{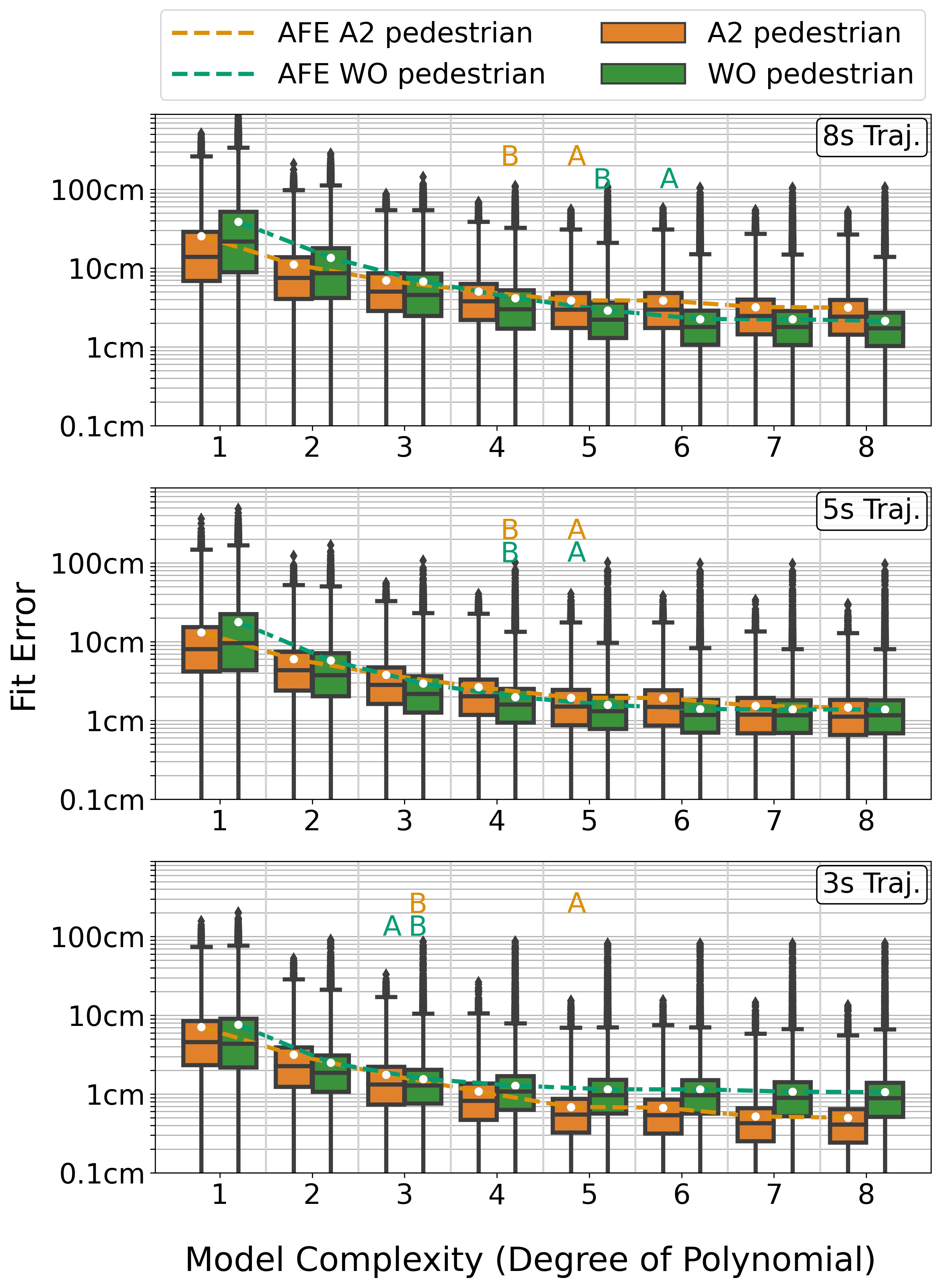} 
\hfill
\hfill
\caption{The fit error of models for cyclist (left) and pedestrian (right) trajectories in A2 and WO with $T \in [3\mathrm{s}, 5\mathrm{s}, 8\mathrm{s}]$. "A, B" denote the model complexity $n=\hat{n}$ that maximizes AIC and BIC, respectively. The upper whisker denotes the 99.9\% percentile.}
\label{fig:trade_off_agt_cyclist_pedestrian}
\vspace{-1.0em}
\end{figure*}

Figure \ref{fig:trade_off_agt_vehicle} shows Box-plots for the longitudinal and lateral fit error for vehicle trajectory samples with $T \in \{3\mathrm{s}, 5\mathrm{s}, 8\mathrm{s}\}$. Figure \ref{fig:trade_off_agt_cyclist_pedestrian} presents the fit error for cyclist and pedestrian trajectories with the same time horizons. We indicate the best model complexity according to AIC and BIC. Table \ref{tab: result summary} gives the numerical results for both ego and agent trajectories at the model complexity $n=\hat{n}$ that maximizes AIC.

\begin{table}[h]
\centering
\caption{Estimated Observation Noise for $T=5$s at $\hat{n}$ maximizing AIC}
\begin{tabularx}{3.2in}{c c !{\vrule width 1.5pt} >{\centering\arraybackslash}X | >{\centering\arraybackslash}X | >{\centering\arraybackslash}X}
\Xhline{3\arrayrulewidth}
\multicolumn{2}{c}{Datasets}&\multicolumn{1}{c}{A1} & \multicolumn{1}{c}{A2} & \multicolumn{1}{c}{WO}\\
\Xhline{3\arrayrulewidth}
\multirow{2}{*}{$\boldsymbol{\theta}_{ego}$} & $\sigma_{diag}$ [m] &0.024& 0.012 & 0.008\\
\cline{2-5} 
&$\sigma_{cov}$ $[m^2]$ &2e-4&3e-6& -1e-7\\
\hline
\hline
\multirow{5}{*}{\thead{$\boldsymbol{\theta}_{agent}$ \\ vehicle}} 
& $\sigma_{\alpha}$ [rad] &1e-3& 6e-4& 3e-4\\
\cline{2-5}
& $\sigma_{c}$ [m] &0.161 & 0.044& 0.017\\
\cline{2-5}
& \multirow{3}{*}{\thead{$\sigma_{r}$ [m] \\  $r=10\mathrm{m}, 20\mathrm{m}, 40\mathrm{m}$}} &0.128&0.055& 0.019\\
&&0.176&0.062& 0.027\\
& & 0.246&0.085& 0.042\\
\hline
\hline
\multirow{5}{*}{\thead{$\boldsymbol{\theta}_{agent}$ \\ cyclist}} 
& $\sigma_{\alpha}$ [rad] &-& 2e-4& 3e-4\\
\cline{2-5}
& $\sigma_{c}$ [m] &- & 0.027& 0.027\\
\cline{2-5}
& \multirow{3}{*}{\thead{$\sigma_{r}$ [m] \\  $r=10\mathrm{m}, 20\mathrm{m}, 40\mathrm{m}$}} &-&0.014& 0.010\\
&&-&0.022& 0.015\\
&&-&0.039& 0.026\\
\hline
\hline
\multirow{5}{*}{\thead{$\boldsymbol{\theta}_{agent}$ \\ pedestrian}} 
& $\sigma_{\alpha}$ [rad] &-& 3e-4& 2e-5\\
\cline{2-5}
& $\sigma_{c}$ [m] &- & 0.018& 0.015\\
\cline{2-5}
& \multirow{3}{*}{\thead{$\sigma_{r}$ [m] \\  $r=10\mathrm{m}, 20\mathrm{m}, 40\mathrm{m}$}} &-&0.007& 4e-4\\
&&-&0.010& 6e-4\\
& & -&0.017& 0.001\\
\Xhline{3\arrayrulewidth}
\vspace{-1em}
\label{tab: observation noise}
\end{tabularx}
\end{table}

\begin{figure*}[tbh]
\centering
\hfill
\includegraphics[width=3.2in]{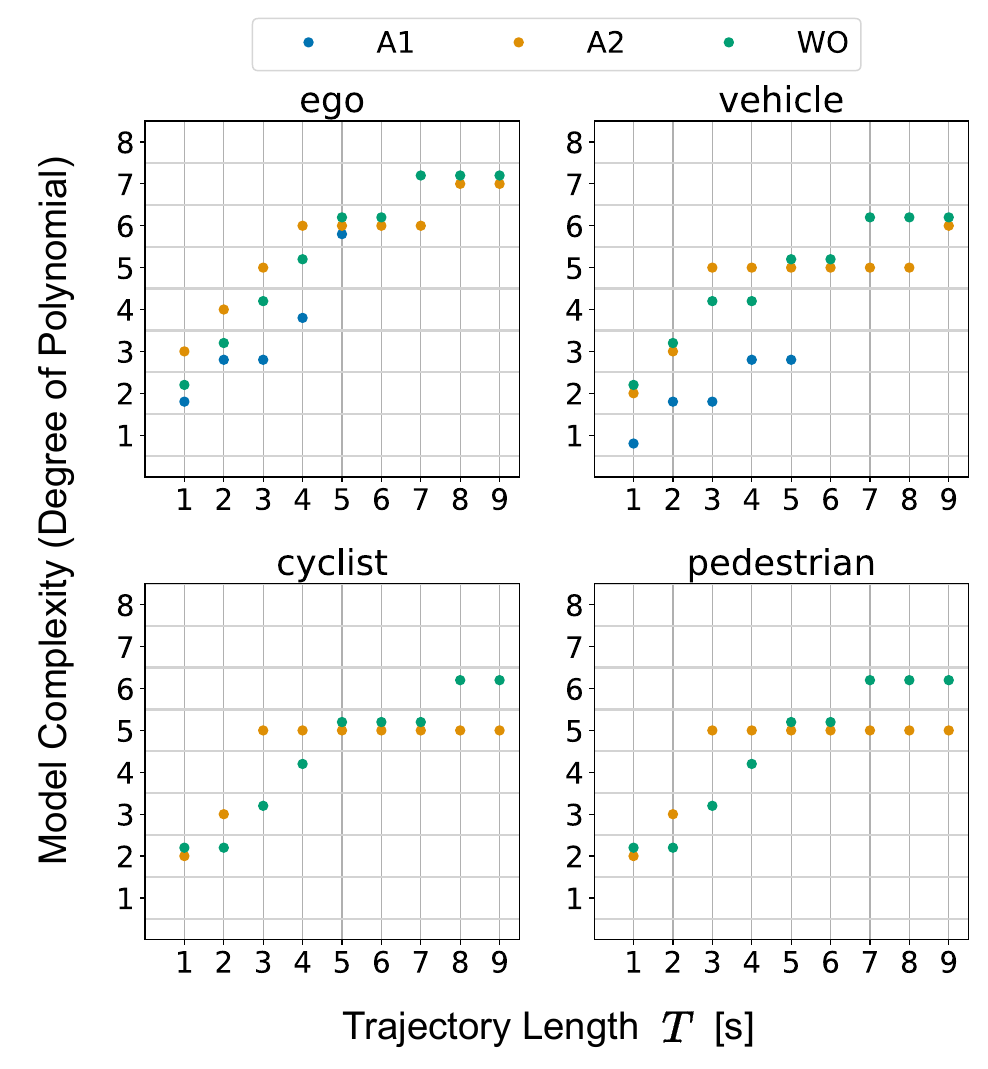} 
\hfill
\includegraphics[width=3.2in]{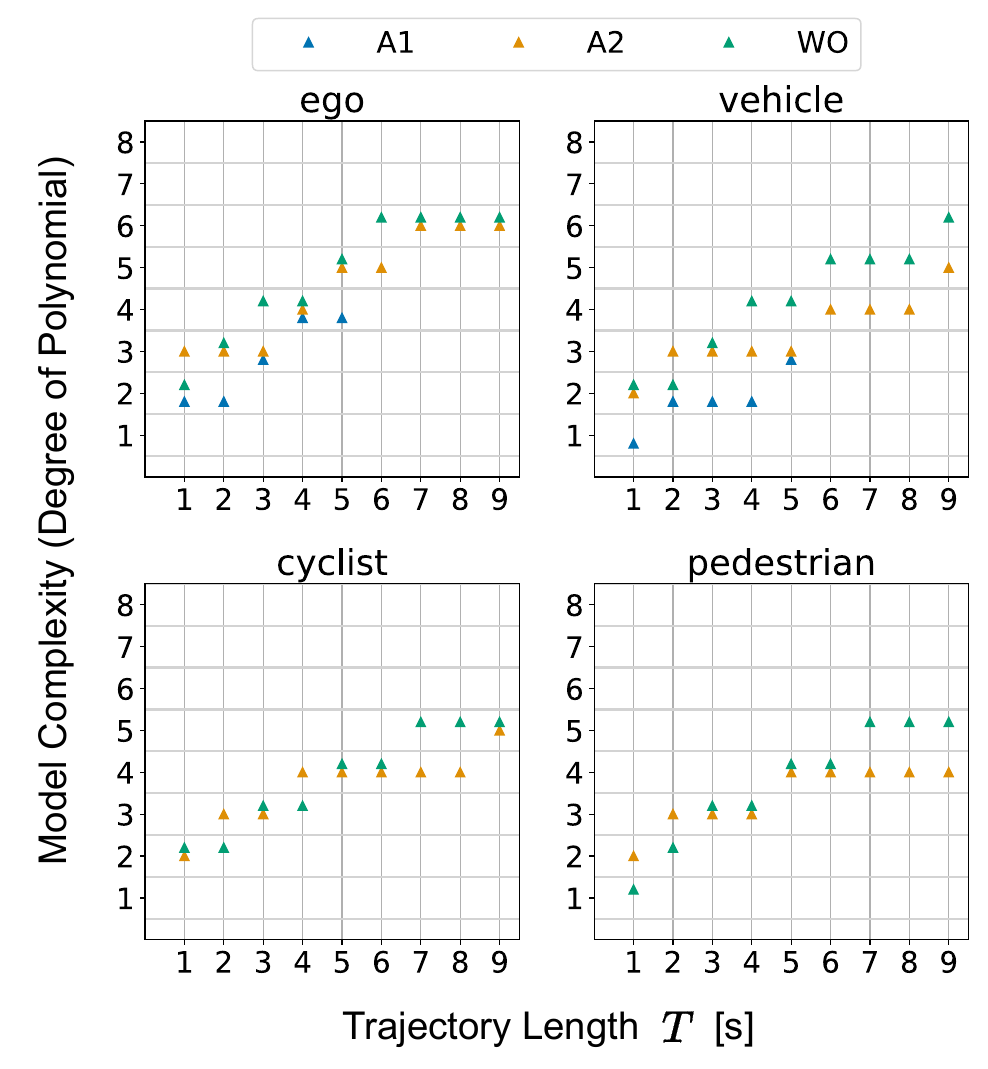} 
\hfill
\hfill
\caption{Optimal model complexity suggested by AIC (left) and BIC (right).}
\vspace{-0.5em}
\end{figure*}

Figure \ref{fig:trade_off_agt_vehicle}, Figure \ref{fig:trade_off_agt_cyclist_pedestrian} and Table \ref{tab: result summary} clearly show that the longer trajectories warrant higher model complexities for their representation, but also that the benefits of higher $n$ are diminishing beyond an optimal model $\hat{n}$ as indicated by AIC or BIC. Trajectories of various object classes require similar $\hat{n}$ according to AIC or BIC in one dataset. Overall, we see how linear models of moderate complexity can represent trajectories with very high fidelity. E.g., the $6^{\mathrm{th}}$ degree polynomial can approximate the 8-seconds vehicle trajectories with $3.7$cm longitudinal and $1.6$cm lateral AFE in WO. However, we also observe large deviations ($>1$m). Inspecting these samples, we find they correspond to physically implausible measurements due to timing jitter or tracking loss that are not excluded by the RTS-Smoother.

\begin{table}[!h]
\vspace{-0.5em}
\begin{threeparttable}
\caption{Fit Error of ego and agent trajectories [m] \\ (Lon.\:\vline\:Lat.)}
\centering
\begin{tabularx}{3.5in}{c c c!{\vrule width 1.5pt}>{\centering\arraybackslash}X|>{\centering\arraybackslash}X|>{\centering\arraybackslash}X}
\Xhline{2\arrayrulewidth}
&\multirow{2}{*}{T [s]}&& \multicolumn{3}{c}{Dataset} \\
\cline{4-6}
&& & A1& A2& WO \\
\Xhline{3\arrayrulewidth}

\multirow{9}{*}{ego} 
&\multirow{3}{*}{3} & $\hat{n}$ & 3 & 5 &4 \\
&&AFE& 0.022\:\vline\:0.002& 0.004\:\vline\:0.001 & 0.004\:\vline\:0.001\\
&&99.9\%&  0.173\:\vline\:0.120 & 0.094\:\vline\:0.012 & 0.066\:\vline\:0.021\\
\cline{2-6}

&\multirow{3}{*}{5} & $\hat{n}$ & 5 & 6 &6 \\
&&AFE& 0.022\:\vline\:0.004& 0.007\:\vline\:0.002 & 0.005\:\vline\:0.002\\
&&99.9\%&  0.157\:\vline\:0.111 &0.143\:\vline\:0.030 & 0.071\:\vline\:0.025\\
\cline{2-6}

&\multirow{3}{*}{8} & $\hat{n}$ & - & 7 & 7 \\ &&AFE& -& 0.019\:\vline\:0.006 & 0.013\:\vline\:0.004\\
&&99.9\%&  - & 0.251\:\vline\:0.090 & 0.114\:\vline\:0.069\\

\hline
\hline
\multirow{9}{*}{\thead{agent \\ vehicle}} 
&\multirow{3}{*}{3} & $\hat{n}$ & 2 & 5 &4 \\
&&AFE& 0.185\:\vline\:0.060& 0.019\:\vline\:0.005 & 0.016\:\vline\:0.005\\
&&99.9\%&  1.753\:\vline\:0.835 & 0.312\:\vline\:0.114 & 0.359\:\vline\:0.056\\
\cline{2-6}

&\multirow{3}{*}{5} & $\hat{n}$ & 3 & 5 &5 \\
&&AFE& 0.191\:\vline\:0.065& 0.051\:\vline\:0.016 & 0.022\:\vline\:0.008\\
&&99.9\%&  1.774\:\vline\:0.848 &0.735\:\vline\:0.284 & 0.408\:\vline\:0.090\\
\cline{2-6}

&\multirow{3}{*}{8} & $\hat{n}$ & - & 5 & 6 \\ &&AFE& -& 0.093\:\vline\:0.033 & 0.037\:\vline\:0.016\\
&&99.9\%&  - & 1.115\:\vline\:0.461 & 0.483\:\vline\:0.186\\

\hline
\hline
\multirow{9}{*}{\thead{agent \\ cyclist}} 
&\multirow{3}{*}{3} & $\hat{n}$ & - & 5 &3 \\
&&AFE& -& 0.008\:\vline\:0.004 & 0.025\:\vline\:0.016\\
&&99.9\%&  - & 0.098\:\vline\:0.054 & 0.179\:\vline\:0.130\\
\cline{2-6}

&\multirow{3}{*}{5} & $\hat{n}$ & - & 5 &5 \\
&&AFE& - & 0.022\:\vline\:0.016 & 0.025\:\vline\:0.016\\
&&99.9\%&  - &0.216\:\vline\:0.183 & 0.174\:\vline\:0.123\\
\cline{2-6}

&\multirow{3}{*}{8} & $\hat{n}$ & - & 5 & 6 \\ &&AFE& -& 0.042\:\vline\:0.040 & 0.031\:\vline\:0.028\\
&&99.9\%&  - & 0.364\:\vline\:0.358 & 0.217\:\vline\:0.242\\

\hline
\hline
\multirow{9}{*}{\thead{agent \\ pedestrian}} 
&\multirow{3}{*}{3} & $\hat{n}$ & - & 5 &3 \\
&&AFE& - & 0.005\:\vline\:0.004 & 0.011\:\vline\:0.009\\
&&99.9\%&  - & 0.058\:\vline\:0.046 & 0.091\:\vline\:0.070\\
\cline{2-6}

&\multirow{3}{*}{5} & $\hat{n}$ & - & 5 &5 \\
&&AFE& - & 0.012\:\vline\:0.012 & 0.011\:\vline\:0.009\\
&&99.9\%&  - &0.135\:\vline\:0.126 & 0.086\:\vline\:0.066\\
\cline{2-6}

&\multirow{3}{*}{8} & $\hat{n}$ & - & 5 & 6 \\ &&AFE& -& 0.023\:\vline\:0.026 & 0.016\:\vline\:0.012\\
&&99.9\%&  - & 0.257\:\vline\:0.255 & 0.133\:\vline\:0.108\\
\Xhline{2\arrayrulewidth}
\label{tab: result summary}
\vspace{-1em}
\end{tabularx}
\begin{tablenotes}
\small \item 
$\hat{n}$ denotes the best polynomial degree according to AIC. $99.9\%$ means the 99.9 percentile of the fit error.
\end{tablenotes}
\end{threeparttable}
\vspace{-1em}
\end{table}


\subsubsection{Fit Error vs. total Displacement Error}
Let us compare the price we pay for the bias introduced by our basis functions to the associated benefits in computational efficiency.

\begin{figure}[!h]
\centering
\includegraphics[width=3.0in, height=1.8in]{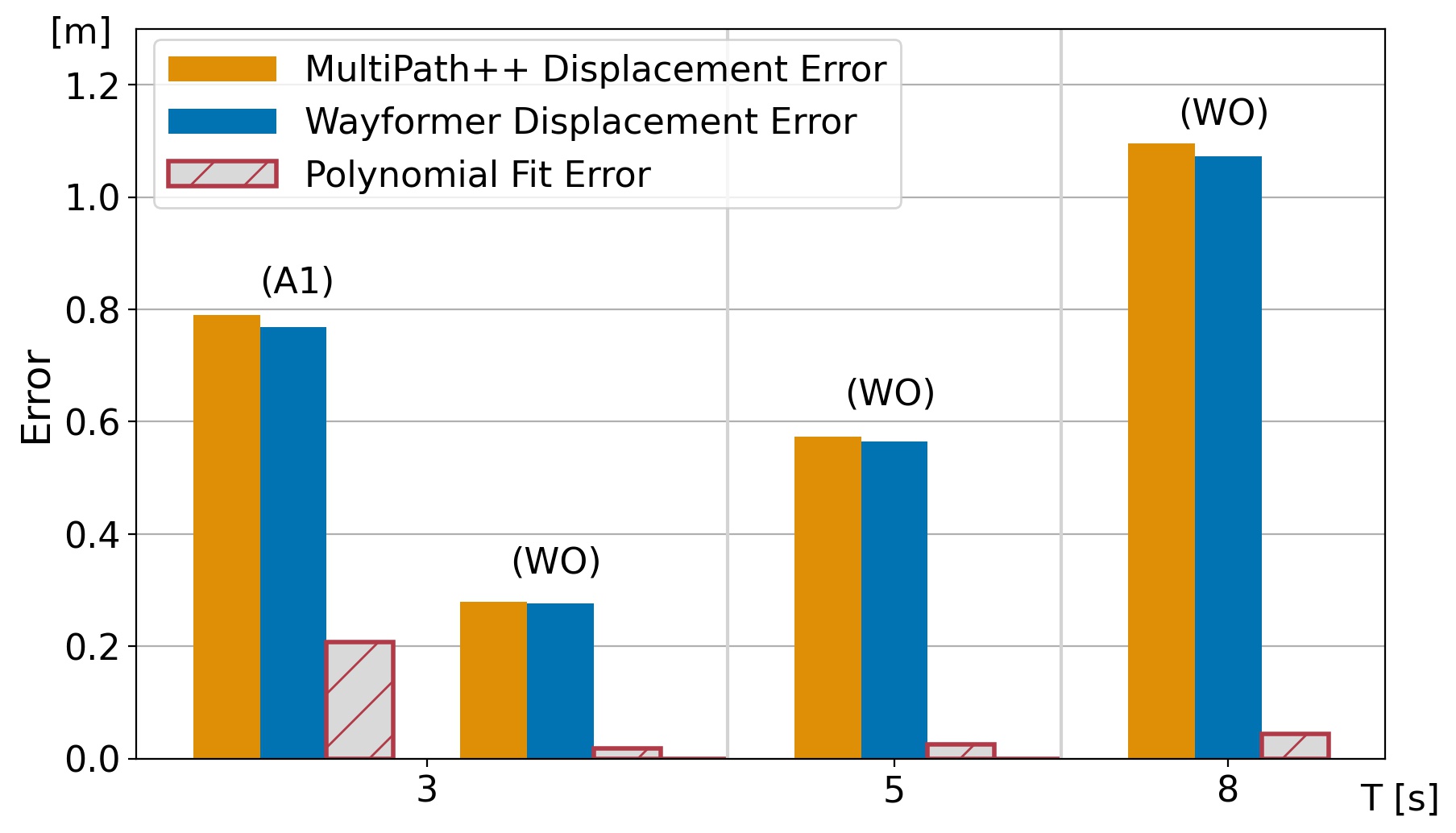} 
\caption{The average fit error $AFE$ of polynomials at $\hat{n}$ maximizing AIC and the average displacement error $minADE_k$ ($k=6$) of Wayformer\cite{nayakanti_wayformer_2022} and MultiPath++\cite{varadarajan_multipath++_2022} for vehicle trajectories with $T \in [3\mathrm{s}, 5\mathrm{s}, 8\mathrm{s}]$. Latest prediction results can be found at \cite{argoverse_leaderboard, waymo_leaderboard}.}
\label{fig:representation vs prediction}
\vspace{-1.0em}
\end{figure}

Figure \ref{fig:representation vs prediction} compares the \emph{fit error} of our linear trajectory models to the \emph{total displacement error} over predicted vehicle trajectories of state-of-the-art prediction methods with unbiased sequence-based trajectory representation. This comparison indicates to what extent a linear trajectory model may, at the most, impact the performance of a prediction system. We compare to $minADE_k$, i.e. the minimum average displacement error over top $k$ most-likely predicted trajectories. Note that both fit error and total displacement error are dependent on the observation noise level as visualized in Figure \ref{fig:error_decompostion}. If the observation noise level is low, as in WO, the representation error of a linear trajectory formulation, upper bounded by fit error, is negligible in the prediction task. It is still significantly lower than the displacement error of predictions in datasets, where the noise level is likely higher such as A1. If a prediction method can perfectly predict the parameters of future trajectories, its total displacement error will be close to the fit error of polynomials.

On the other hand, we deem the computational benefits of linear models large. For $n=5$, a trajectory is specified completely by six kinematic constraints (position, velocity and acceleration) at start and end points. Since three kinematic constraints at the start can be estimated entirely from past observation, the prediction problem is in fact reduced to predicting the kinematics at the end point. Hence, $50\%$ of the accuracy of trajectory prediction is due to exact tracking of initial kinematics and $50\%$ is due to accurate prediction of kinematics at the \emph{end} of the prediction horizon, only.

\section{Conclusion}     %

Object trajectories for various object classes can be modelled with high fidelity by simple linear combinations of polynomial basis functions. \emph{Independent} of any particular prediction method, we have characterized the trade-off between model complexity and fit error by an empirical analysis of several large public datasets. Using an Empirical Bayes approach, we have estimated models for observation noise and prior distributions over model parameters. The estimated observation noise parameters can (and should) be considered when training trajectory prediction models with a Gaussian Log-Likelihood loss, particularly when combining different datasets. The prior parameters can inform the motion models of trajectory tracking and filtering models \cite{reichardt_trajectories_2022} or regularize trajectory prediction models.

It is important to note that the fit error introduced by a linear formulation is small compared to the total displacement errors of current state-of-the-art prediction models. This indicates that the inherent bias in linear models is much smaller than the epistemic uncertainty in the prediction task. We further suggest using linear trajectory models in future works with inherent mathematical benefits. 


\section*{ACKNOWLEDGMENT}
Authors would like to thank Andreas Philipp for many useful discussions and acknowledge funding from the German Federal Ministry for Economic Affairs and Climate Action within the project “KI Wissen – Automotive AI powered by Knowledge”.

\begin{newpage}
\printbibliography[title = {References}]
\end{newpage}

\onecolumn
\begin{appendices}




\newpage
\section{10 5-seconds vehicle trajectories with highest fit error in A1 (Fitted with $\hat{n} = 3$)}
\centering
\includegraphics[width=5.5in, height=1.8in]{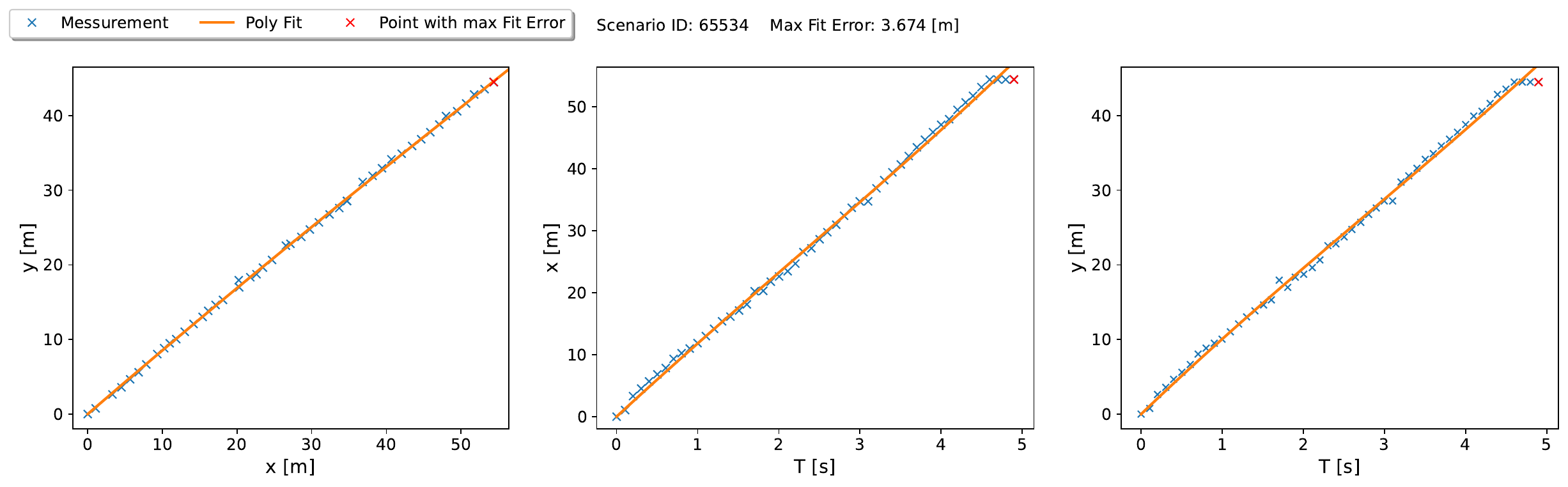} 

\includegraphics[width=5.5in, height=1.8in]{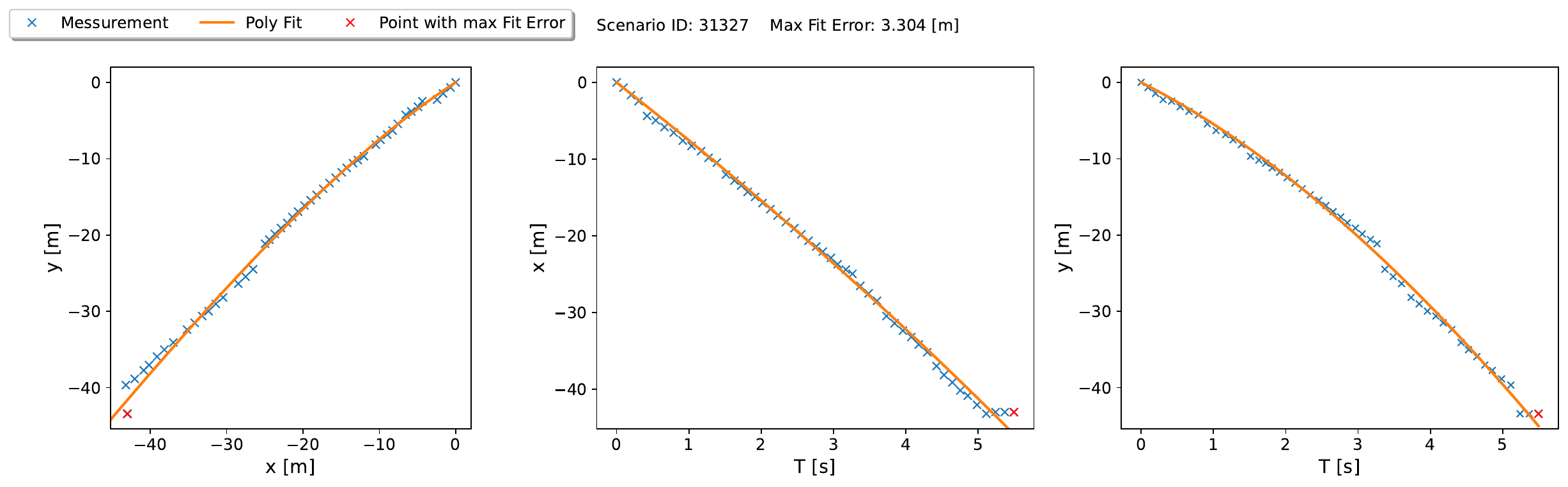} 

\includegraphics[width=5.5in, height=1.8in]{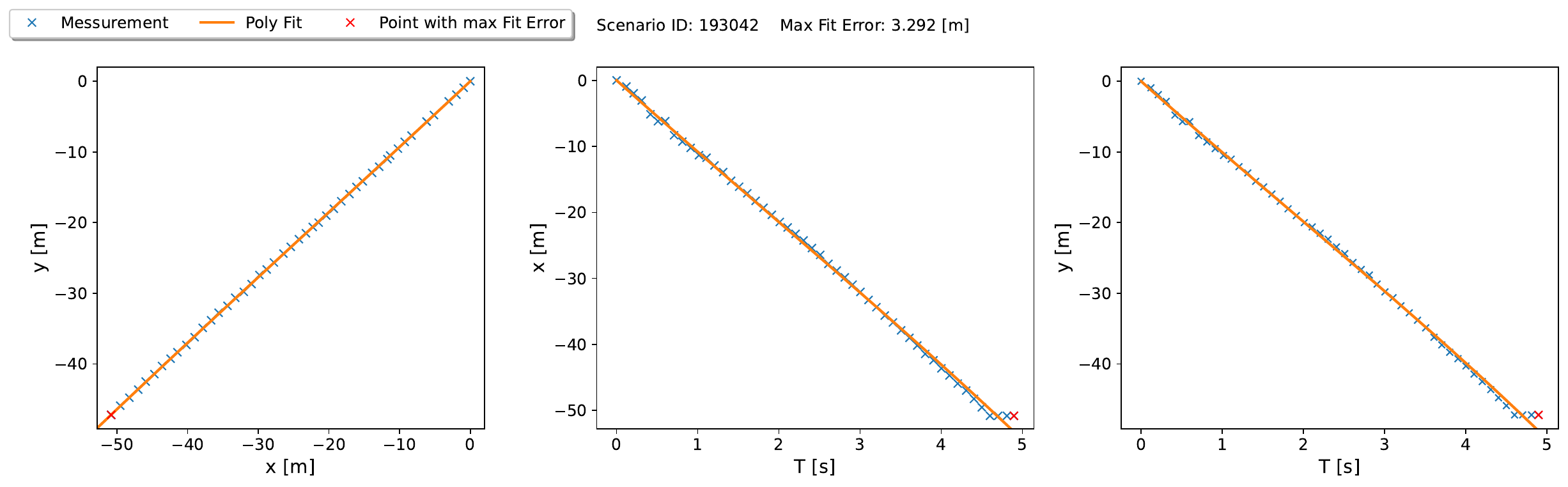} 

\includegraphics[width=5.5in, height=1.8in]{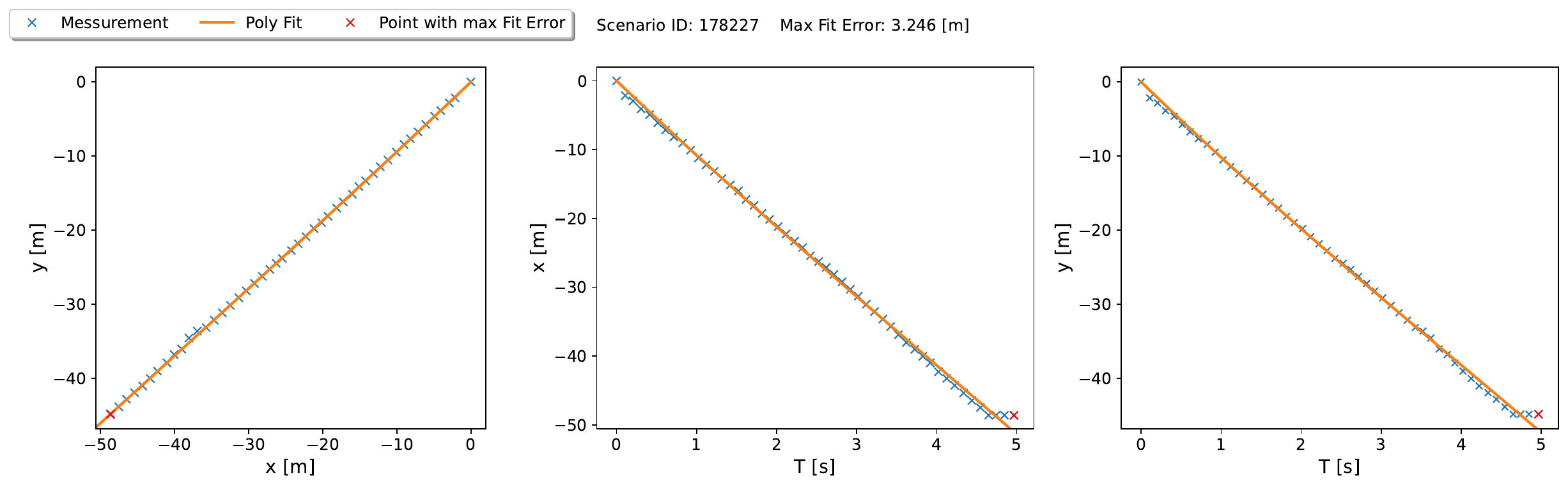} 

\includegraphics[width=5.5in, height=1.7in]{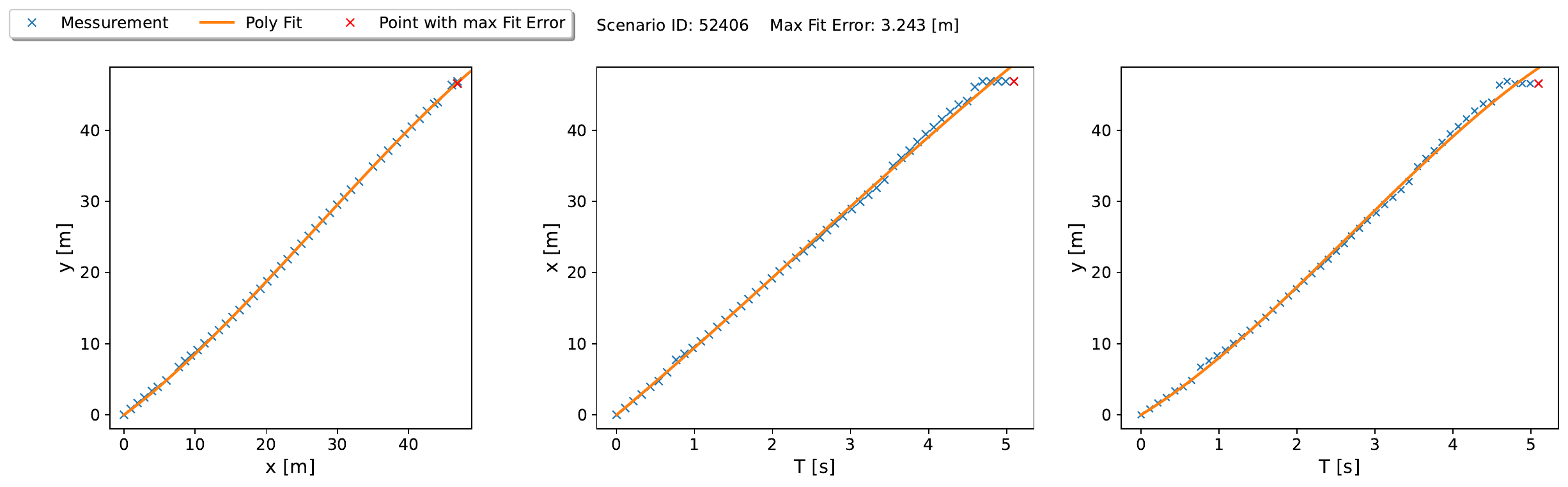} 

\includegraphics[width=5.5in, height=1.8in]{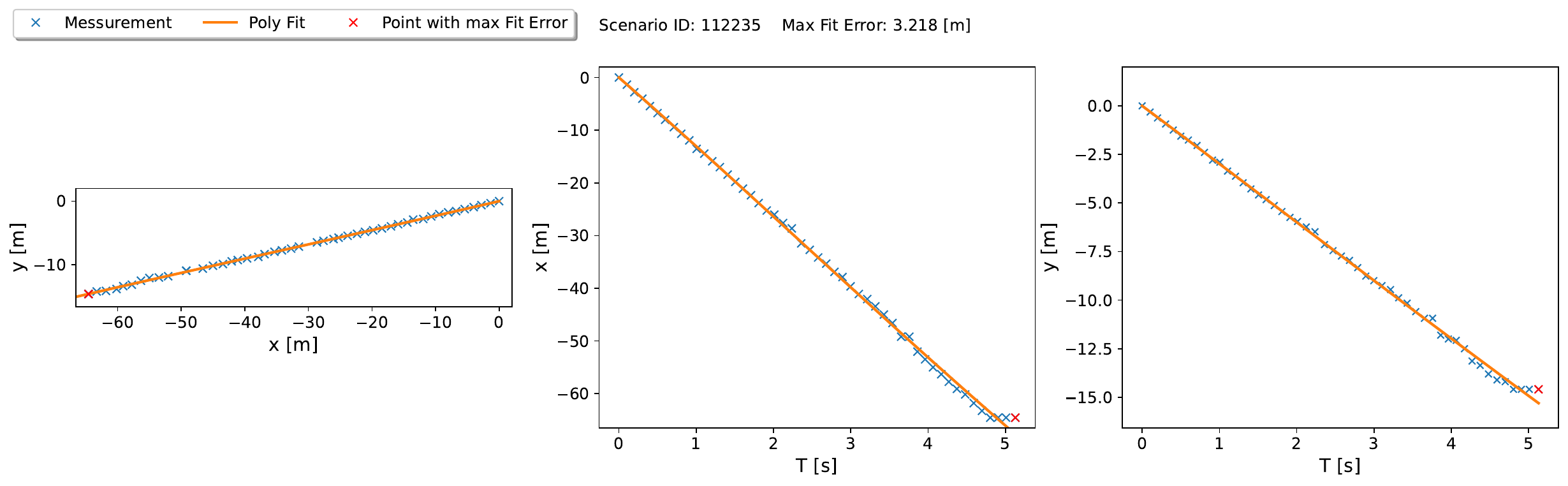} 

\includegraphics[width=5.5in, height=1.8in]{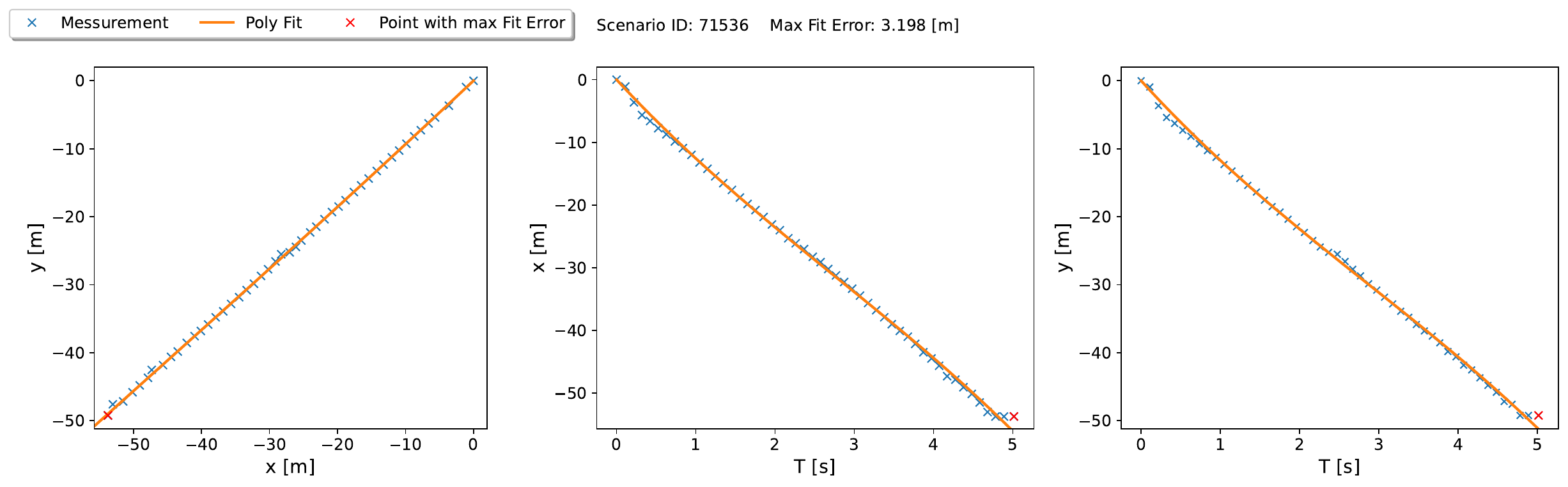} 

\includegraphics[width=5.5in, height=1.8in]{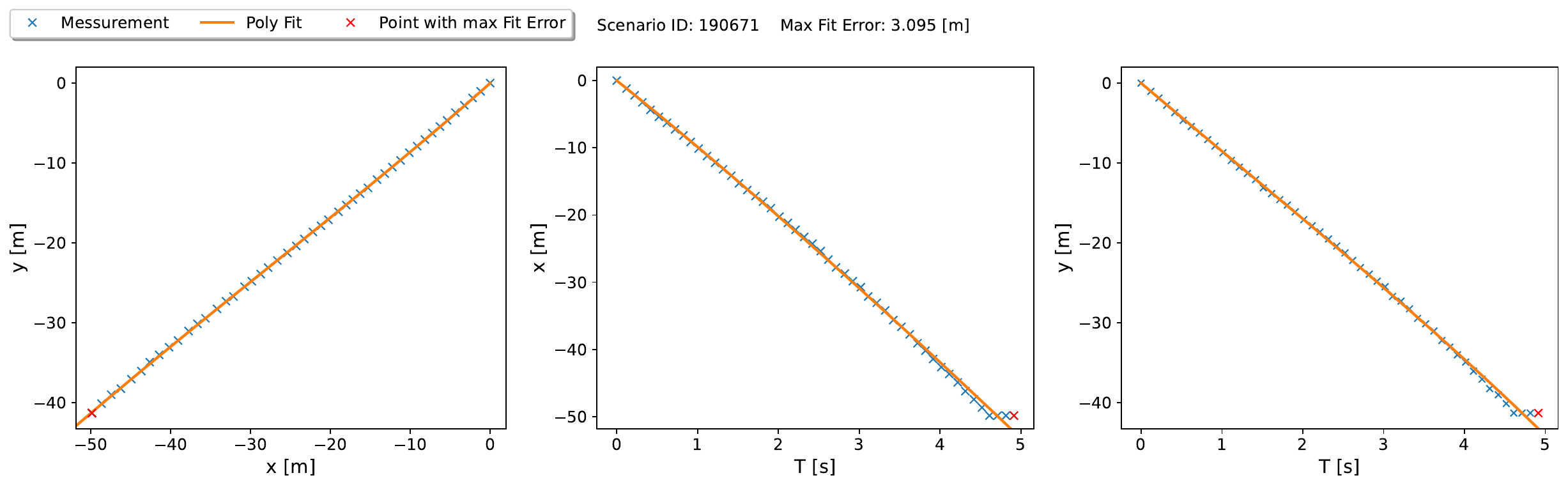} 

\includegraphics[width=5.5in, height=1.8in]{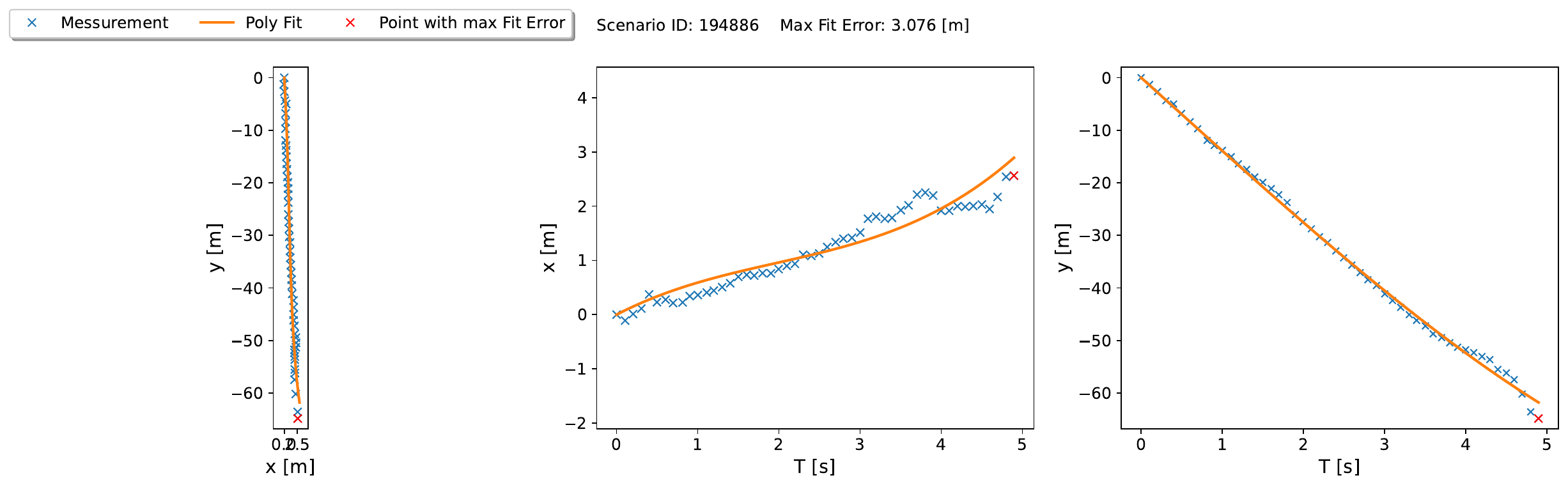} 

\includegraphics[width=5.5in, height=1.7in]{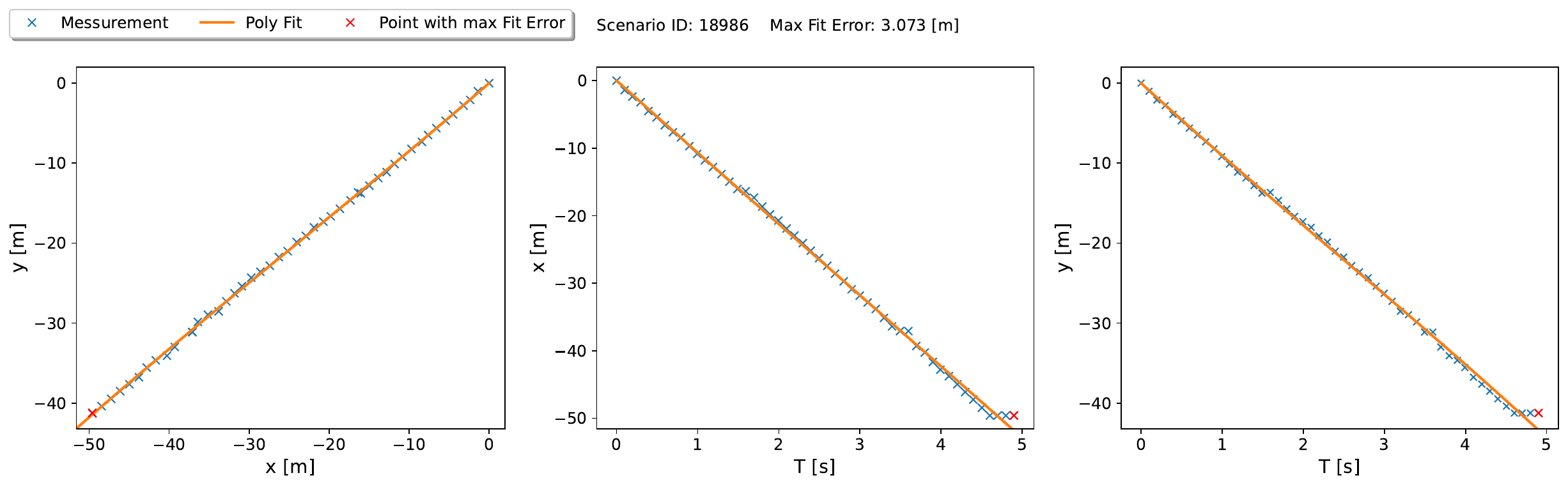} 

\section{10 random 5-seconds vehicle trajectories in A1 (Fitted with $\hat{n} = 3$)}
\centering
\includegraphics[width=5.5in, height=1.8in]{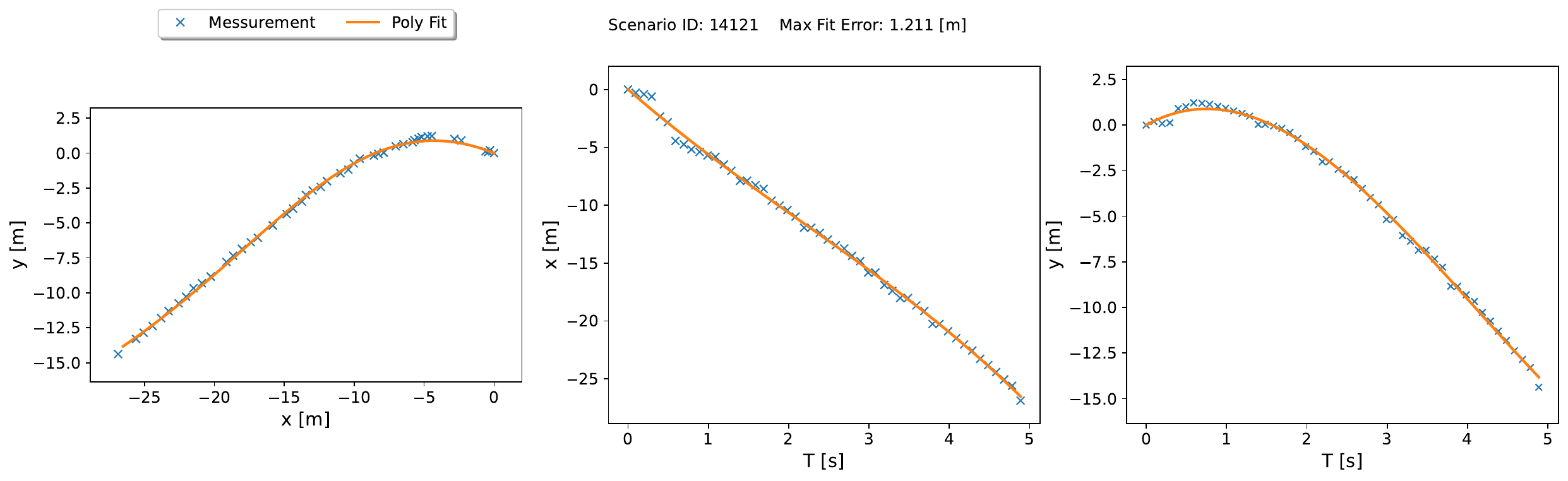} 

\includegraphics[width=5.5in, height=1.8in]{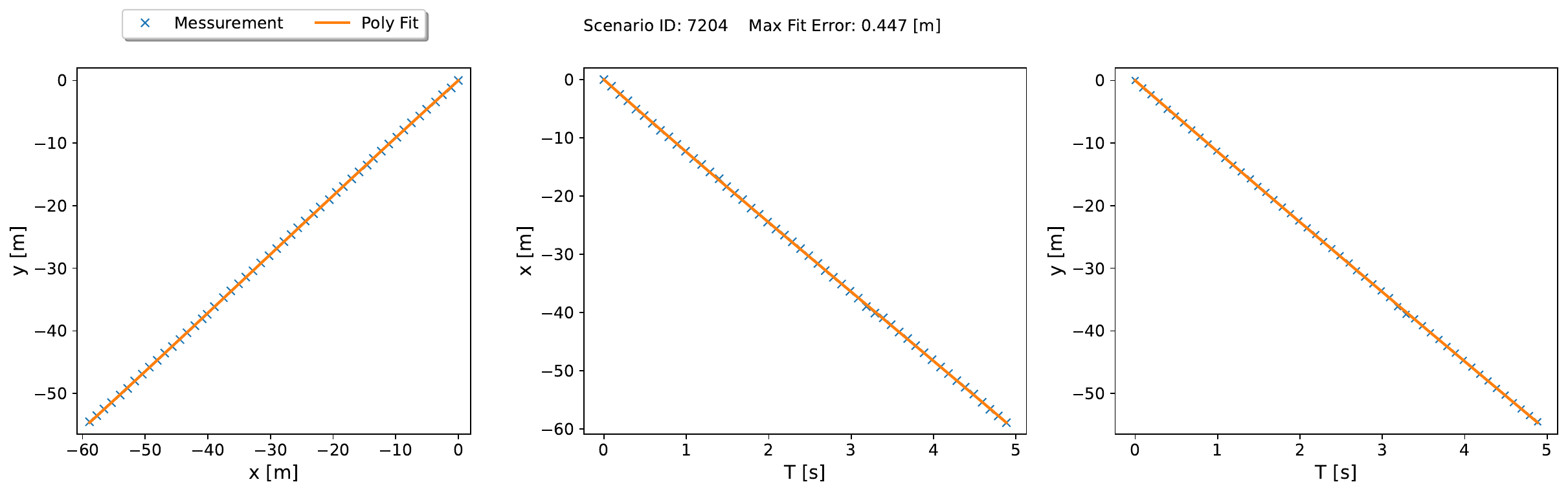} 

\includegraphics[width=5.5in, height=1.8in]{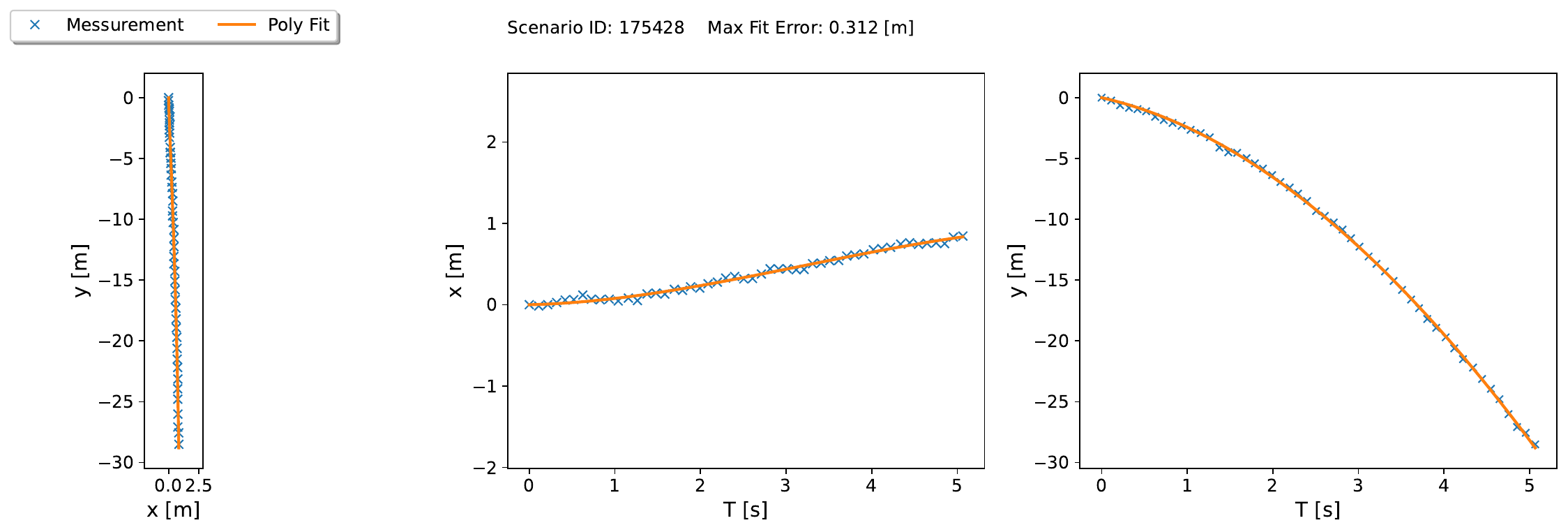} 

\includegraphics[width=5.5in, height=1.8in]{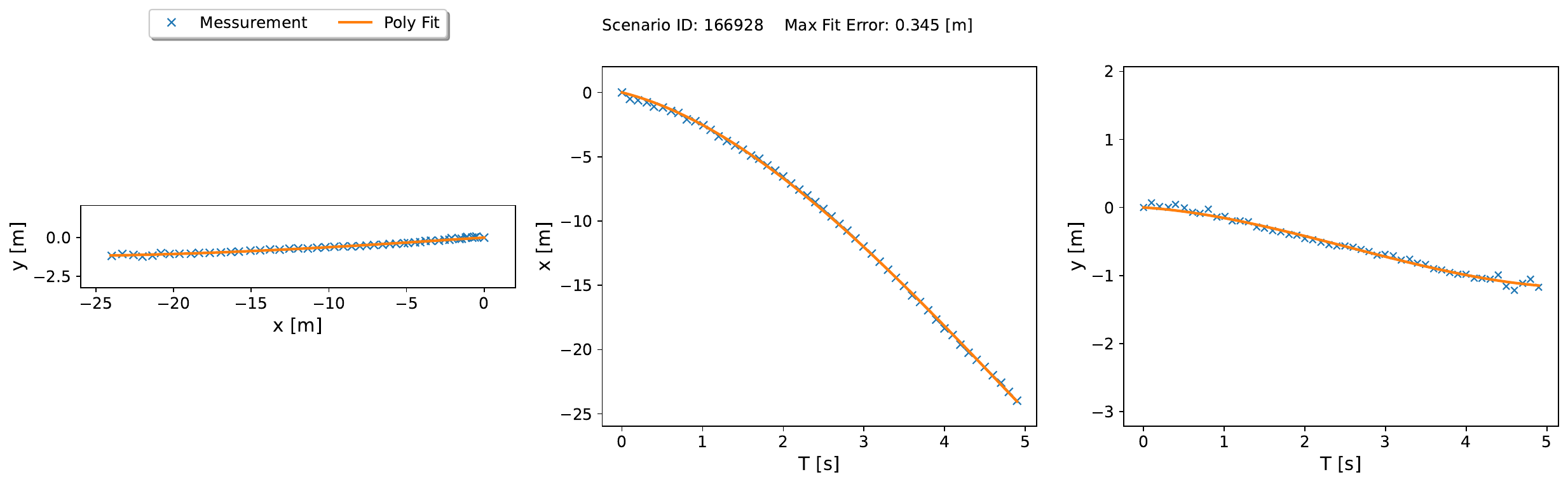} 

\includegraphics[width=5.5in, height=1.7in]{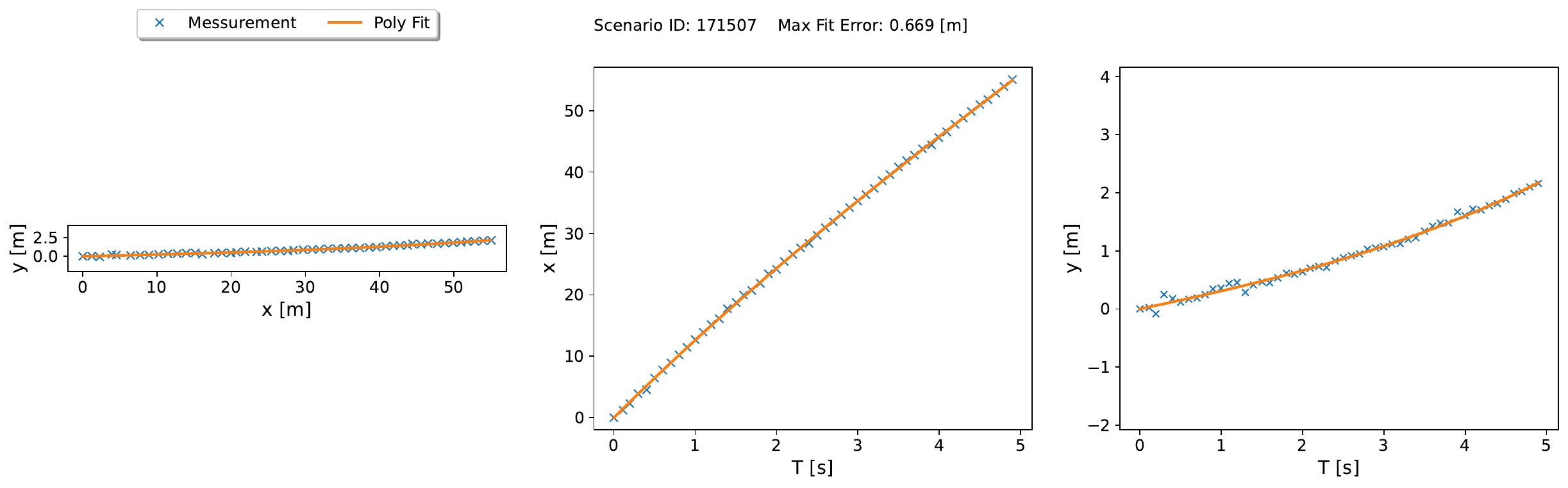} 

\includegraphics[width=5.5in, height=1.8in]{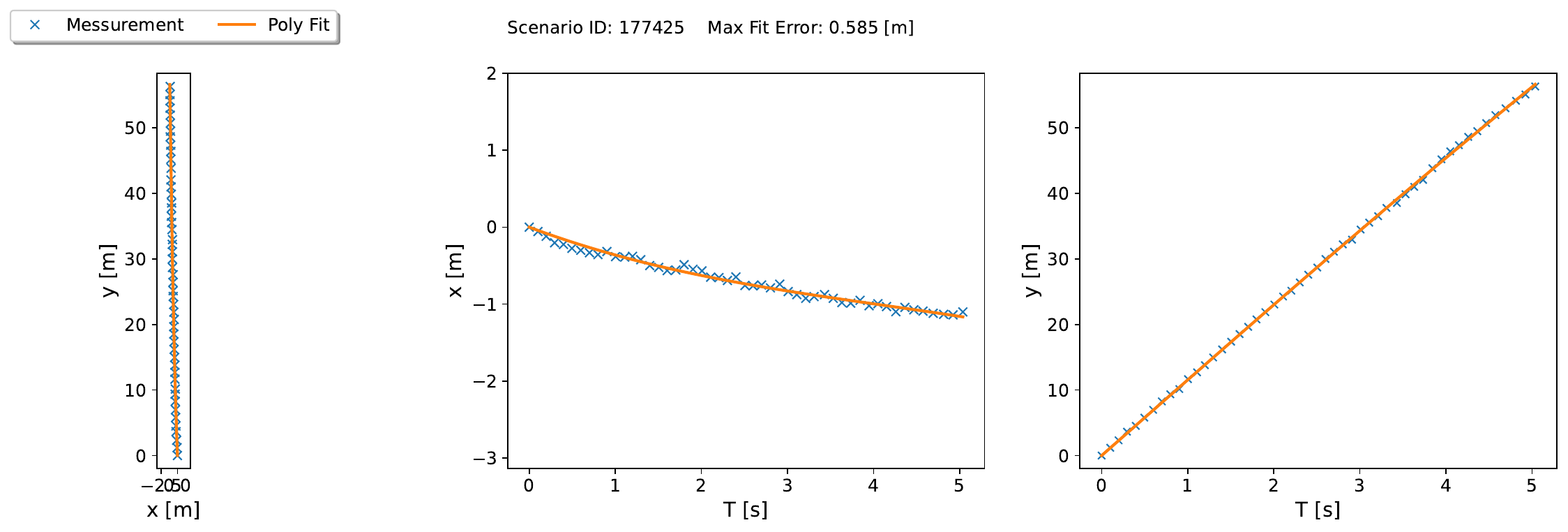} 

\includegraphics[width=5.5in, height=1.8in]{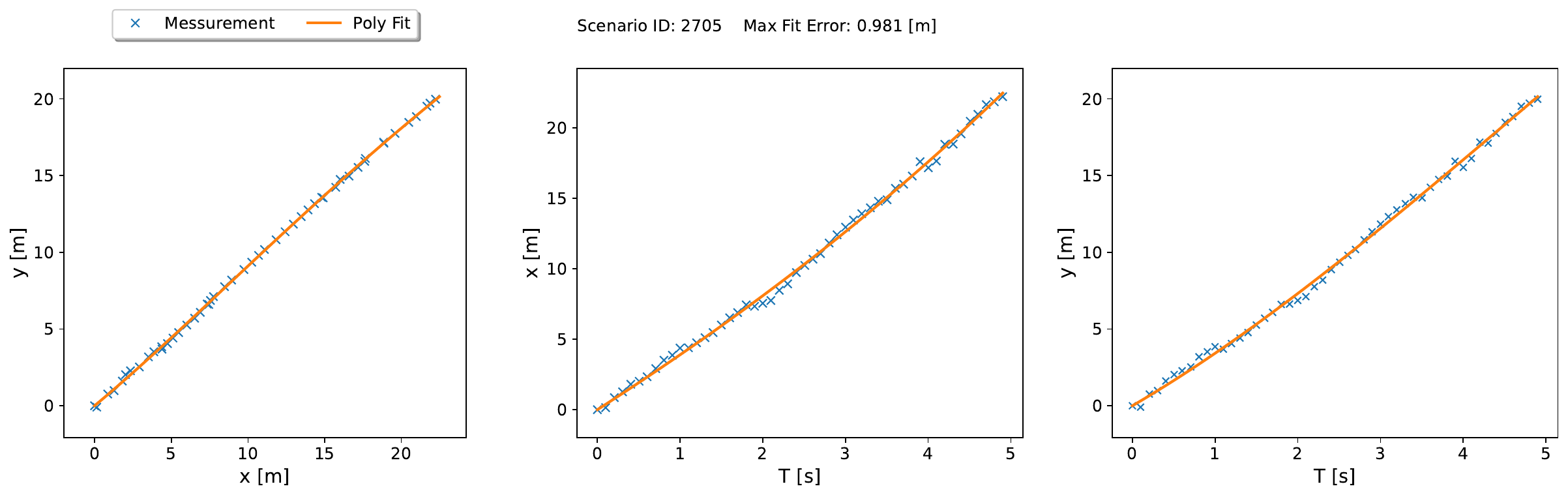} 

\includegraphics[width=5.5in, height=1.8in]{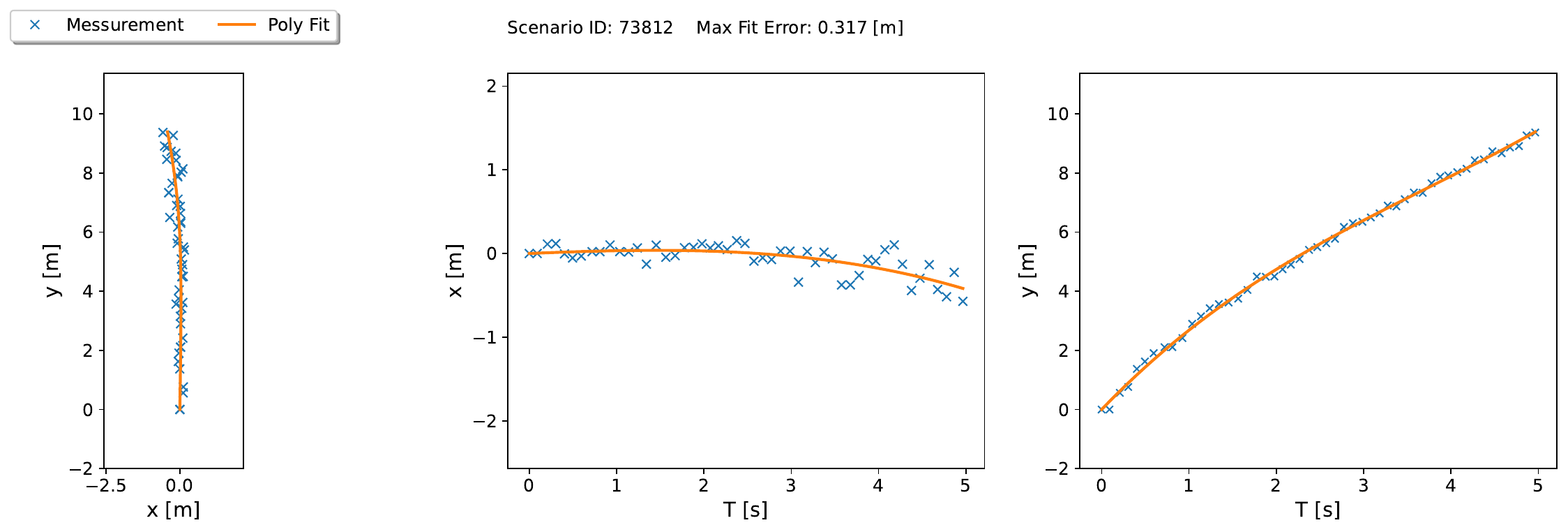} 

\includegraphics[width=5.5in, height=1.8in]{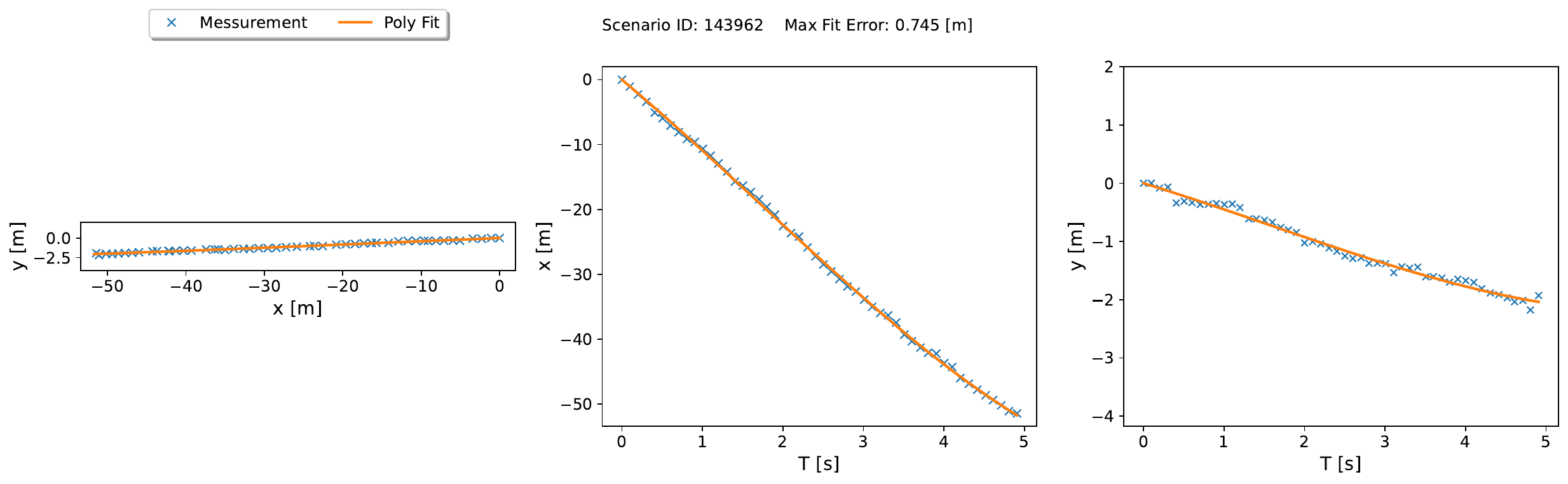} 

\includegraphics[width=5.5in, height=1.7in]{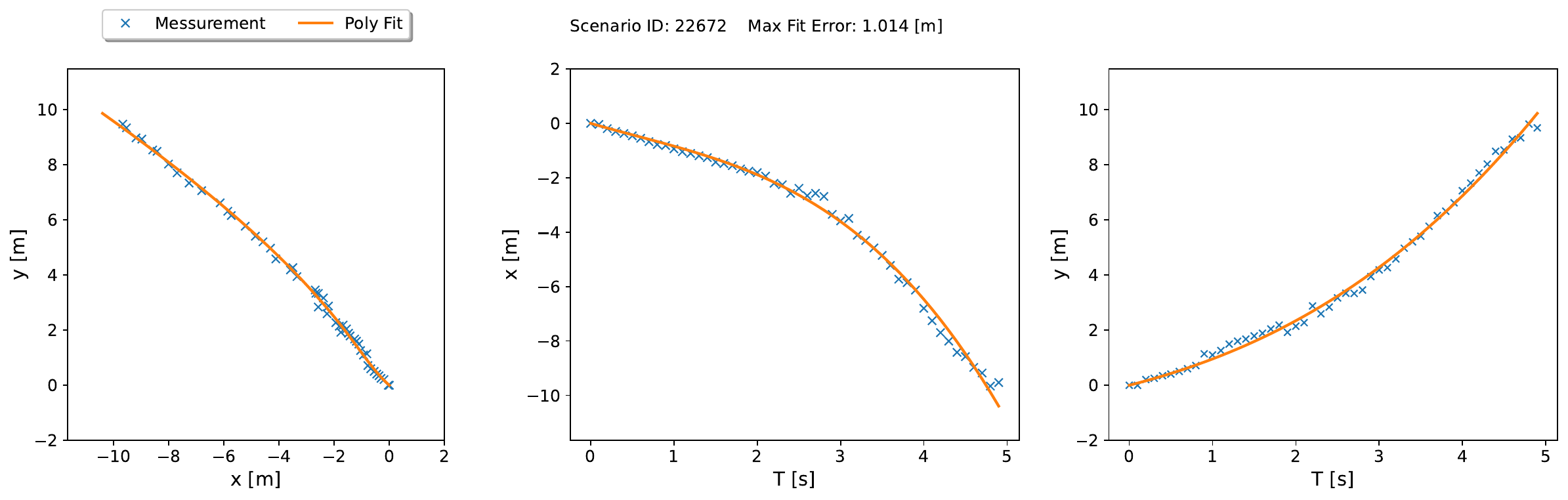} 

\section{10 5-seconds vehicle trajectories with highest fit error in A2 (Fitted with $\hat{n} = 5$)}

\centering
\includegraphics[width=5.5in, height=1.8in]{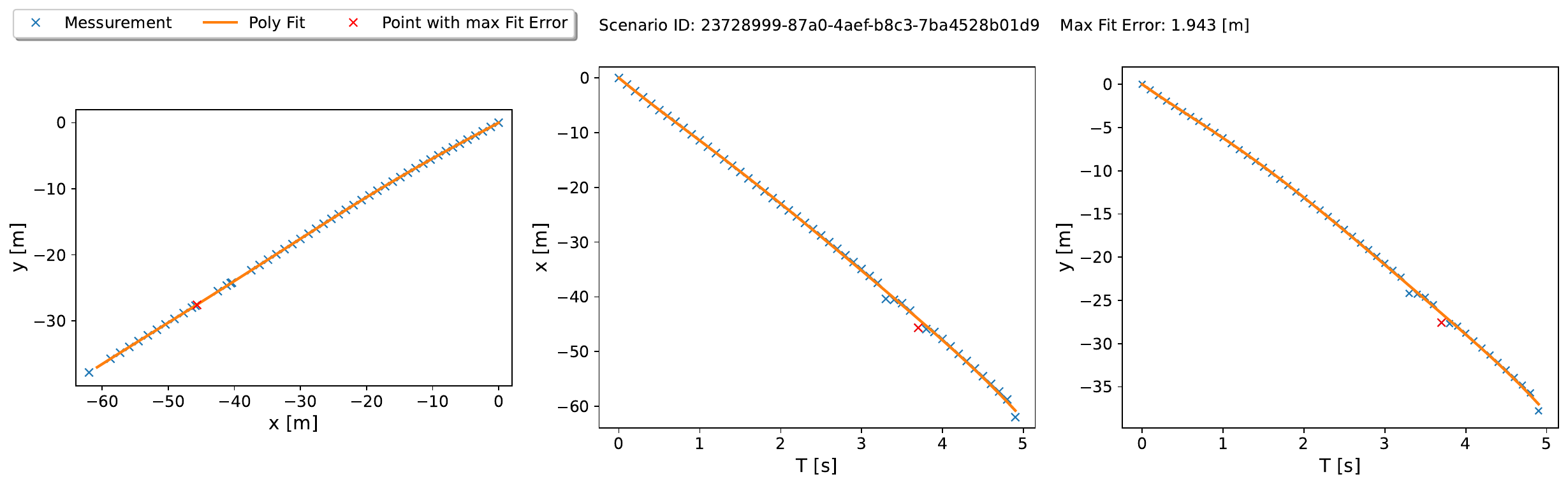} 

\includegraphics[width=5.5in, height=1.8in]{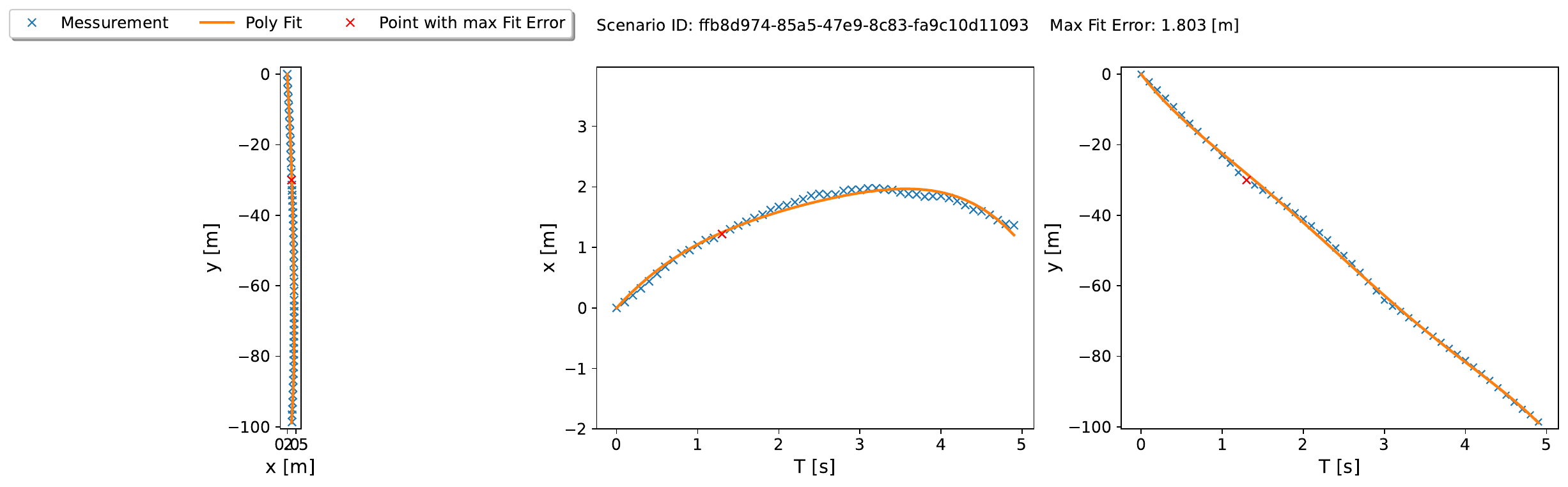}

\includegraphics[width=5.5in, height=1.8in]{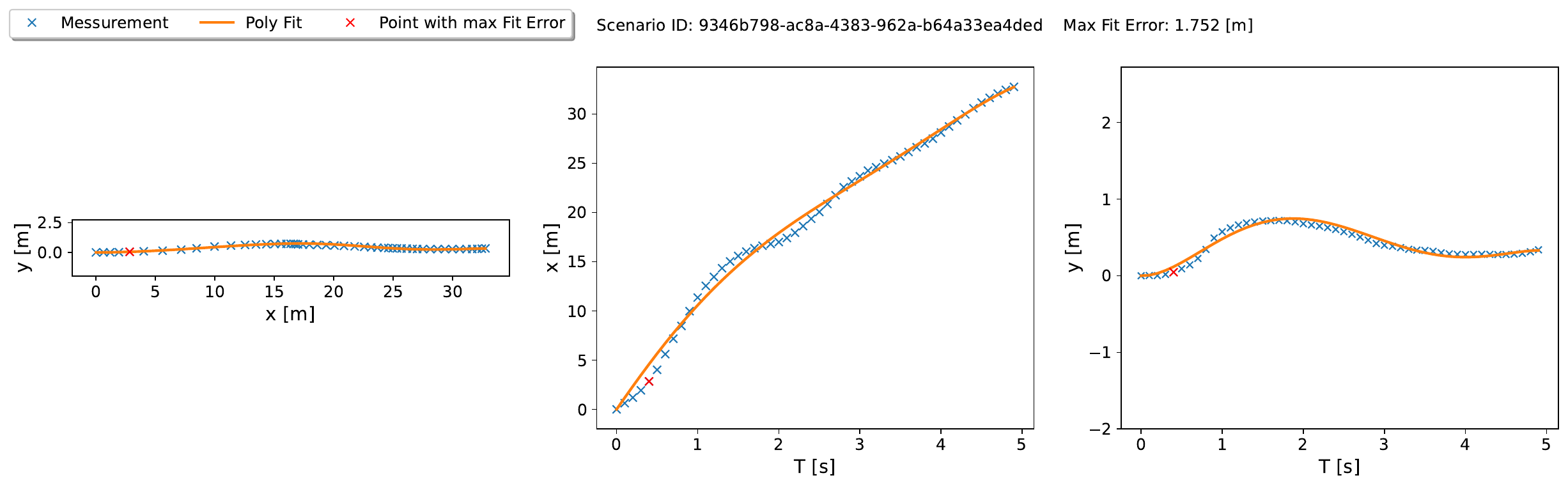} 

\includegraphics[width=5.5in, height=1.8in]{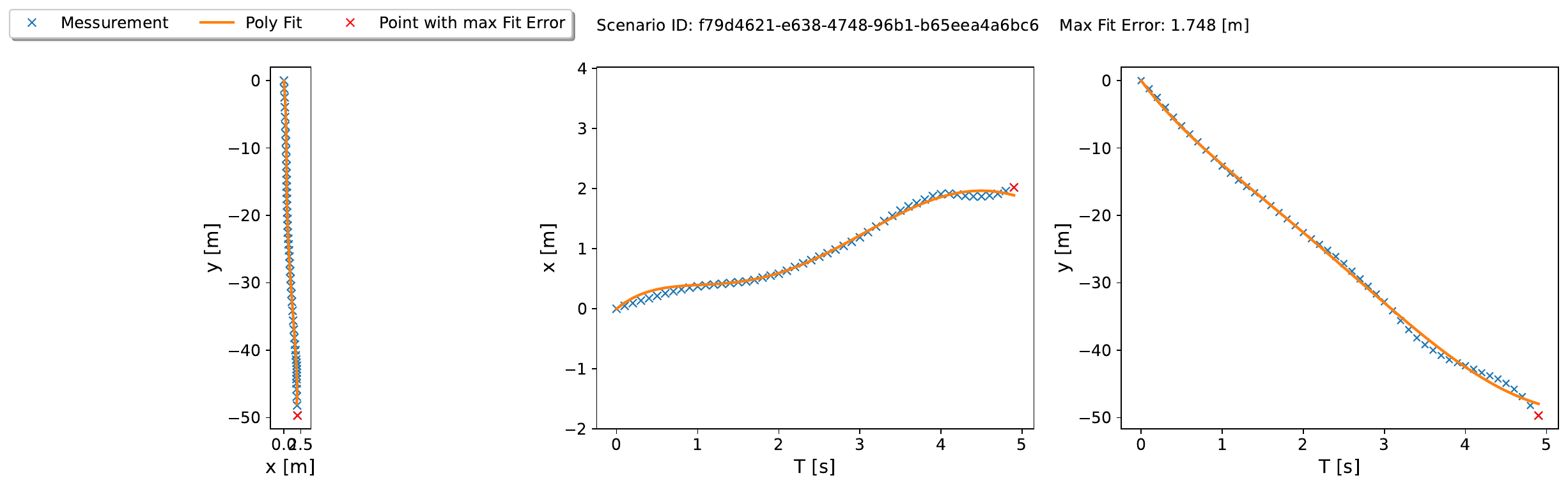} 

\includegraphics[width=5.5in, height=1.7in]{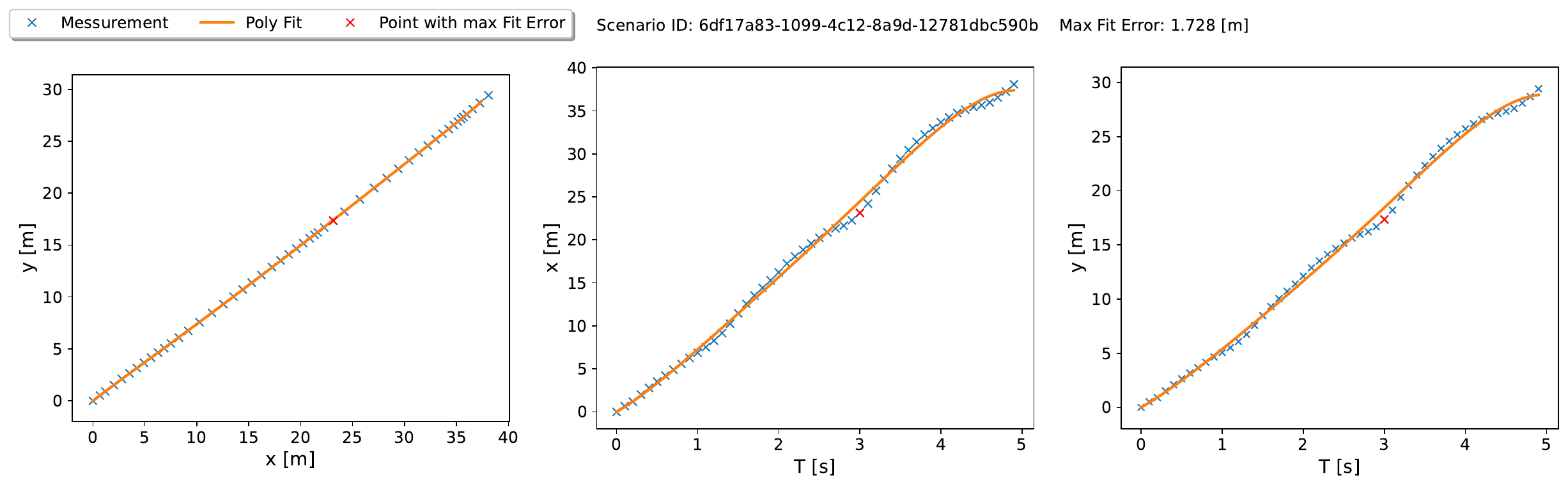} 

\includegraphics[width=5.5in, height=1.8in]{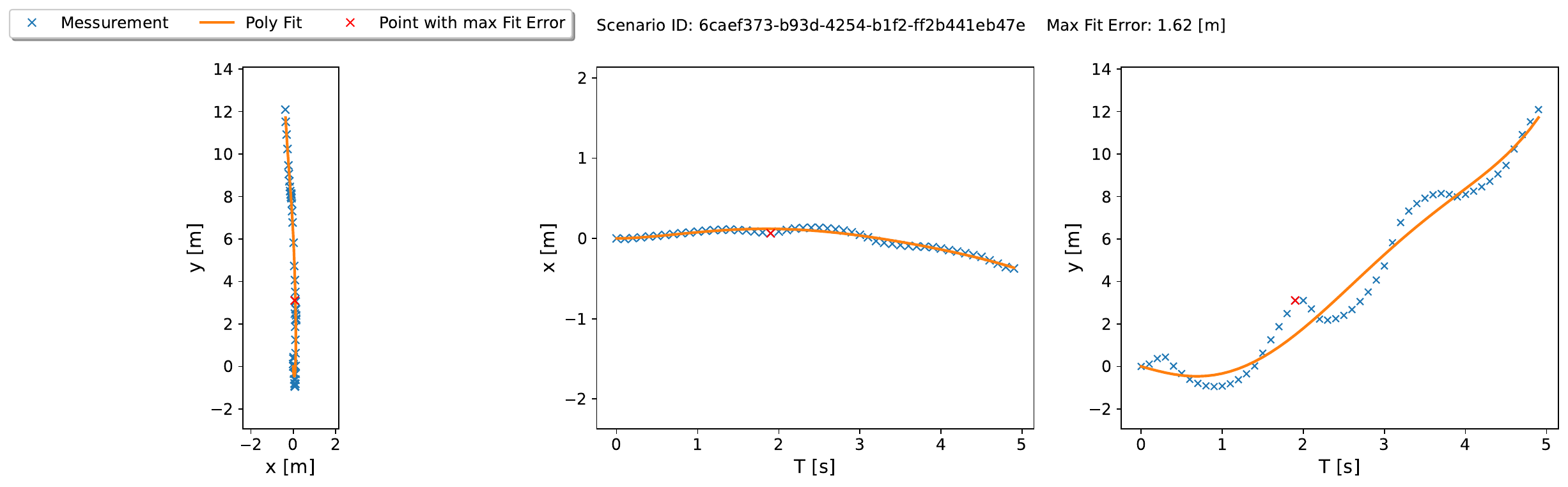} 

\includegraphics[width=5.5in, height=1.8in]{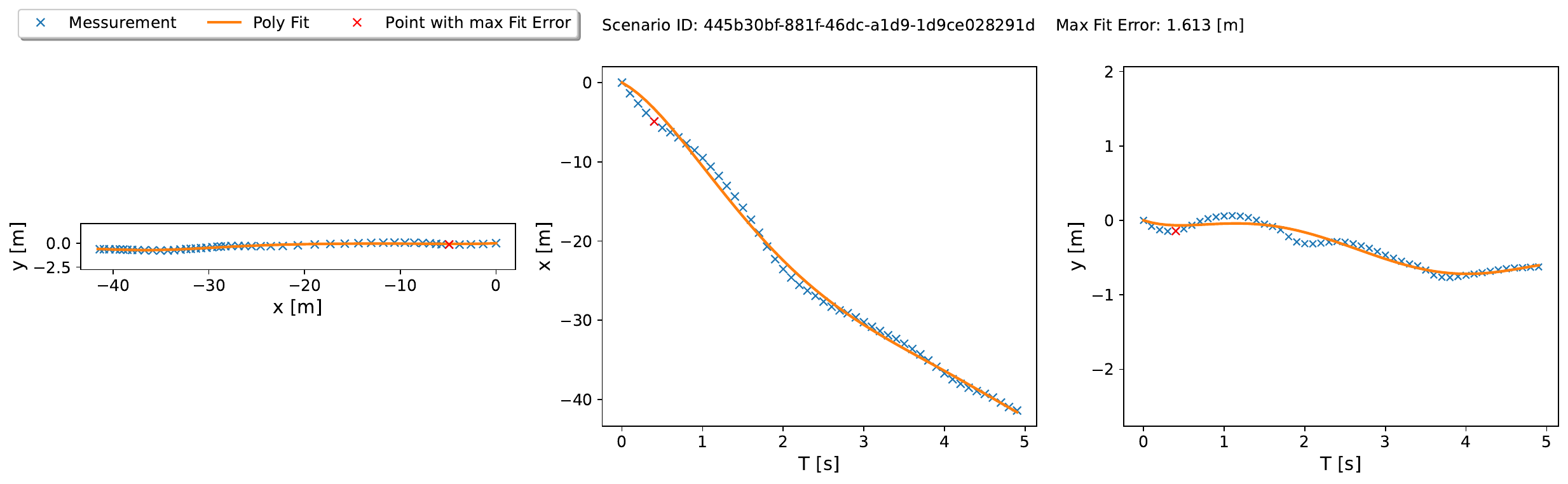} 

\includegraphics[width=5.5in, height=1.8in]{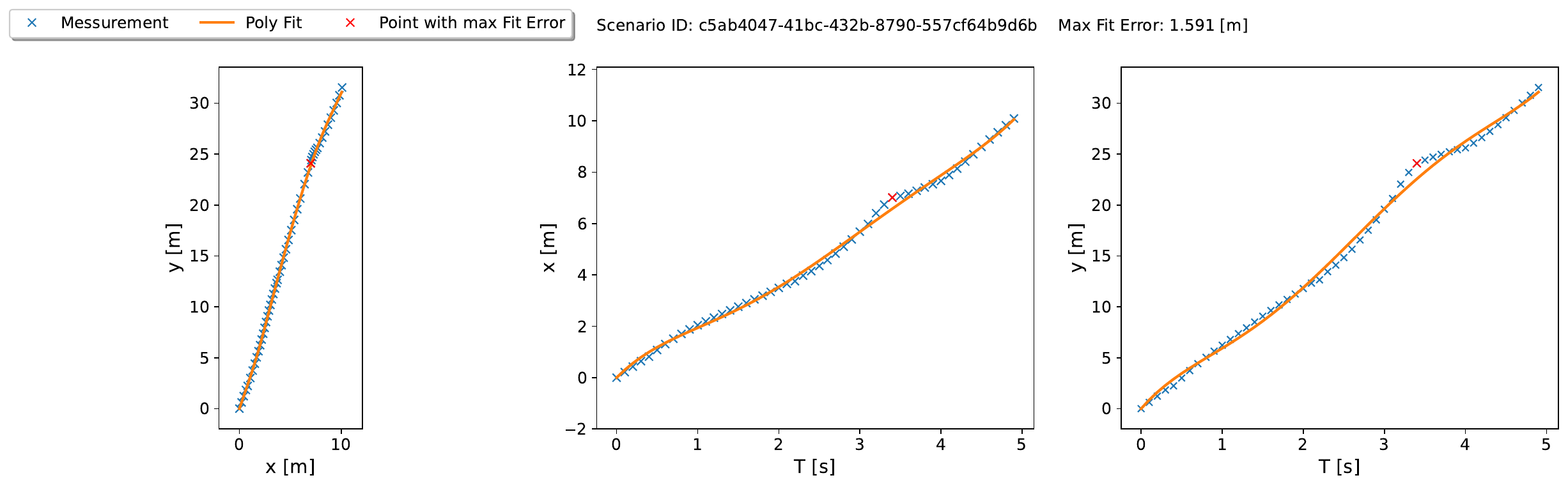} 

\includegraphics[width=5.5in, height=1.8in]{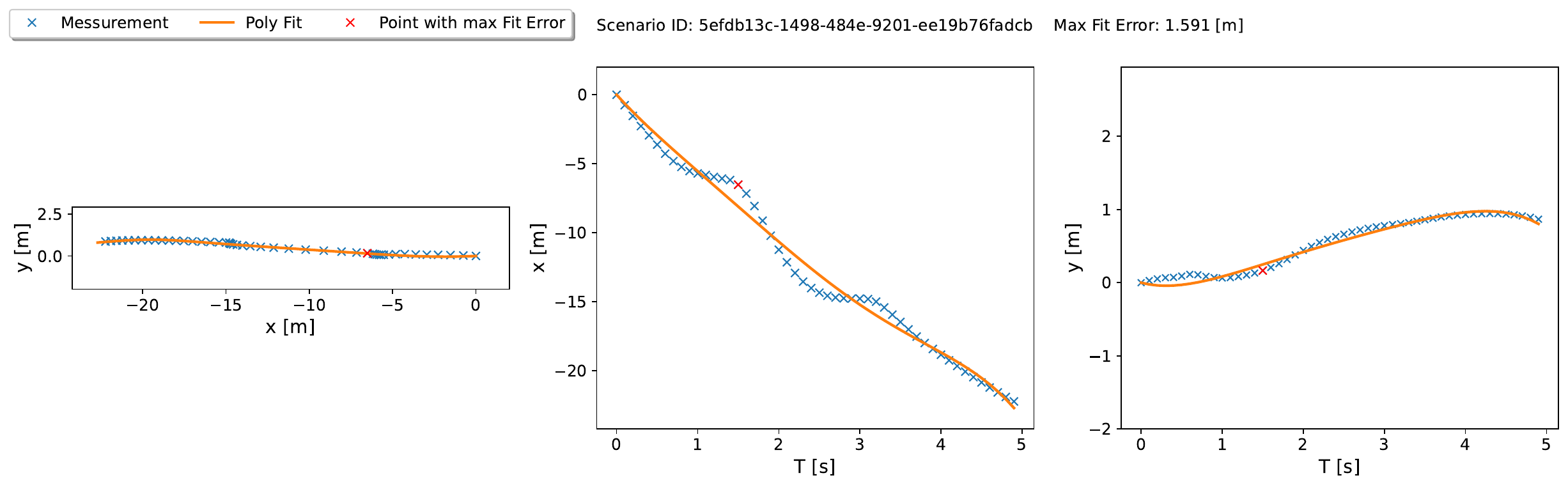} 

\includegraphics[width=5.5in, height=1.7in]{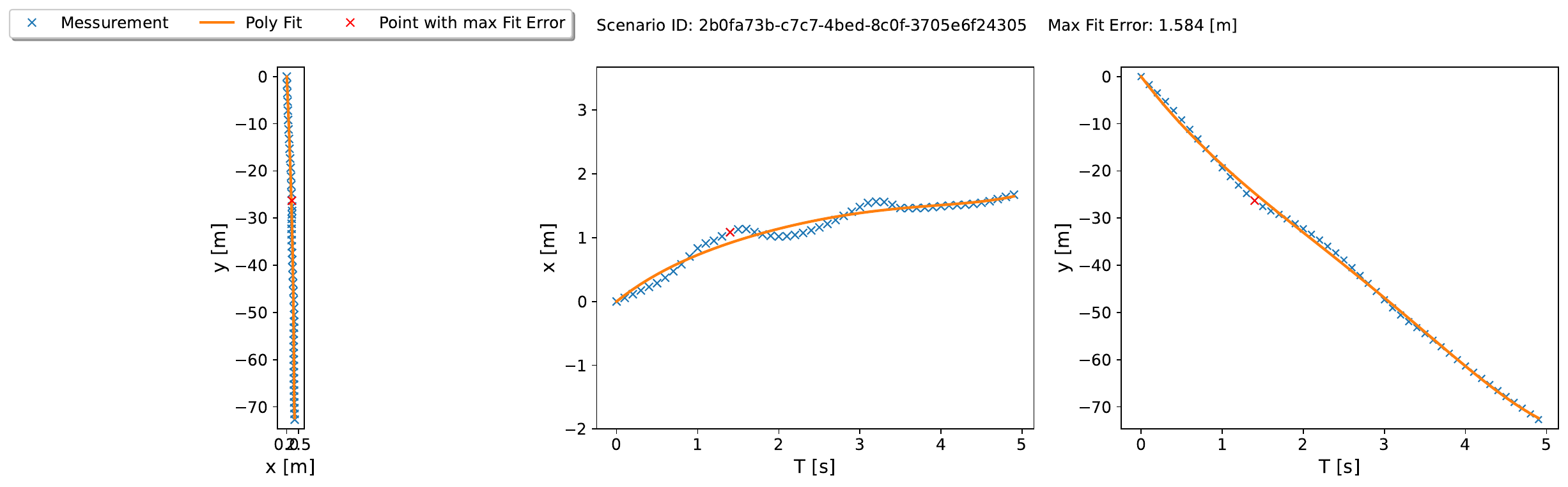} 

\section{10 random 5-seconds vehicle trajectories in A2 (Fitted with $\hat{n} = 5$)}
\centering
\includegraphics[width=5.5in, height=1.8in]{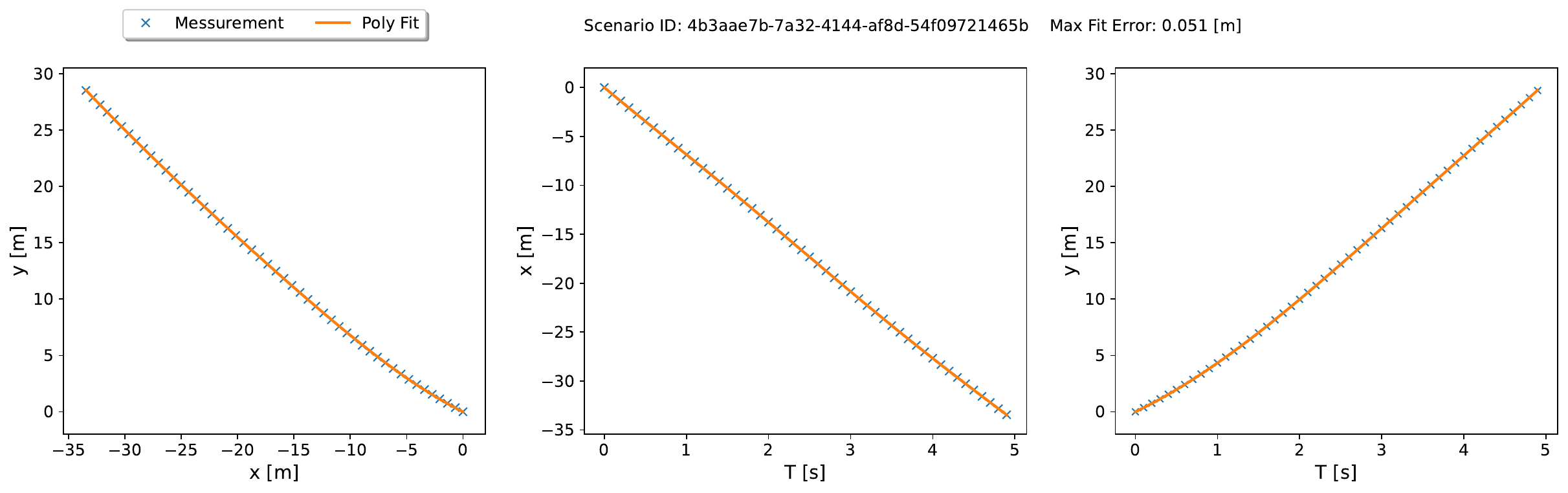} 

\includegraphics[width=5.5in, height=1.8in]{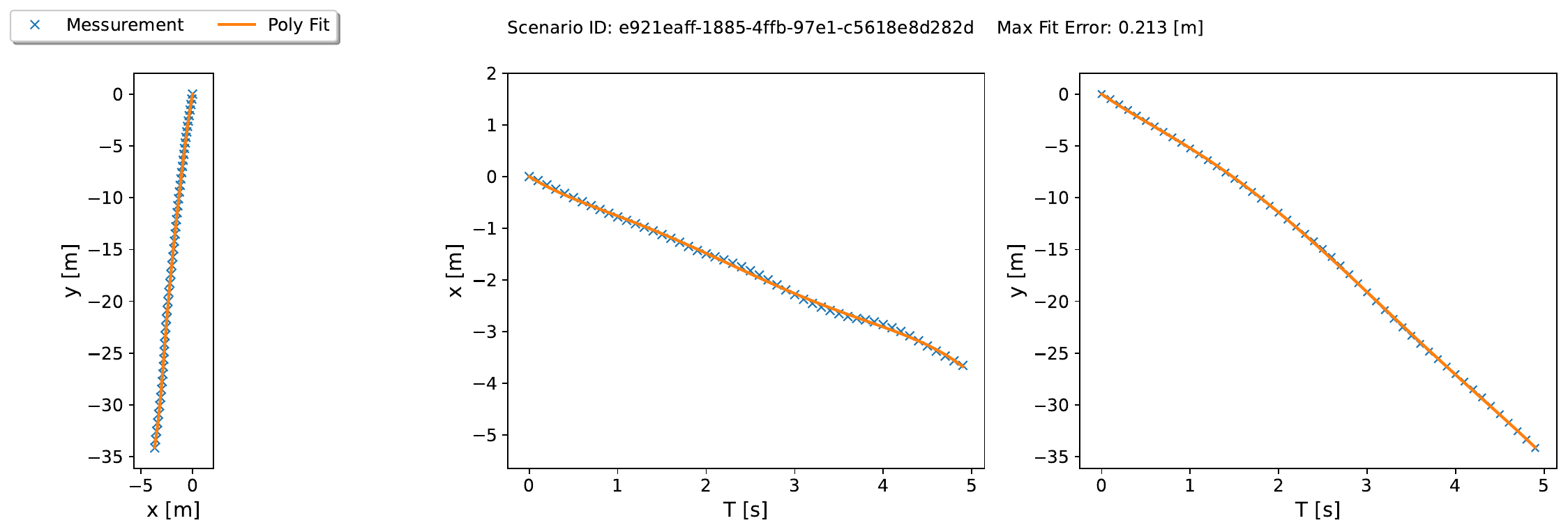} 

\includegraphics[width=5.5in, height=1.8in]{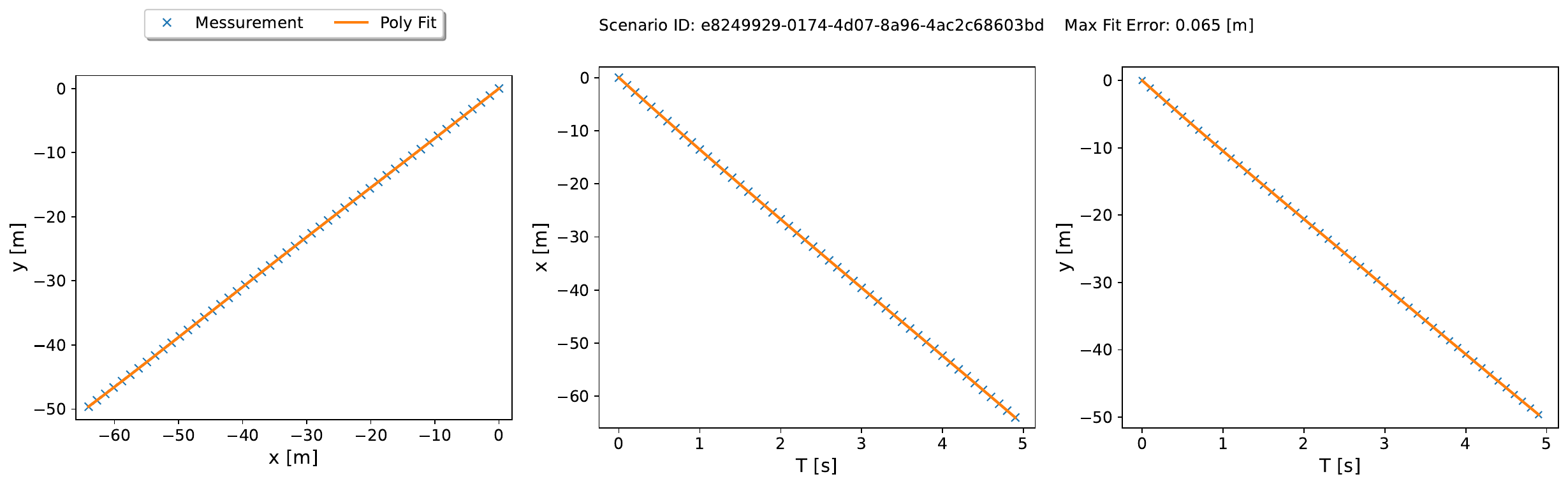} 

\includegraphics[width=5.5in, height=1.8in]{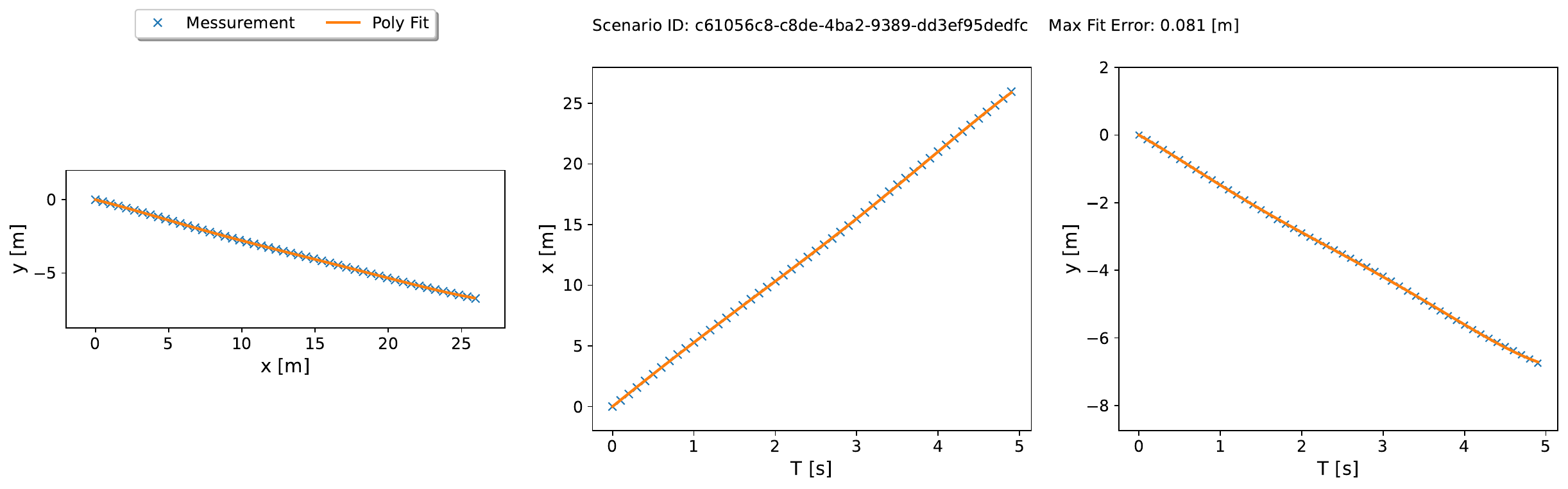} 

\includegraphics[width=5.5in, height=1.7in]{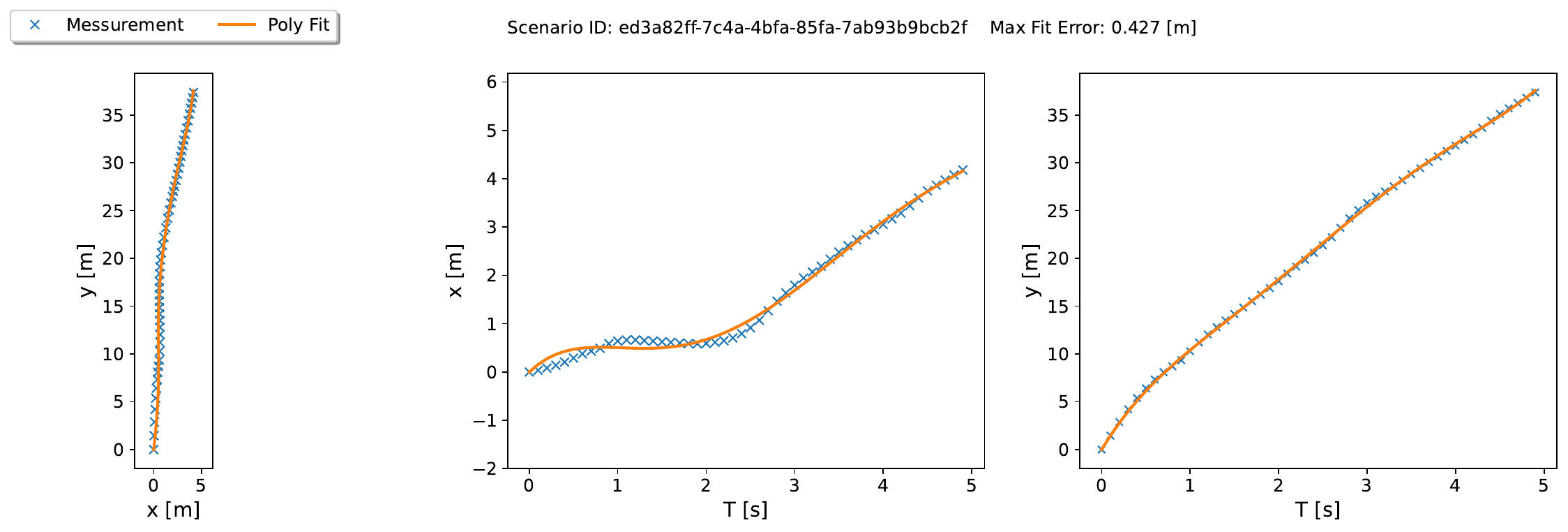} 

\includegraphics[width=5.5in, height=1.8in]{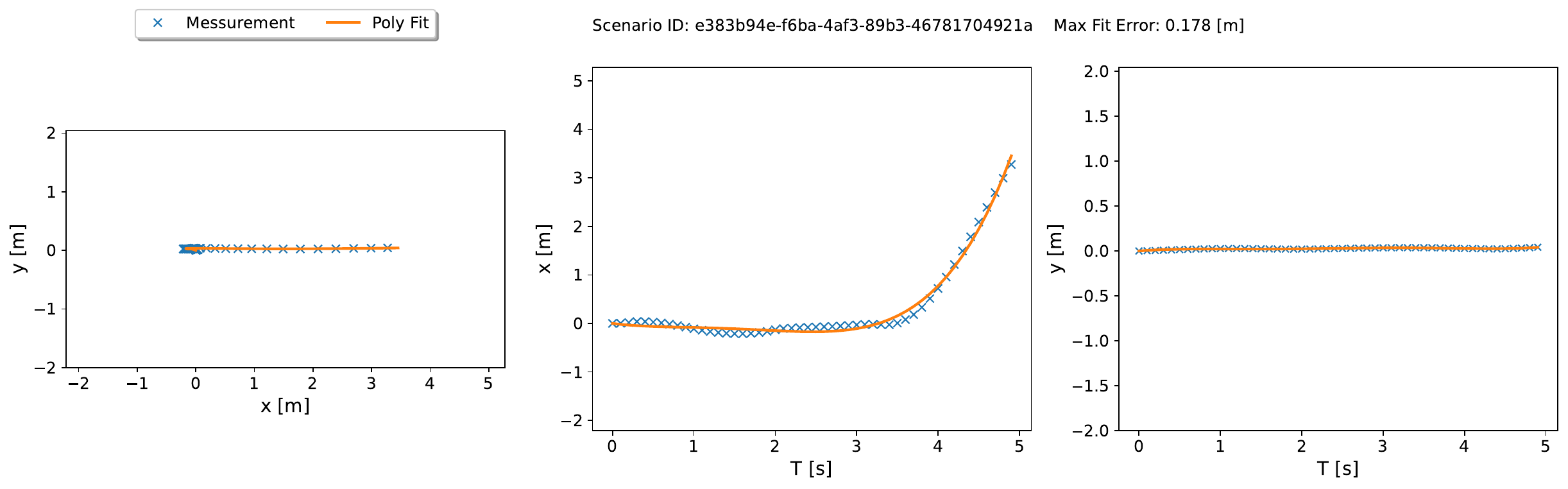} 

\includegraphics[width=5.5in, height=1.8in]{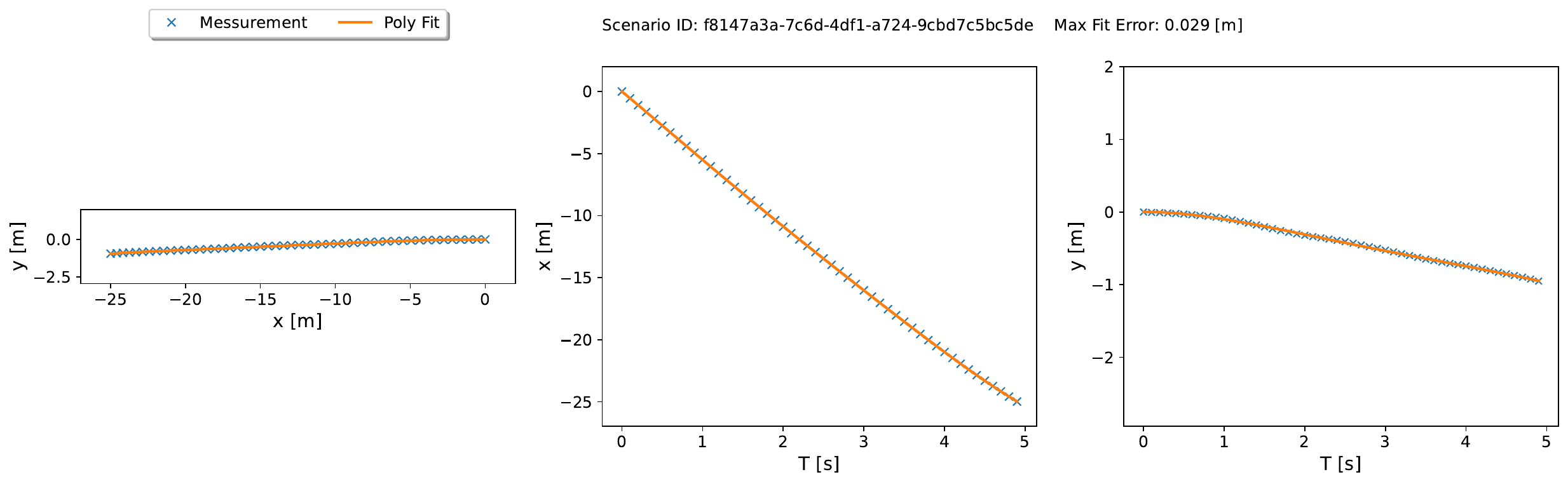} 

\includegraphics[width=5.5in, height=1.8in]{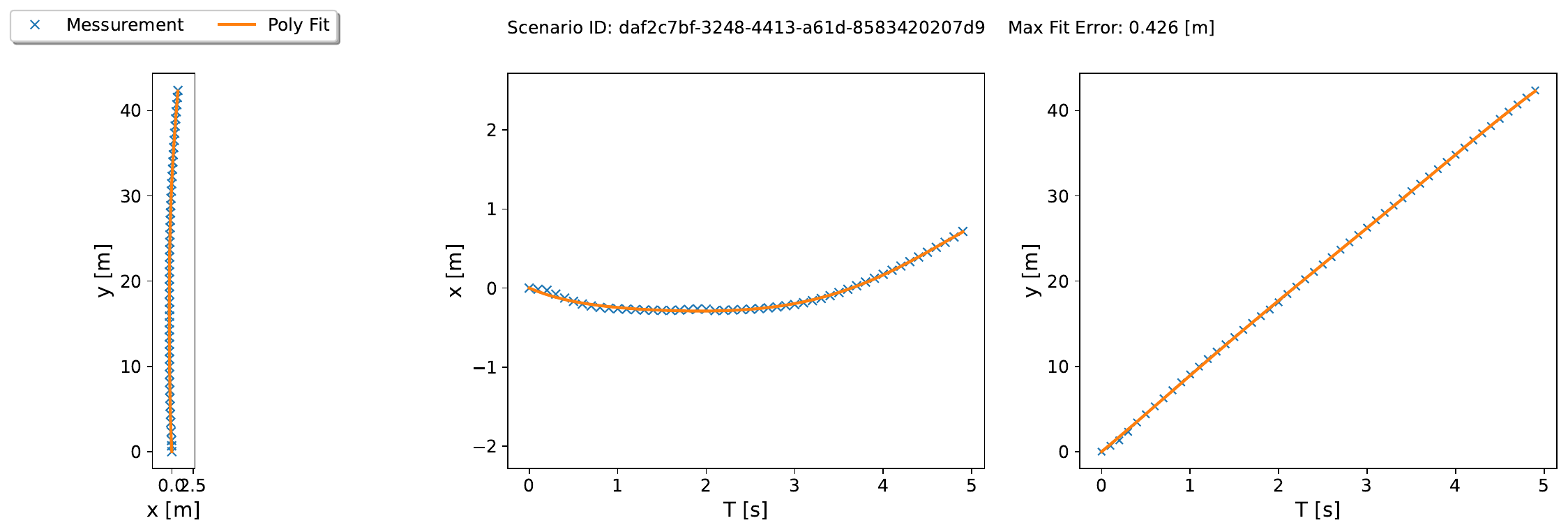} 

\includegraphics[width=5.5in, height=1.8in]{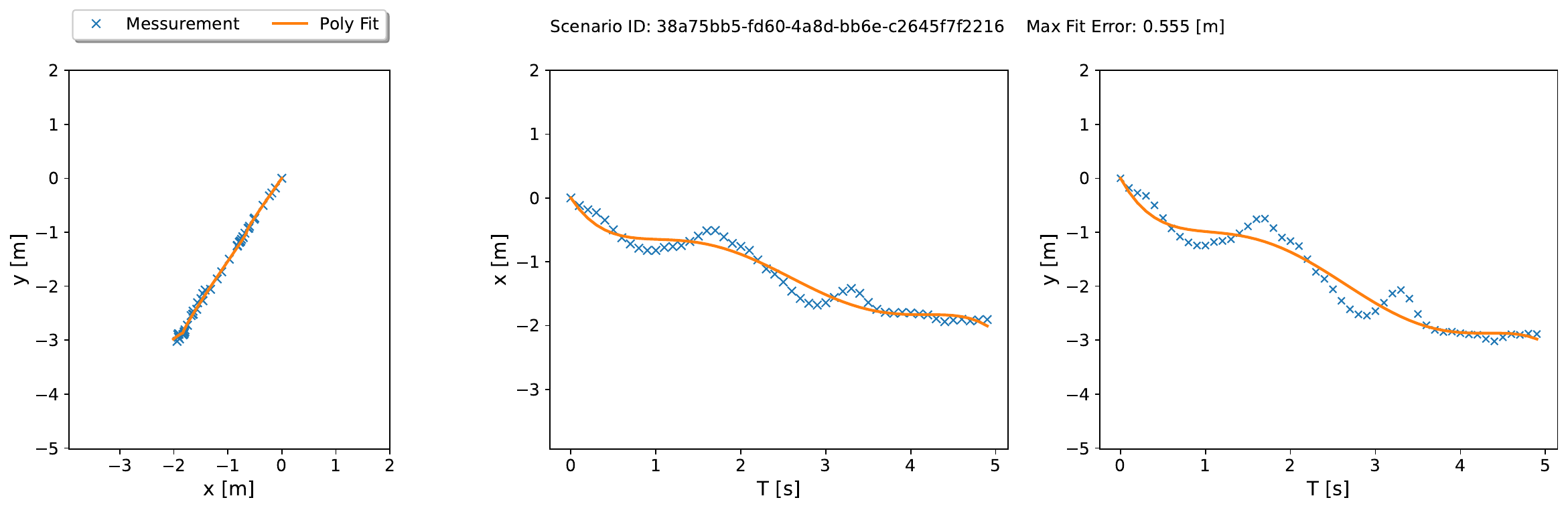} 

\includegraphics[width=5.5in, height=1.7in]{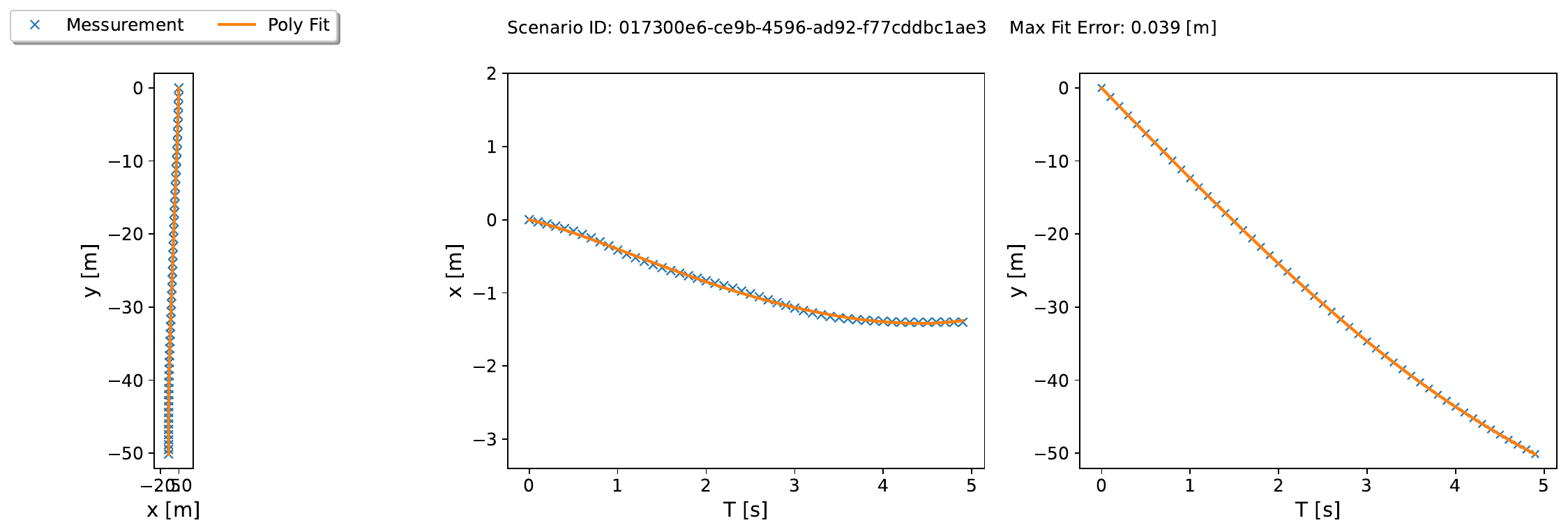} 

\section{10 5-seconds cyclist trajectories with highest fit error in A2 (Fitted with $\hat{n} = 5$)}
\centering
\includegraphics[width=5.5in, height=1.8in]{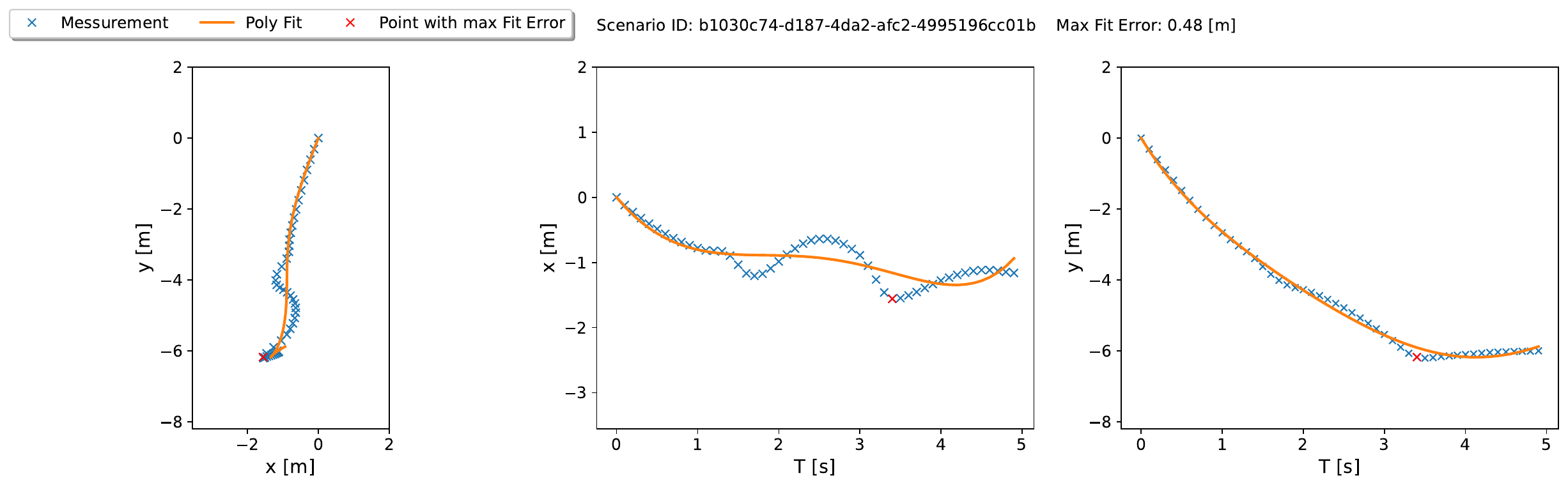} 

\includegraphics[width=5.5in, height=1.8in]{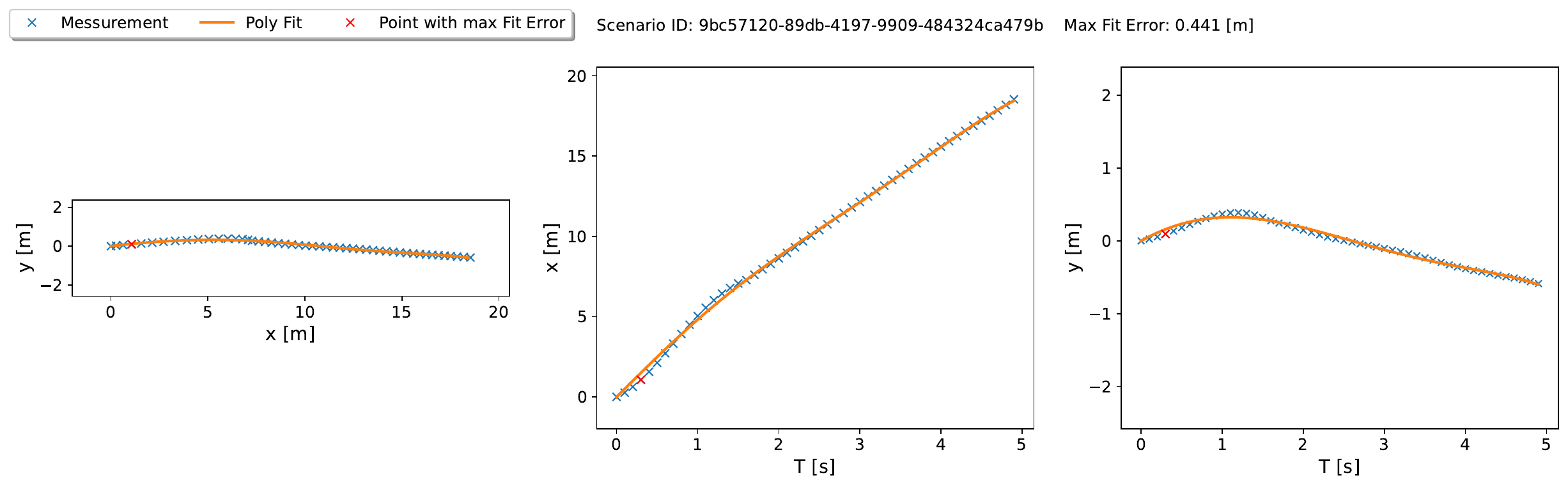} 

\includegraphics[width=5.5in, height=1.8in]{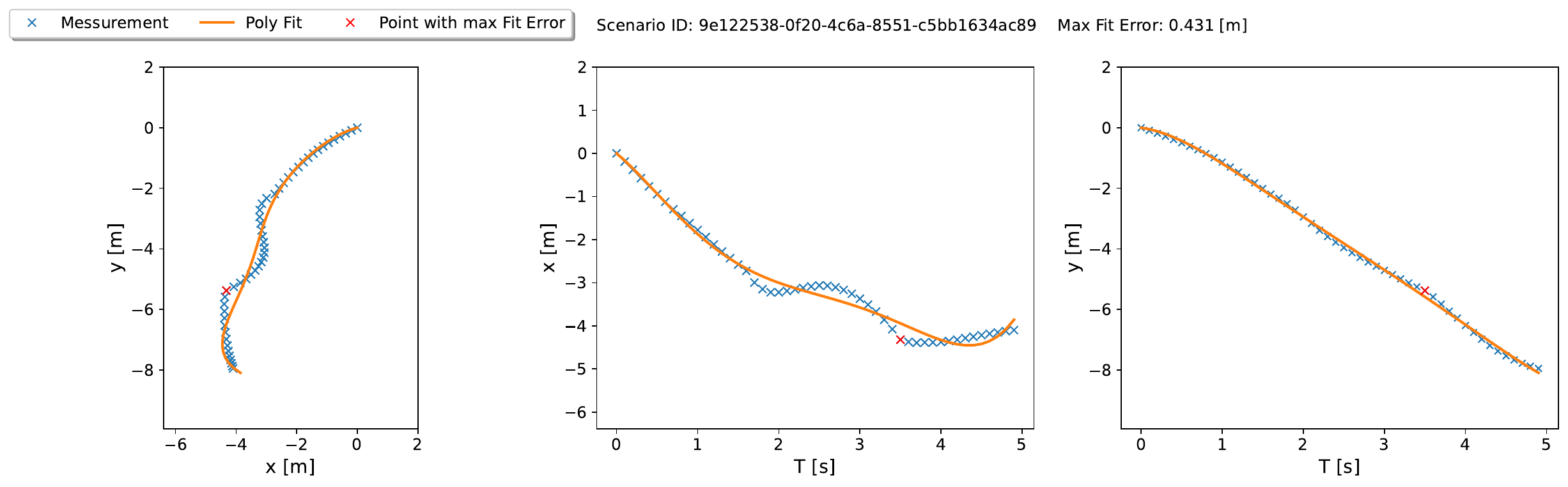} 

\includegraphics[width=5.5in, height=1.8in]{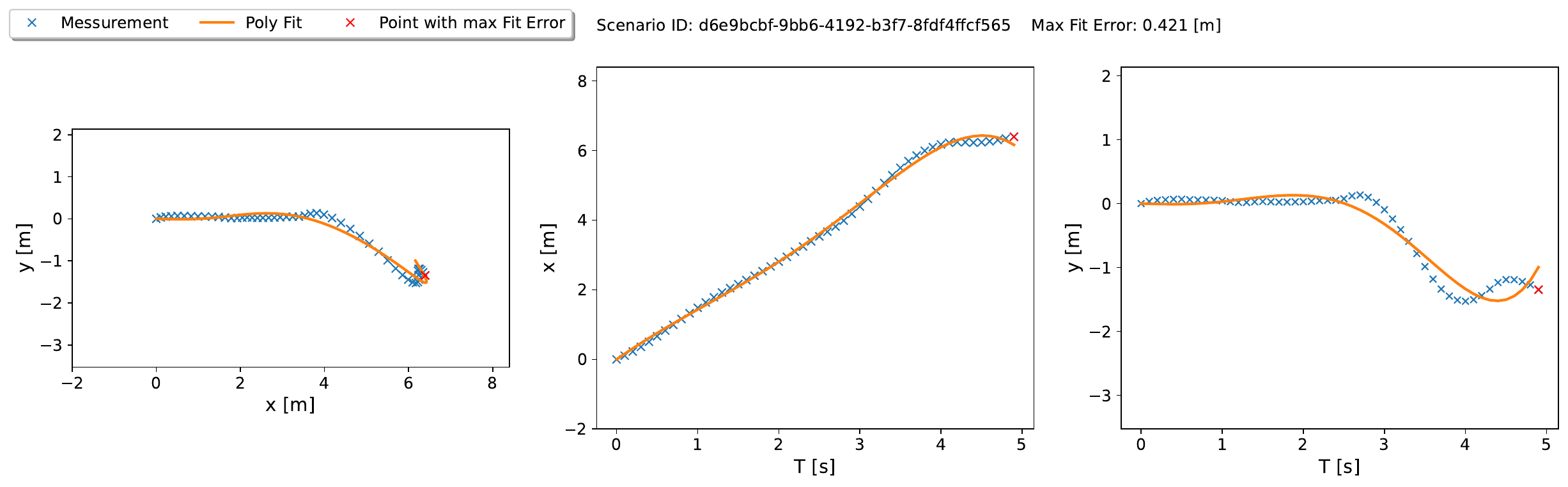} 

\includegraphics[width=5.5in, height=1.7in]{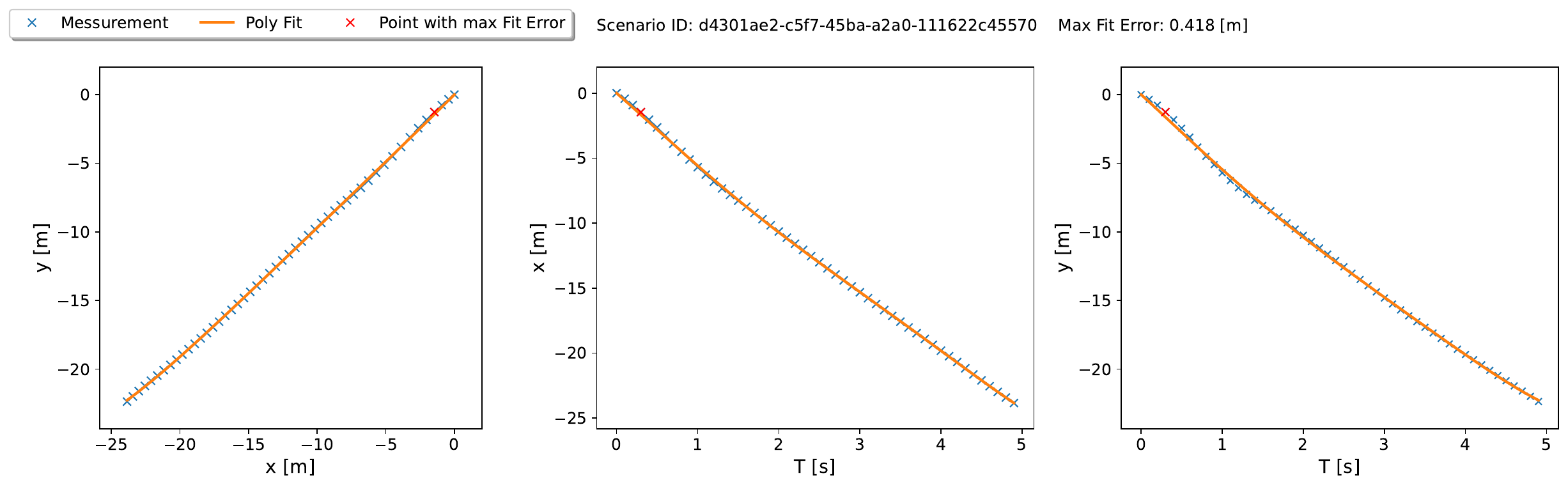} 

\includegraphics[width=5.5in, height=1.8in]{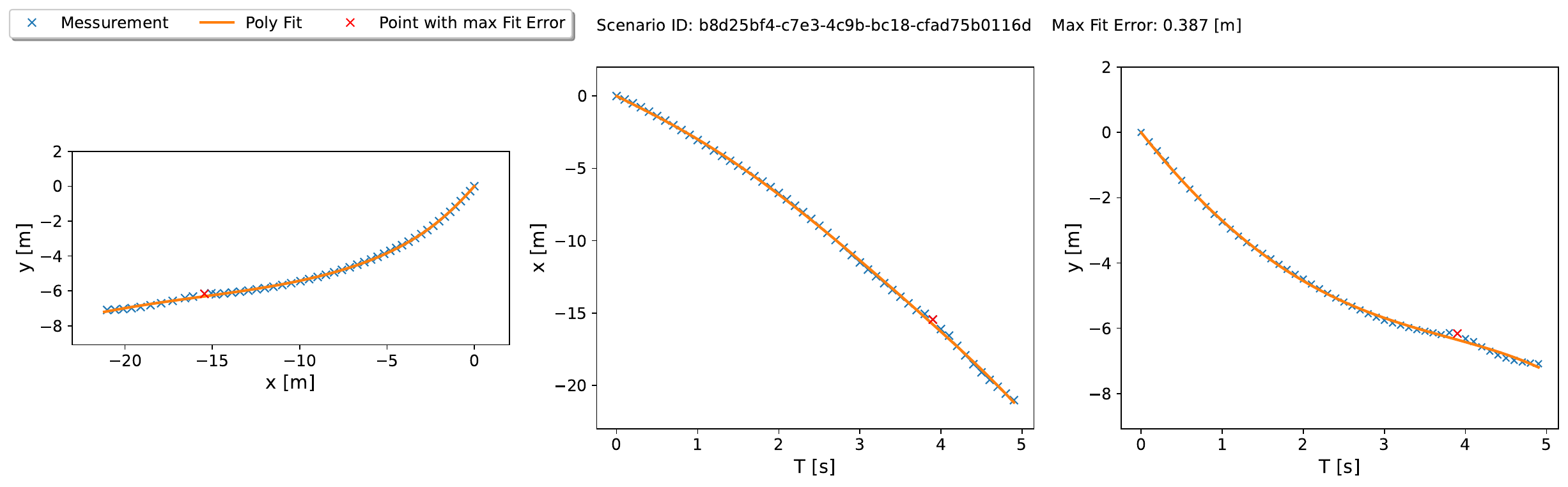} 

\includegraphics[width=5.5in, height=1.8in]{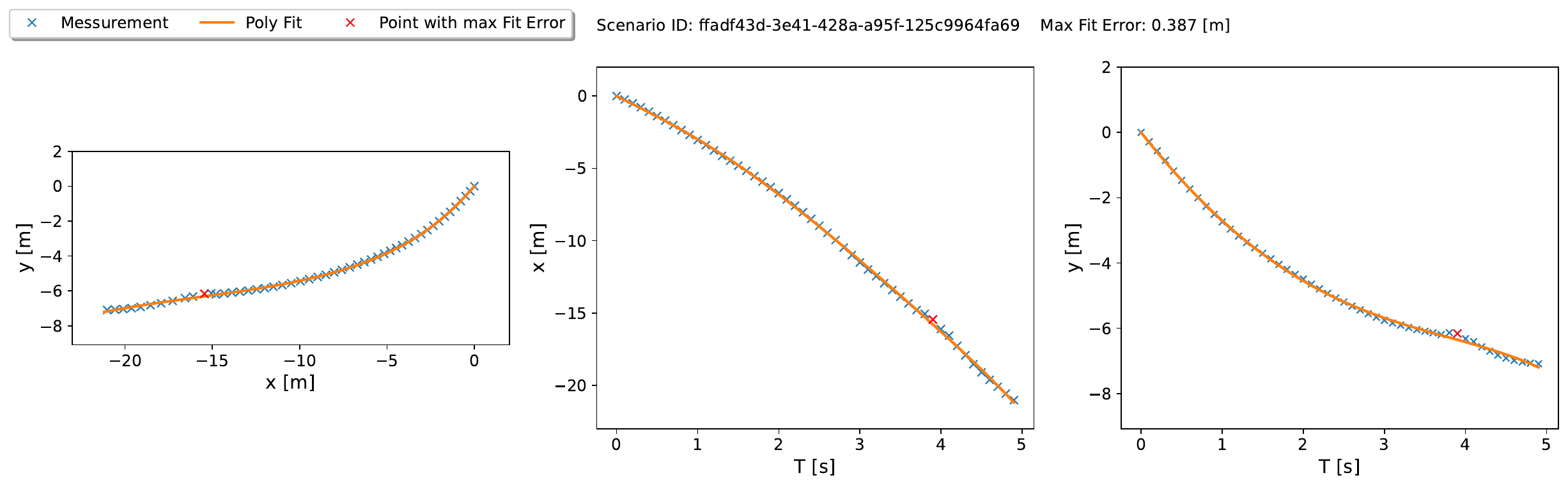} 

\includegraphics[width=5.5in, height=1.8in]{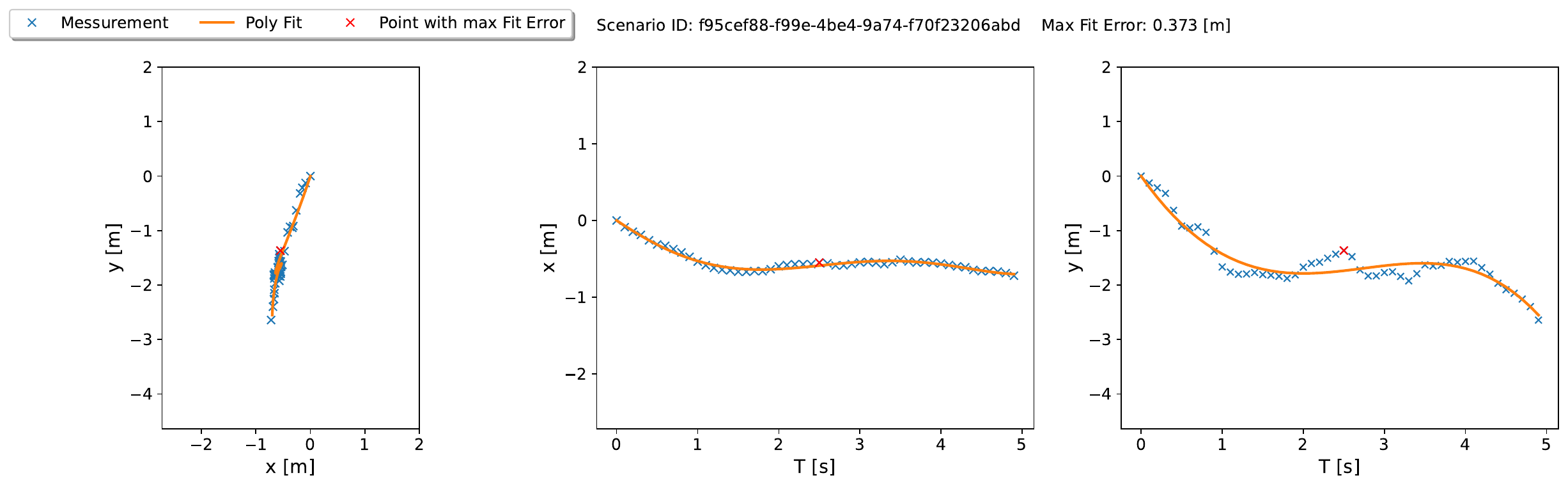} 

\includegraphics[width=5.5in, height=1.8in]{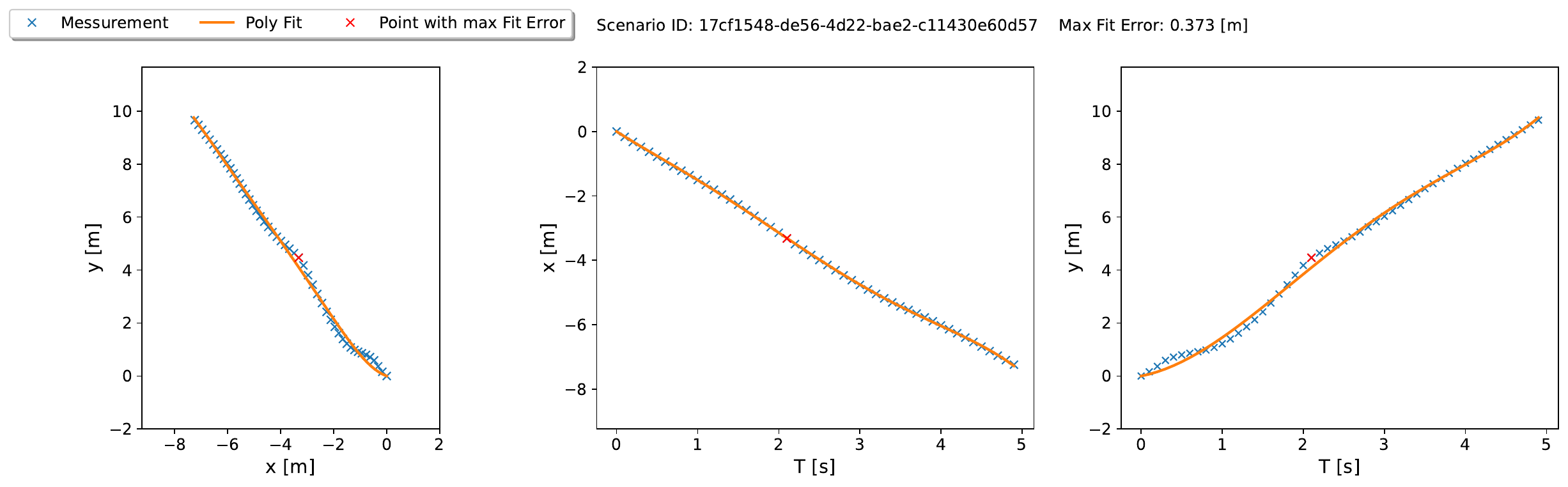} 

\includegraphics[width=5.5in, height=1.7in]{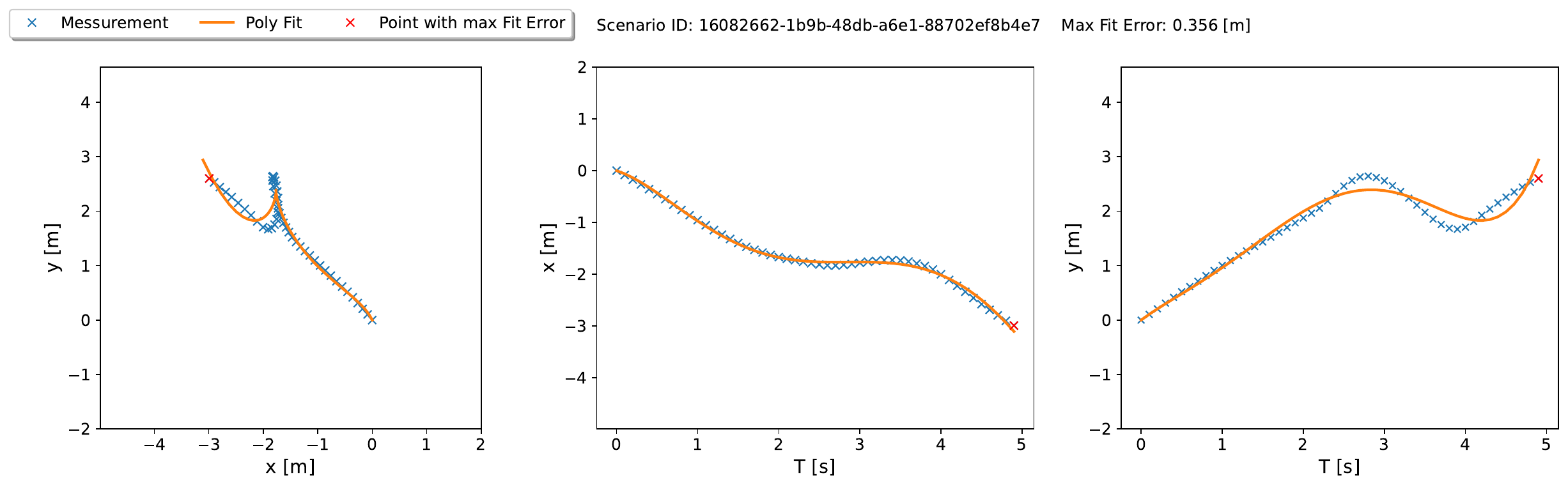} 

\section{10 random 5-seconds cyclist trajectories in A2 (Fitted with $\hat{n} = 5$)}
\centering
\includegraphics[width=5.5in, height=1.8in]{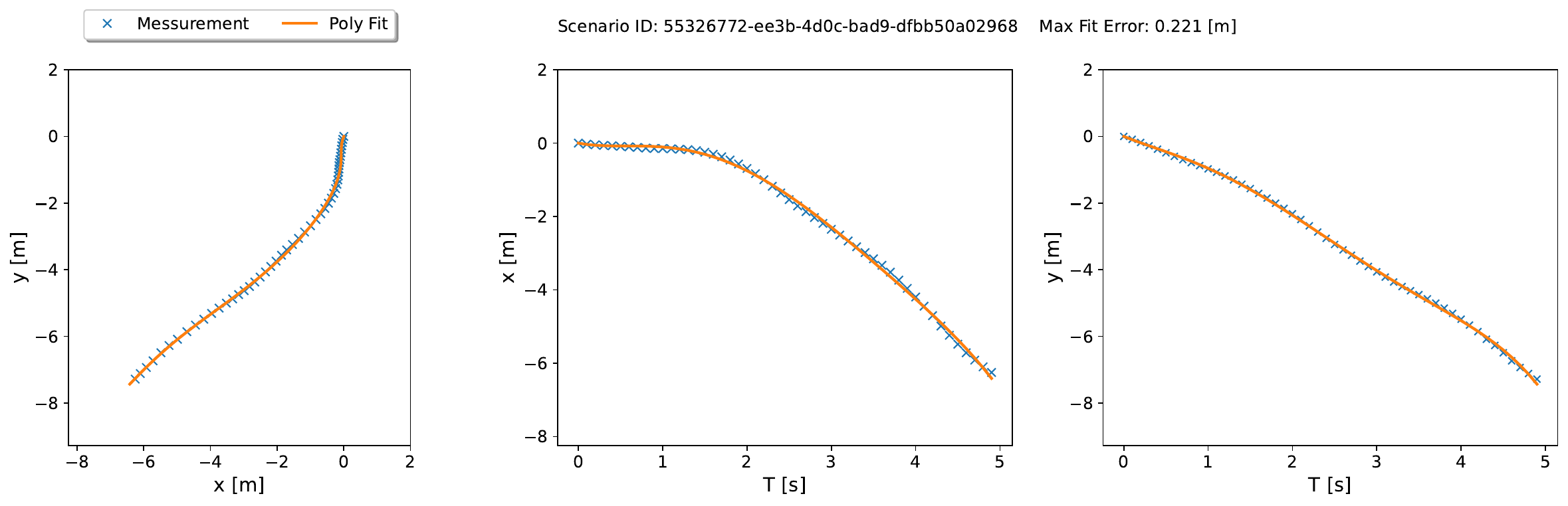} 

\includegraphics[width=5.5in, height=1.8in]{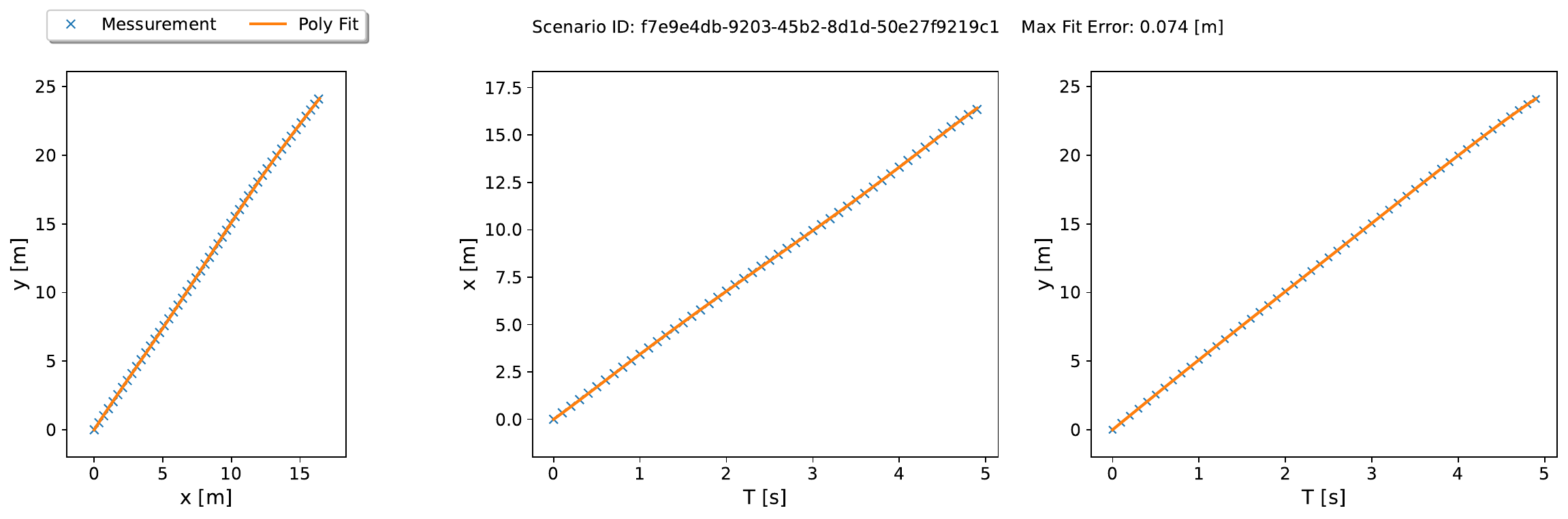} 

\includegraphics[width=5.5in, height=1.8in]{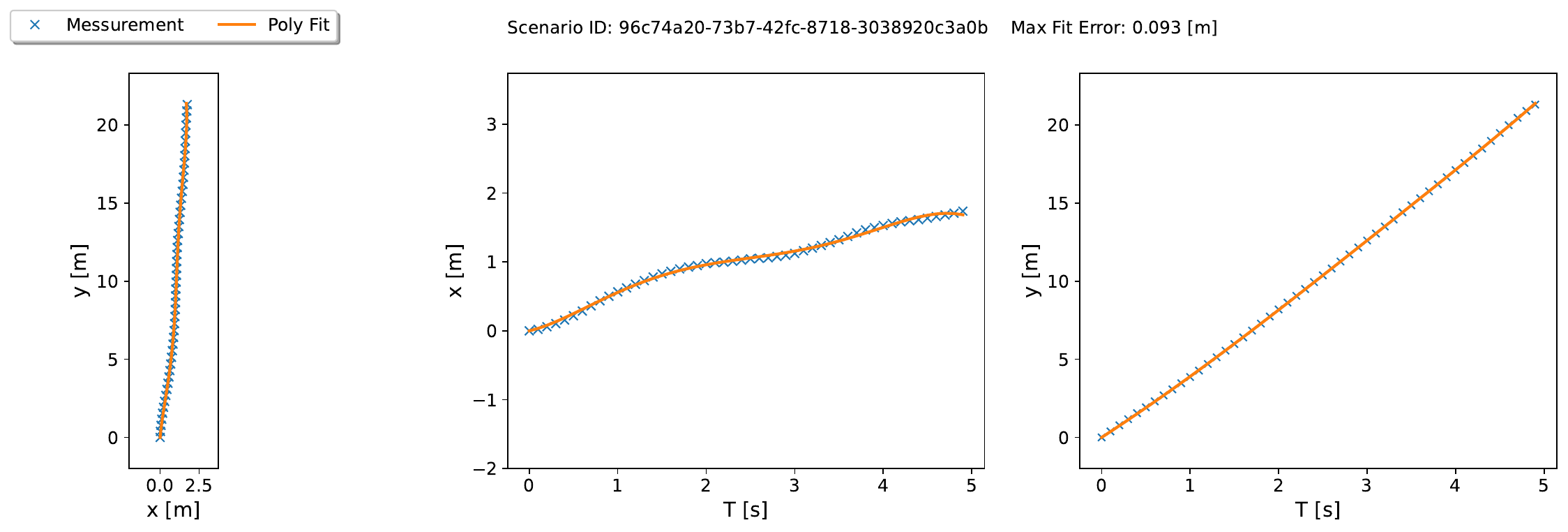} 

\includegraphics[width=5.5in, height=1.8in]{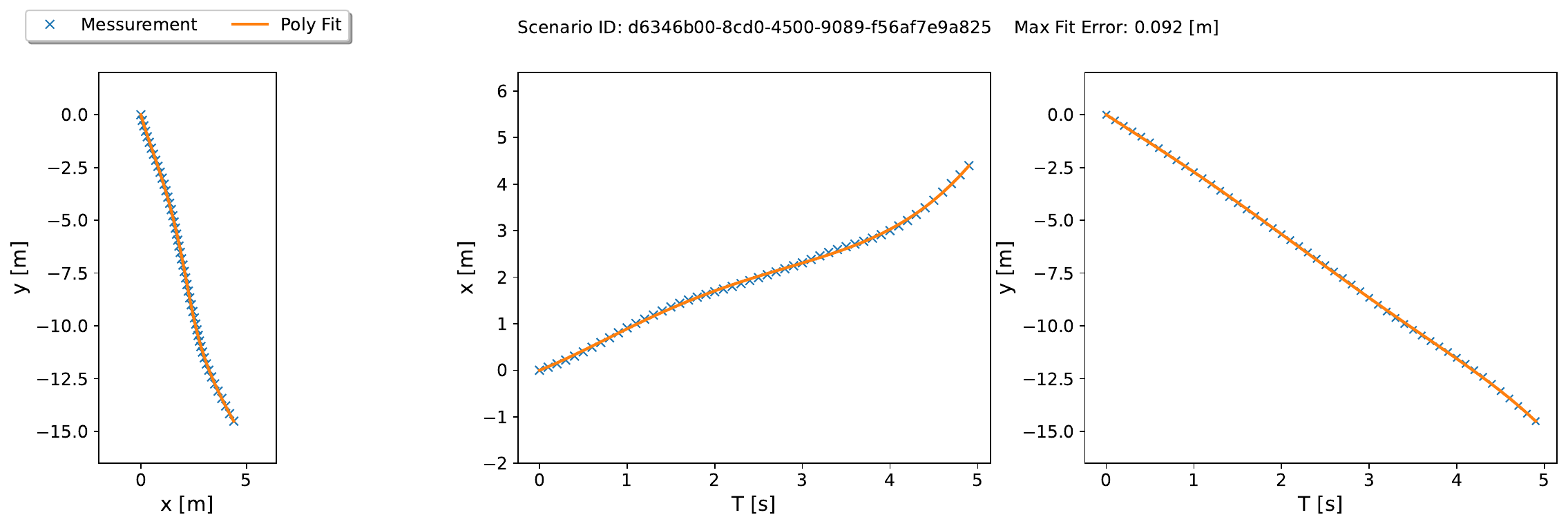} 

\includegraphics[width=5.5in, height=1.7in]{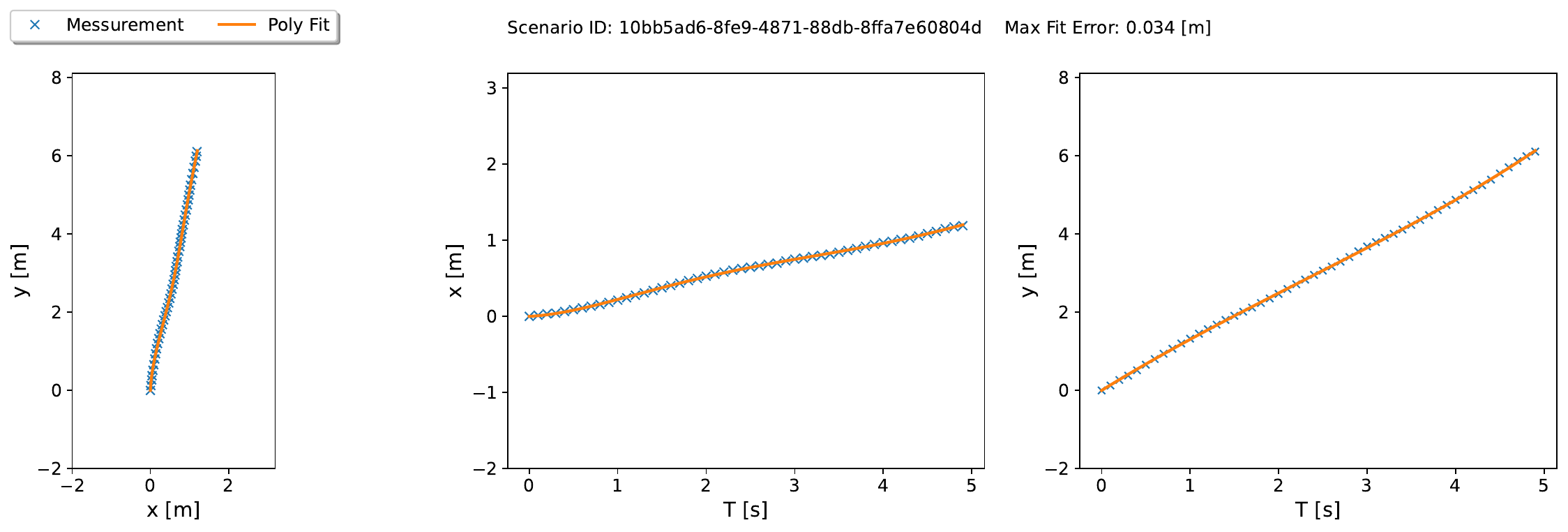} 

\includegraphics[width=5.5in, height=1.8in]{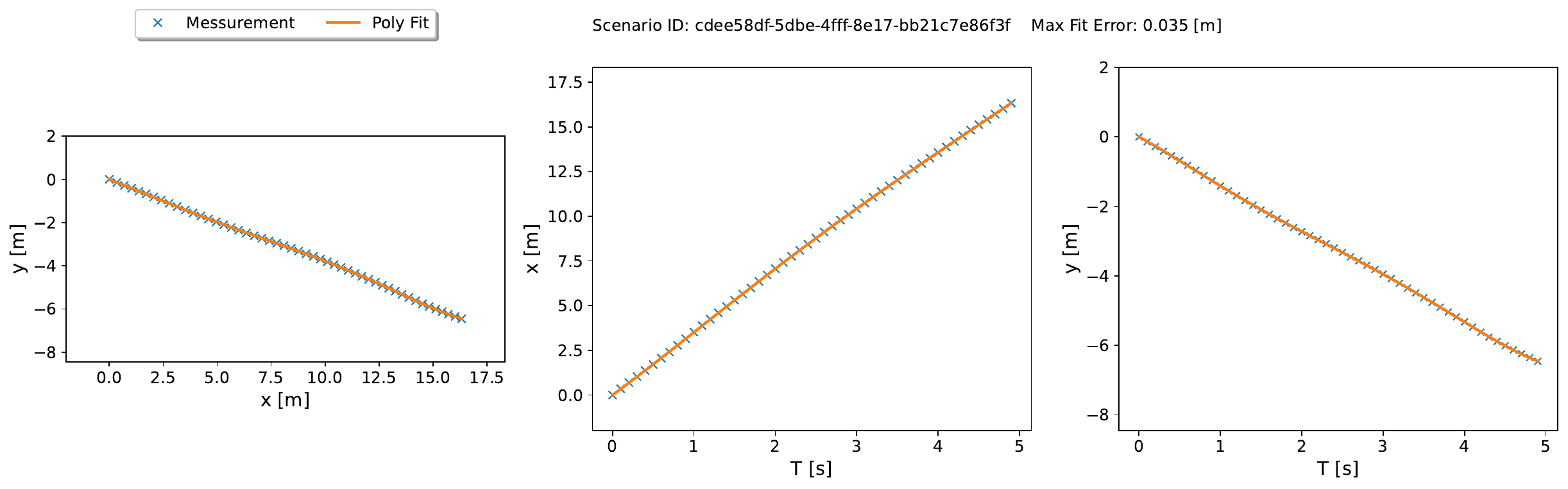} 

\includegraphics[width=5.5in, height=1.8in]{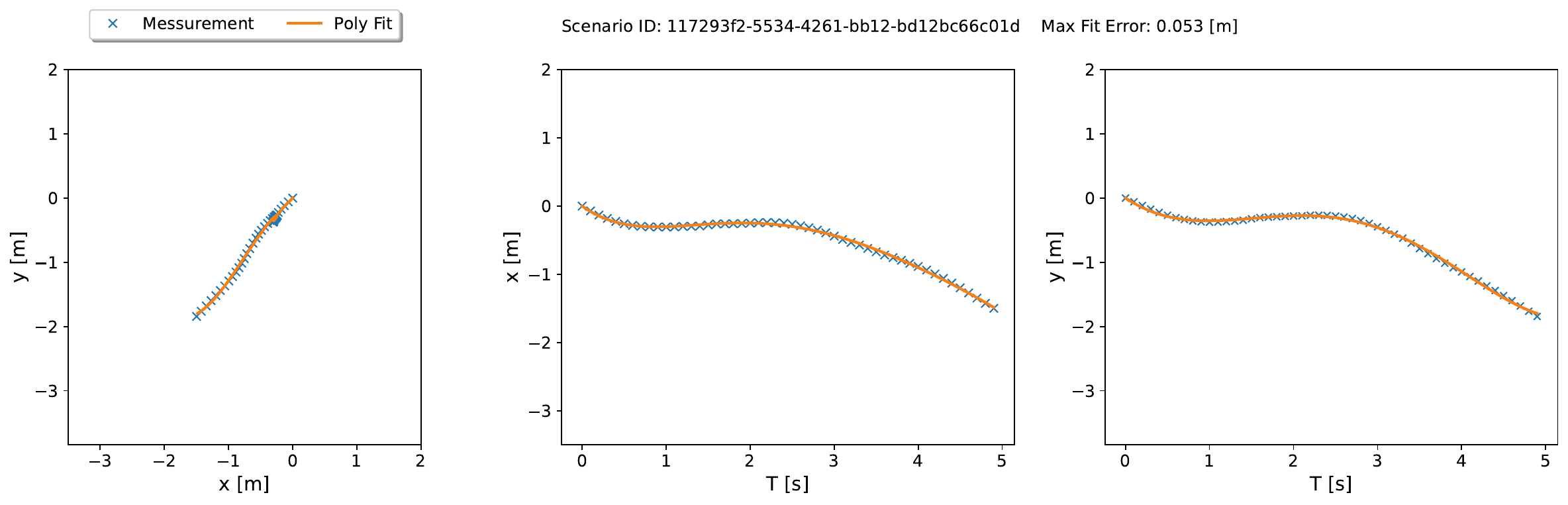} 

\includegraphics[width=5.5in, height=1.8in]{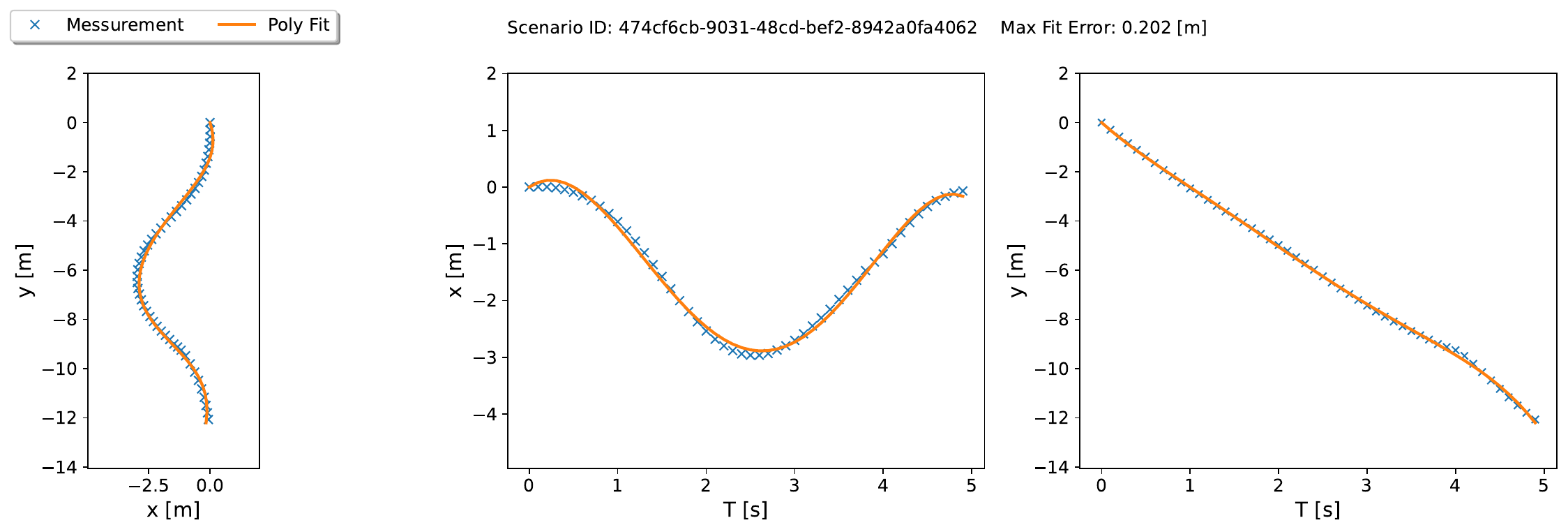} 

\includegraphics[width=5.5in, height=1.8in]{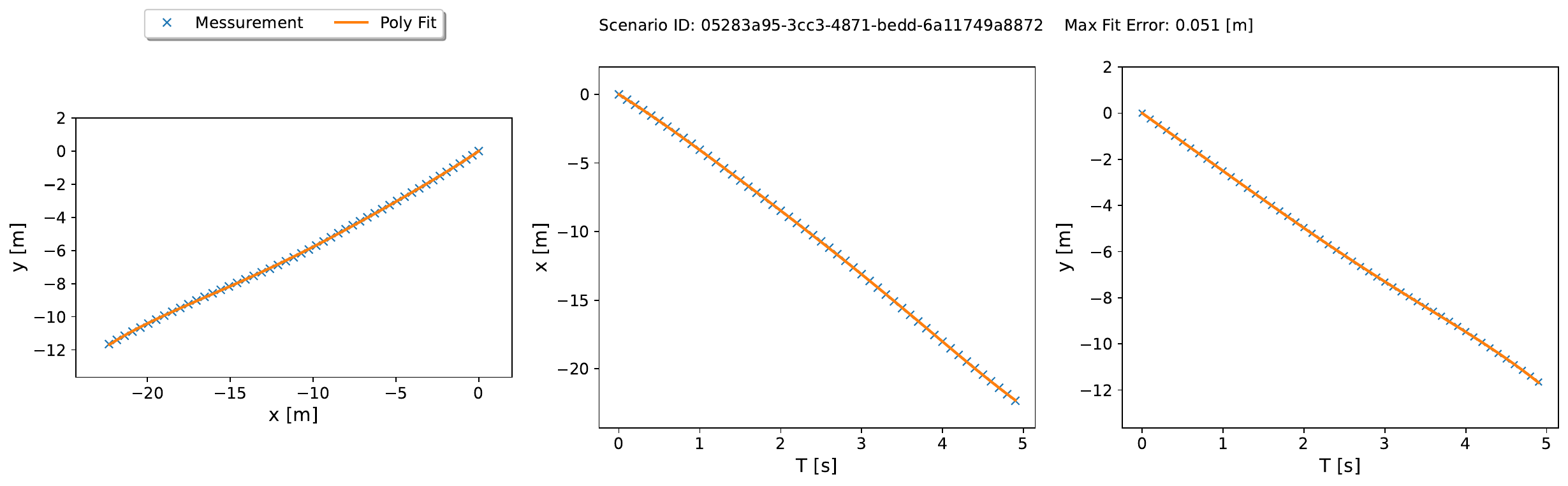} 

\includegraphics[width=5.5in, height=1.7in]{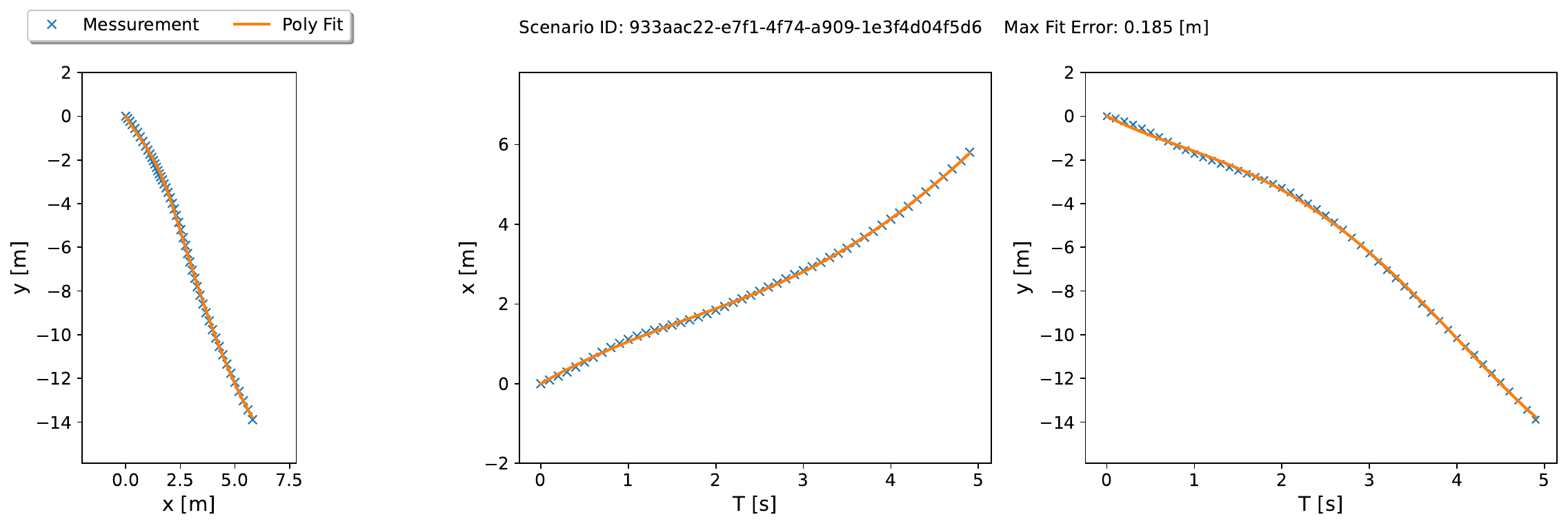} 

\section{10 5-seconds pedestrian trajectories with highest fit error in A2 (Fitted with $\hat{n} = 5$)}
\centering
\includegraphics[width=5.5in, height=1.8in]{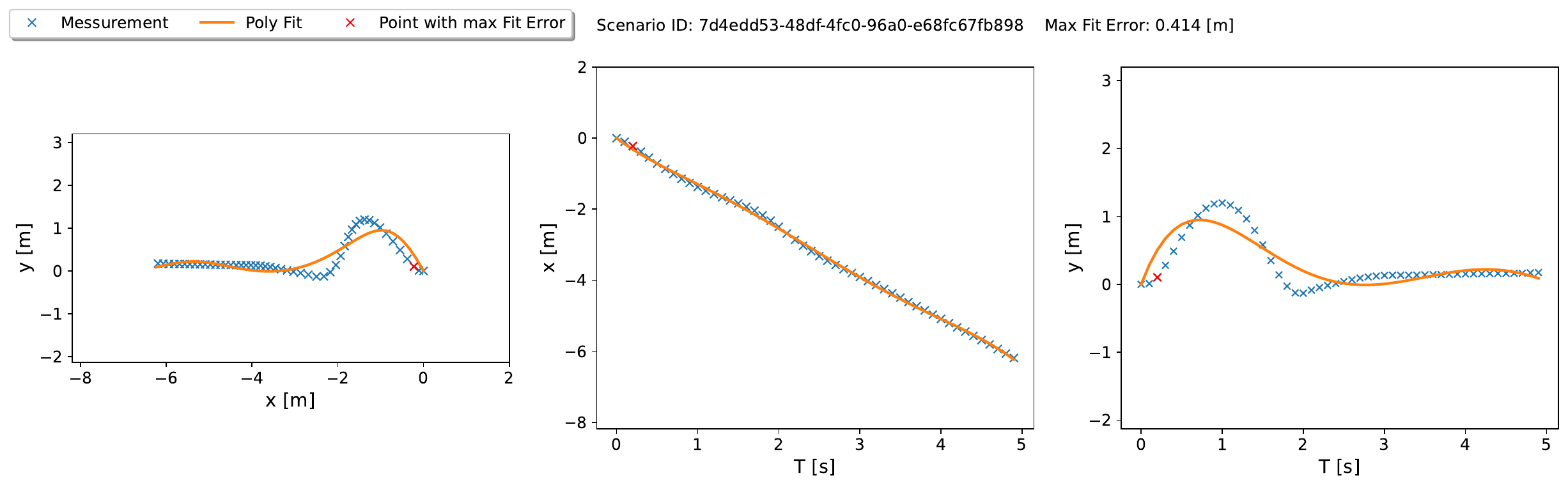} 

\includegraphics[width=5.5in, height=1.8in]{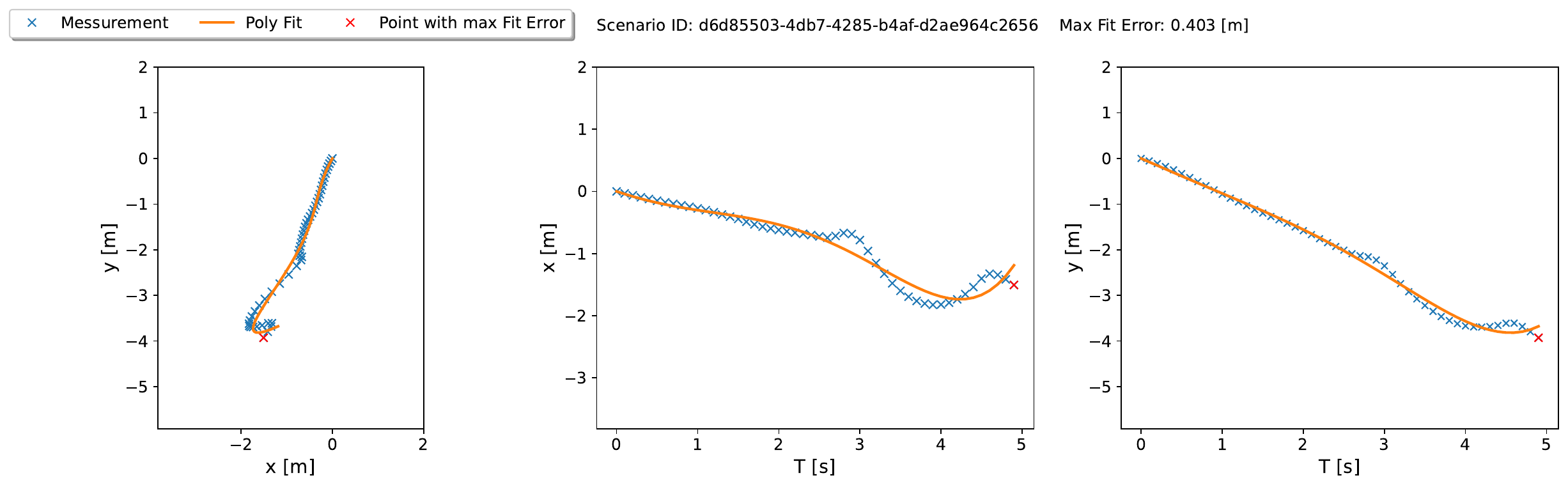} 

\includegraphics[width=5.5in, height=1.8in]{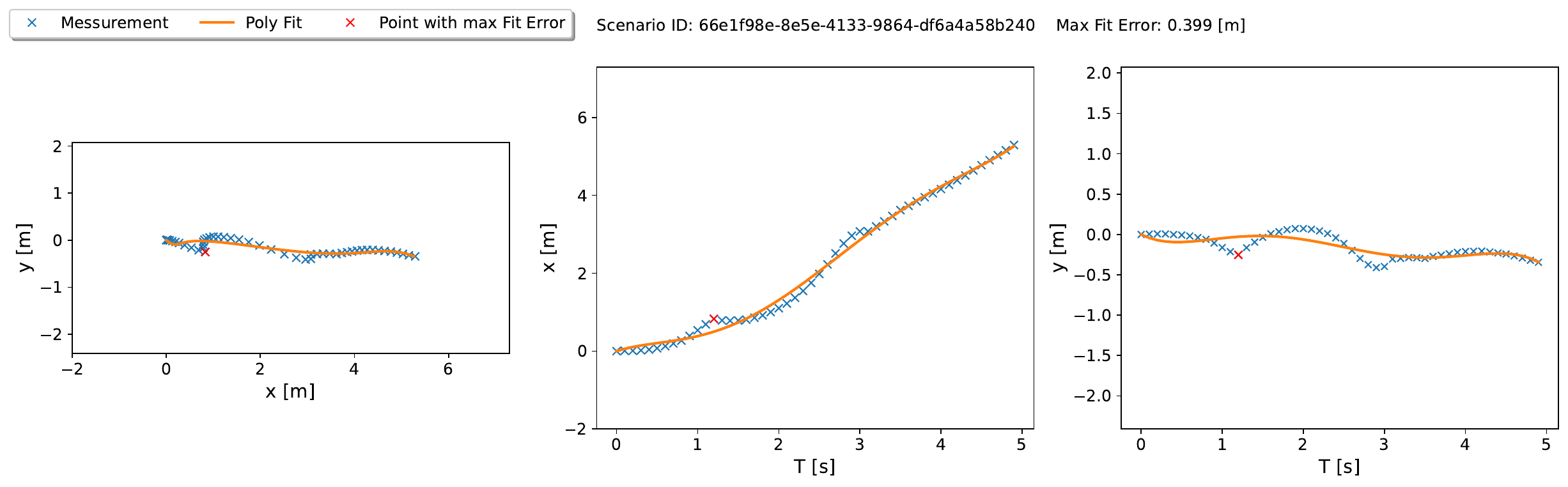} 

\includegraphics[width=5.5in, height=1.8in]{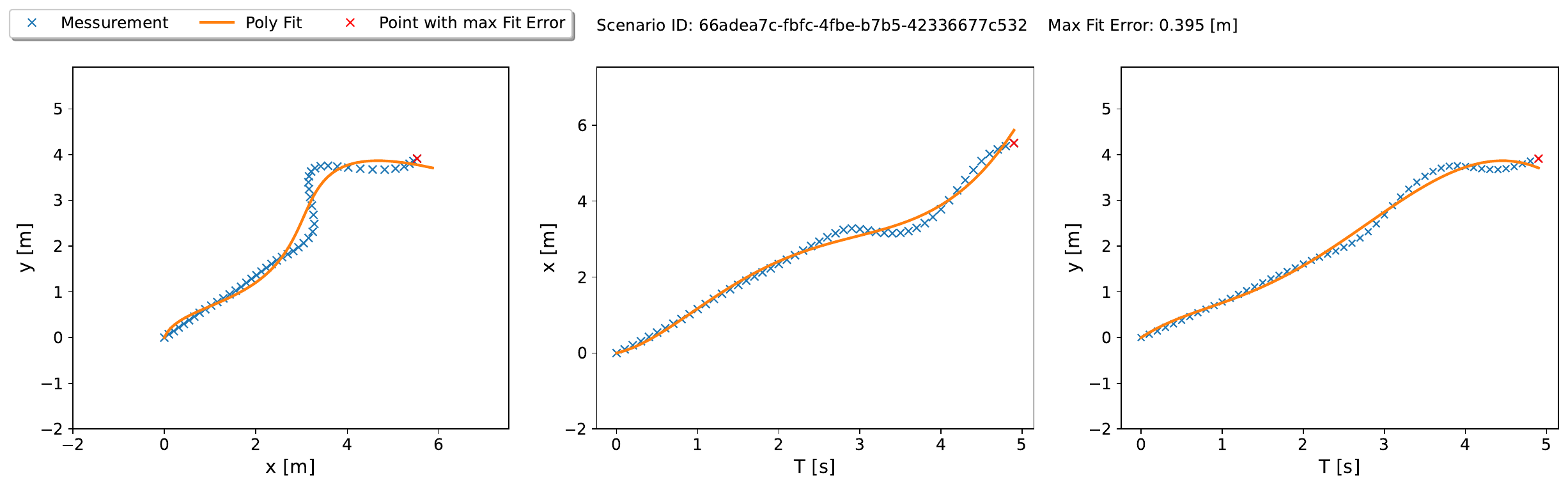} 

\includegraphics[width=5.5in, height=1.7in]{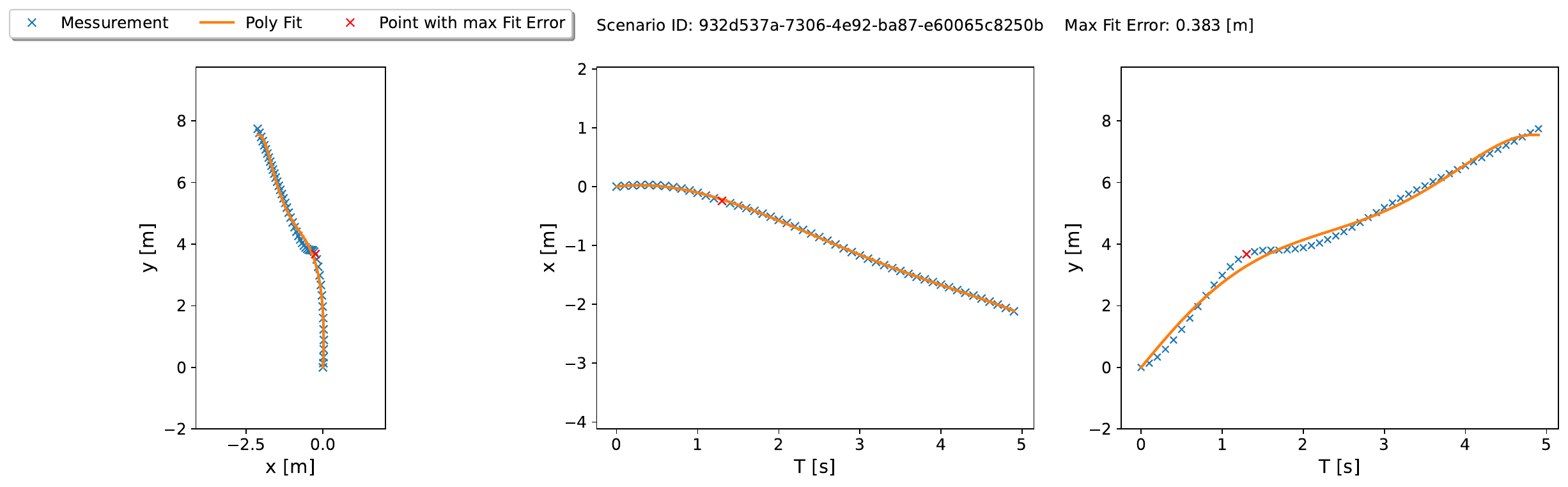} 

\includegraphics[width=5.5in, height=1.8in]{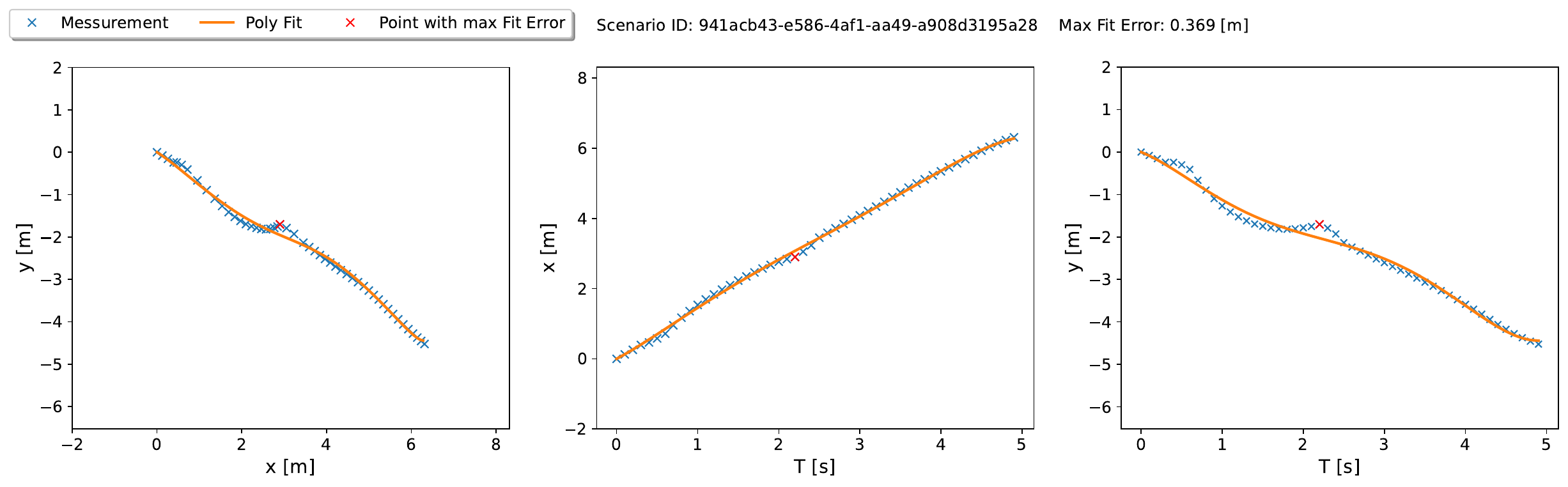} 

\includegraphics[width=5.5in, height=1.8in]{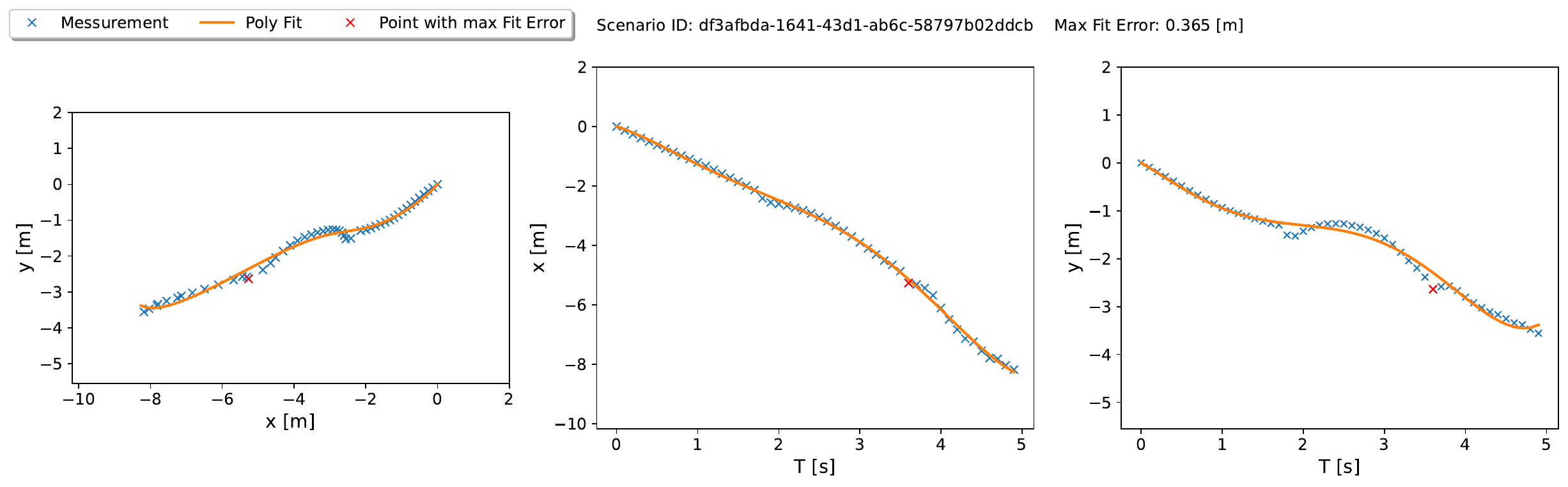} 

\includegraphics[width=5.5in, height=1.8in]{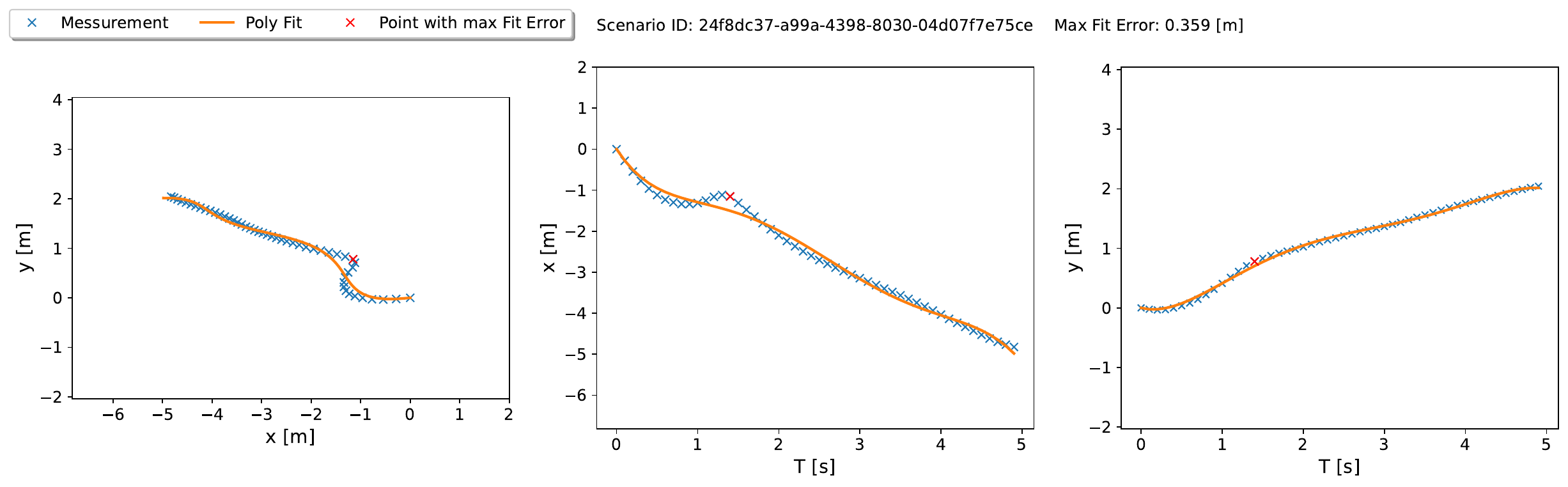} 

\includegraphics[width=5.5in, height=1.8in]{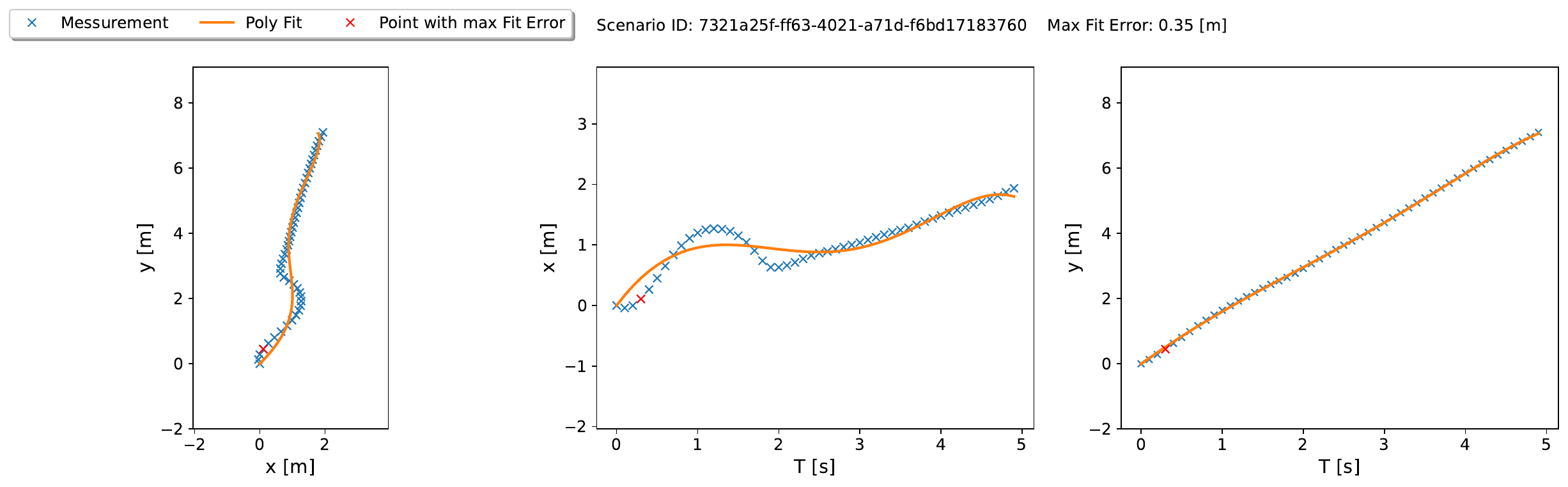} 

\includegraphics[width=5.5in, height=1.7in]{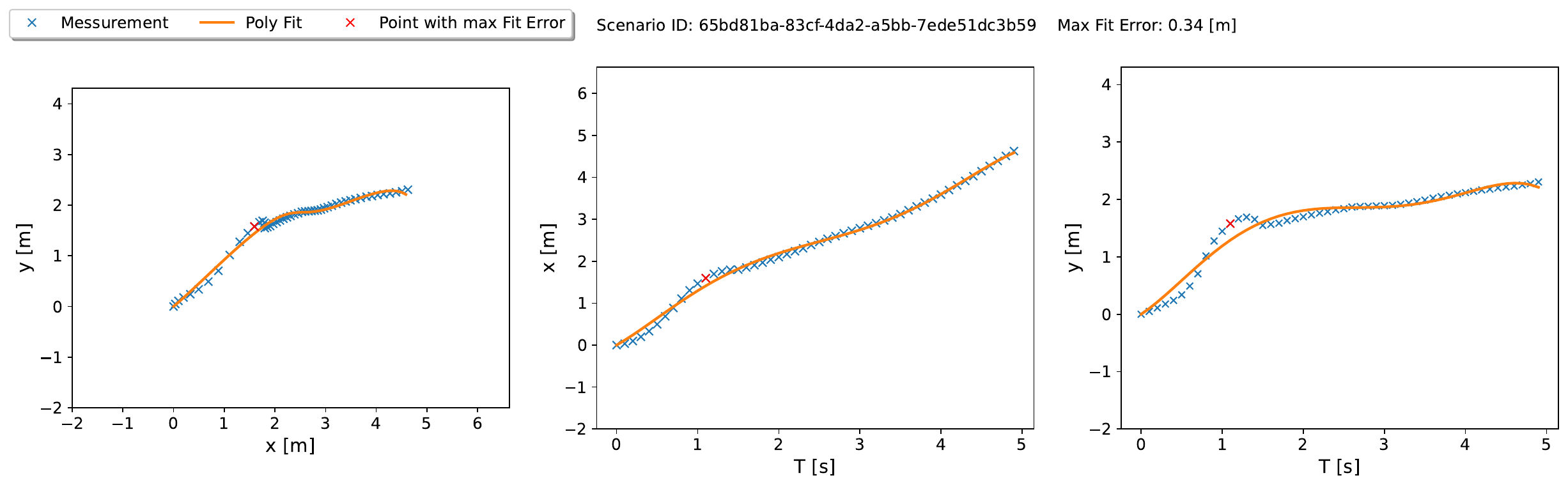} 

\section{10 random 5-seconds pedestrian trajectories in A2 (Fitted with $\hat{n} = 5$)}
\centering
\includegraphics[width=5.5in, height=1.8in]{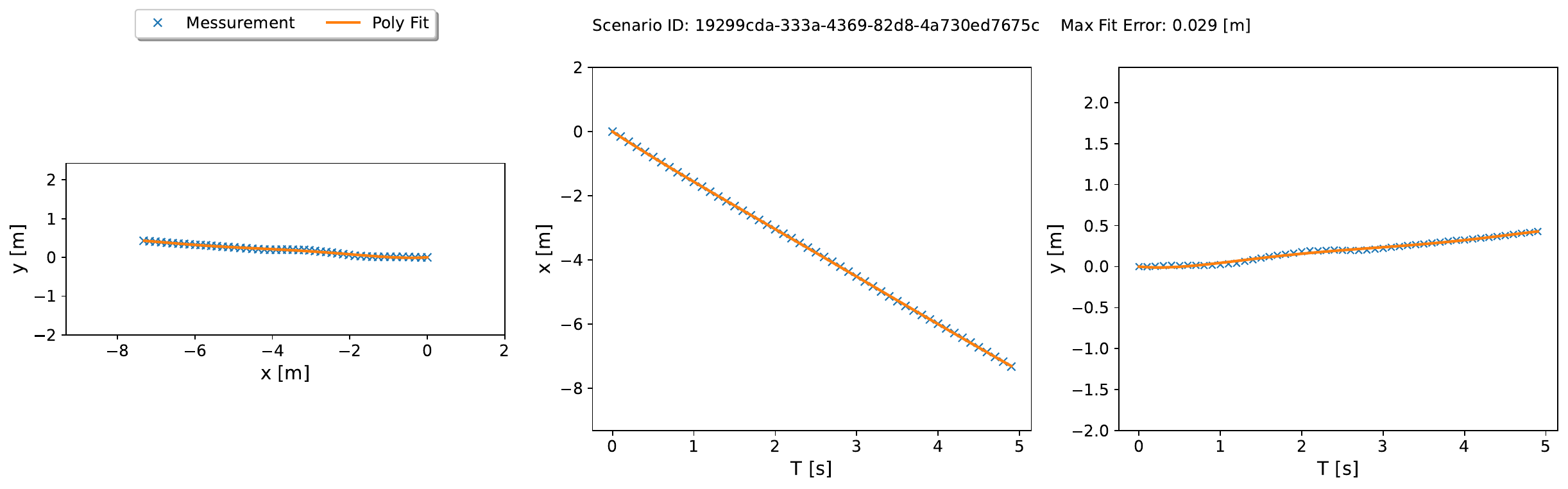} 

\includegraphics[width=5.5in, height=1.8in]{images/appendix_img/A2/pedestrian/50_random_samples_A2_pedestrian_01.pdf} 

\includegraphics[width=5.5in, height=1.8in]{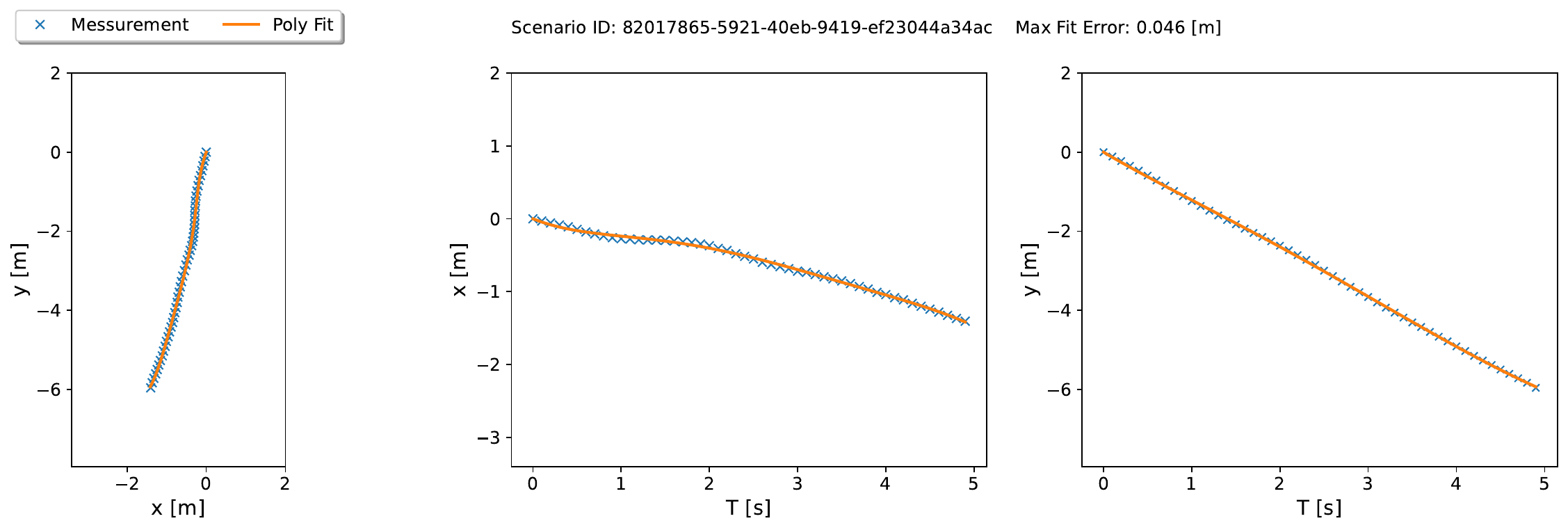} 

\includegraphics[width=5.5in, height=1.8in]{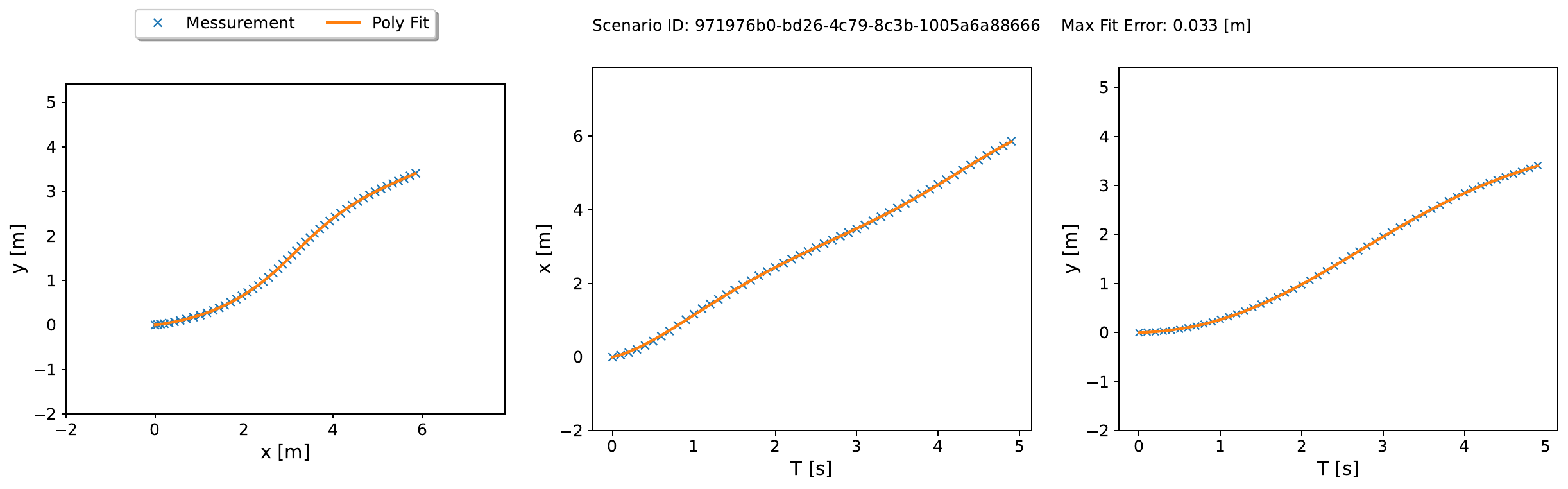} 

\includegraphics[width=5.5in, height=1.7in]{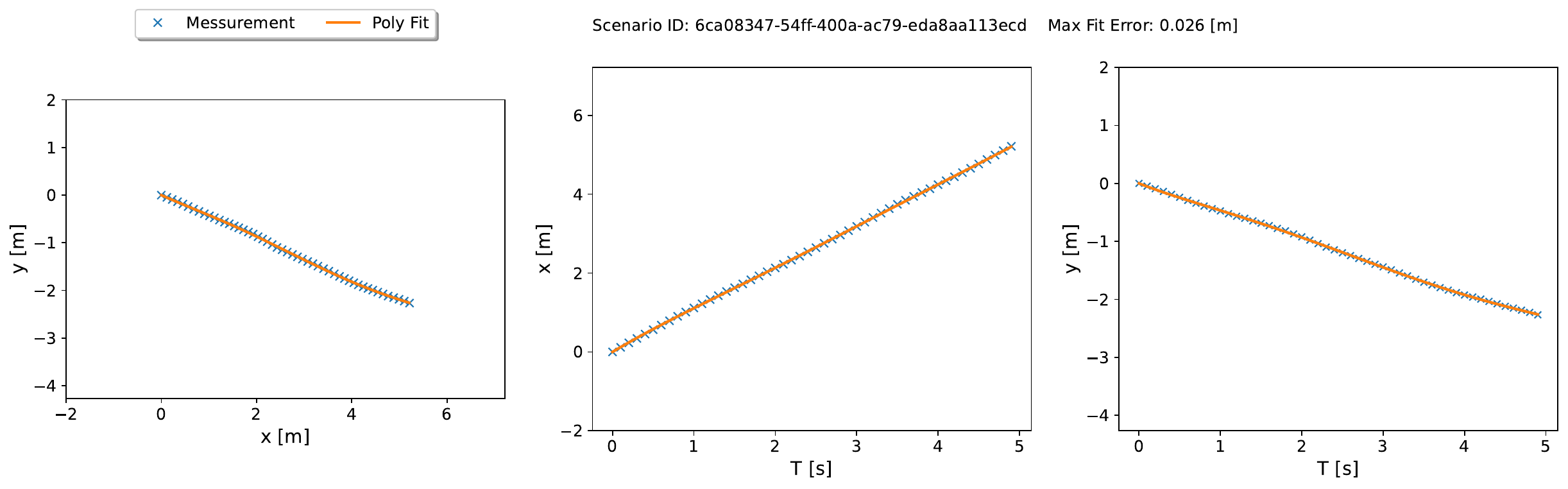} 

\includegraphics[width=5.5in, height=1.8in]{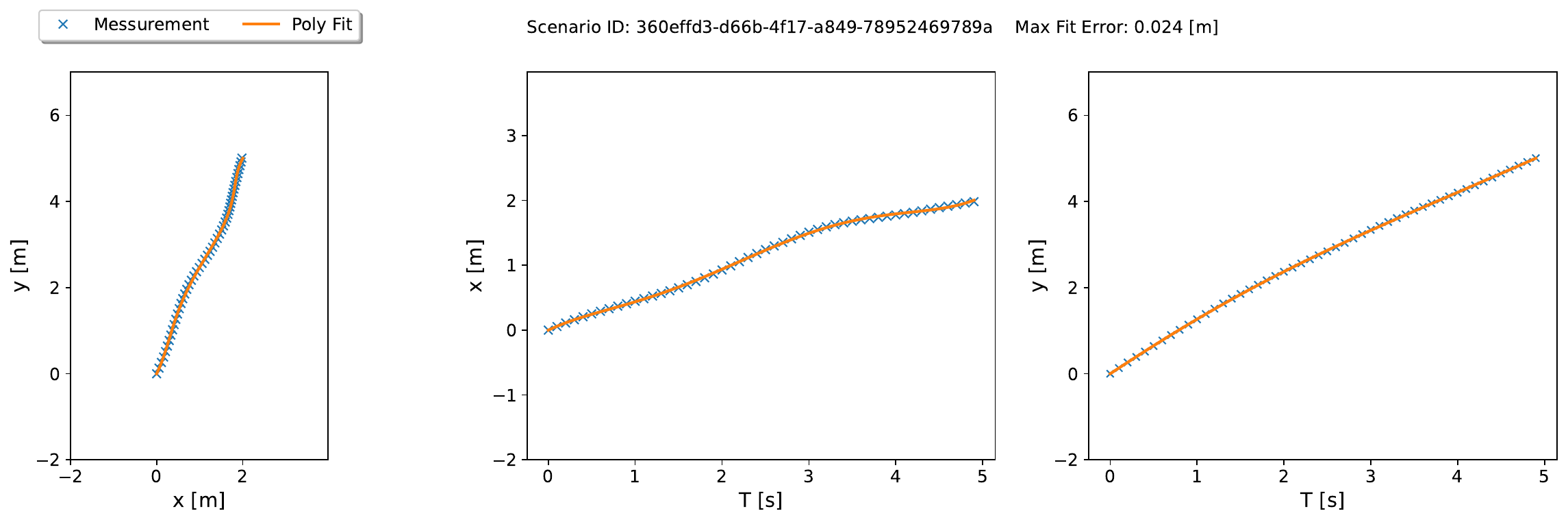} 

\includegraphics[width=5.5in, height=1.8in]{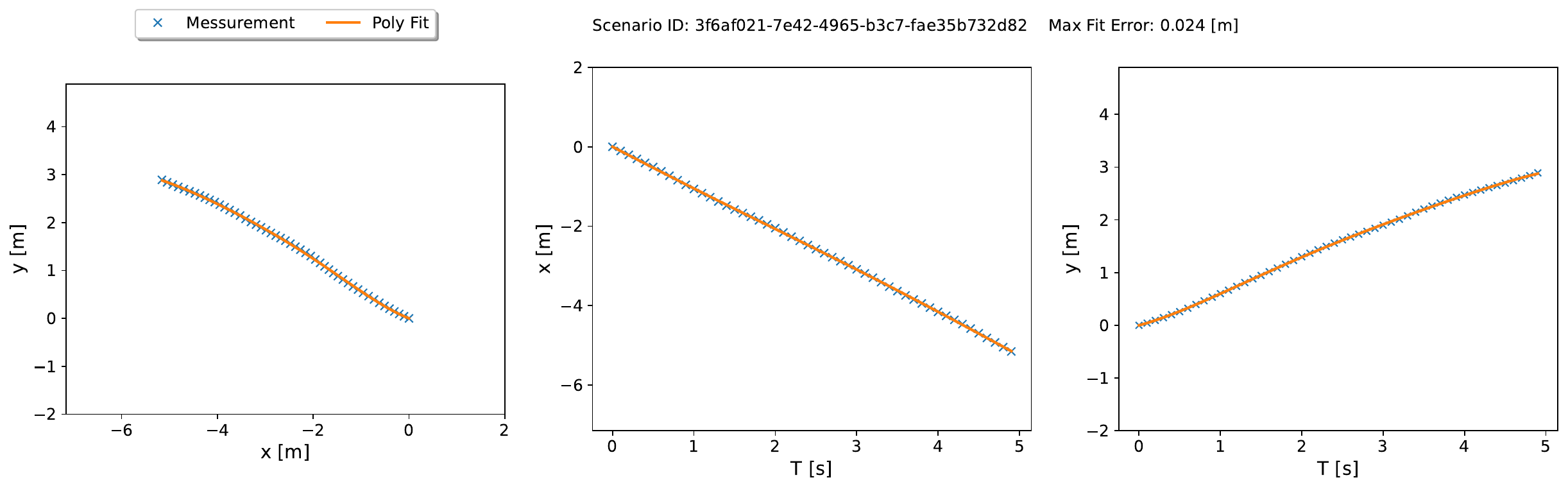} 

\includegraphics[width=5.5in, height=1.8in]{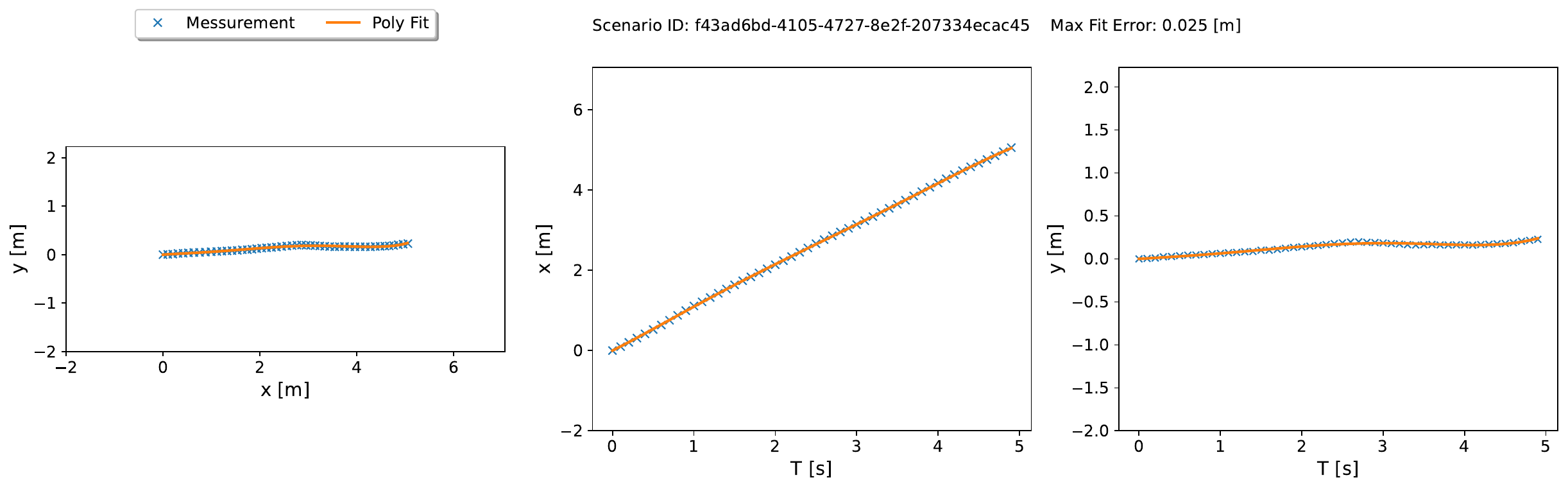} 

\includegraphics[width=5.5in, height=1.8in]{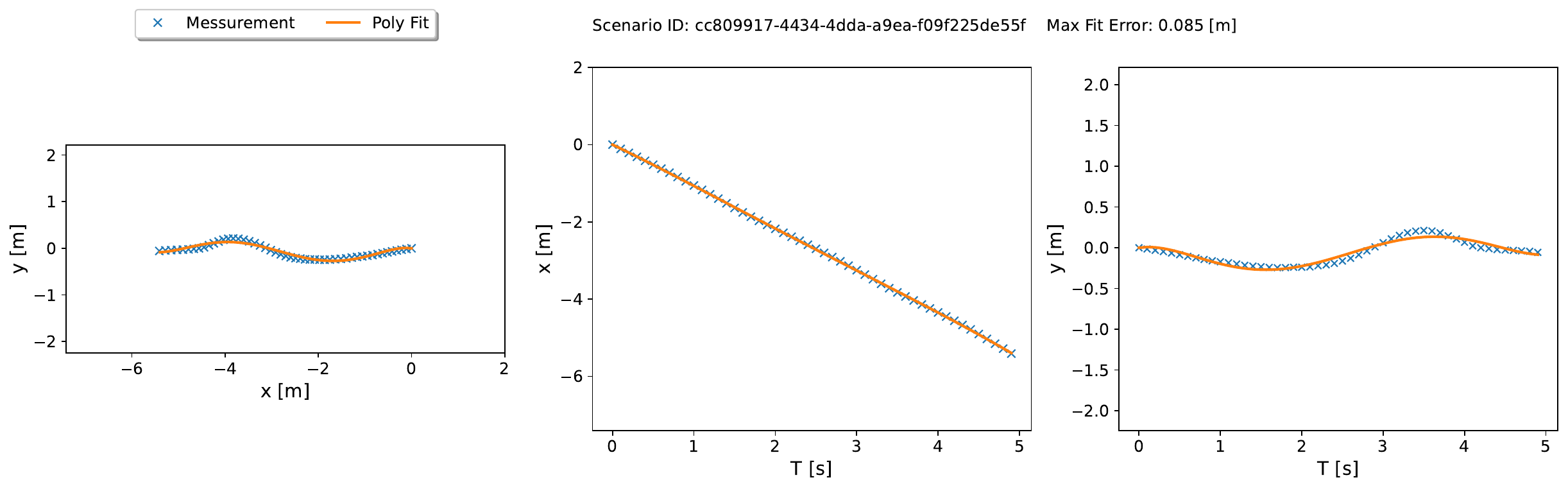} 

\includegraphics[width=5.5in, height=1.7in]{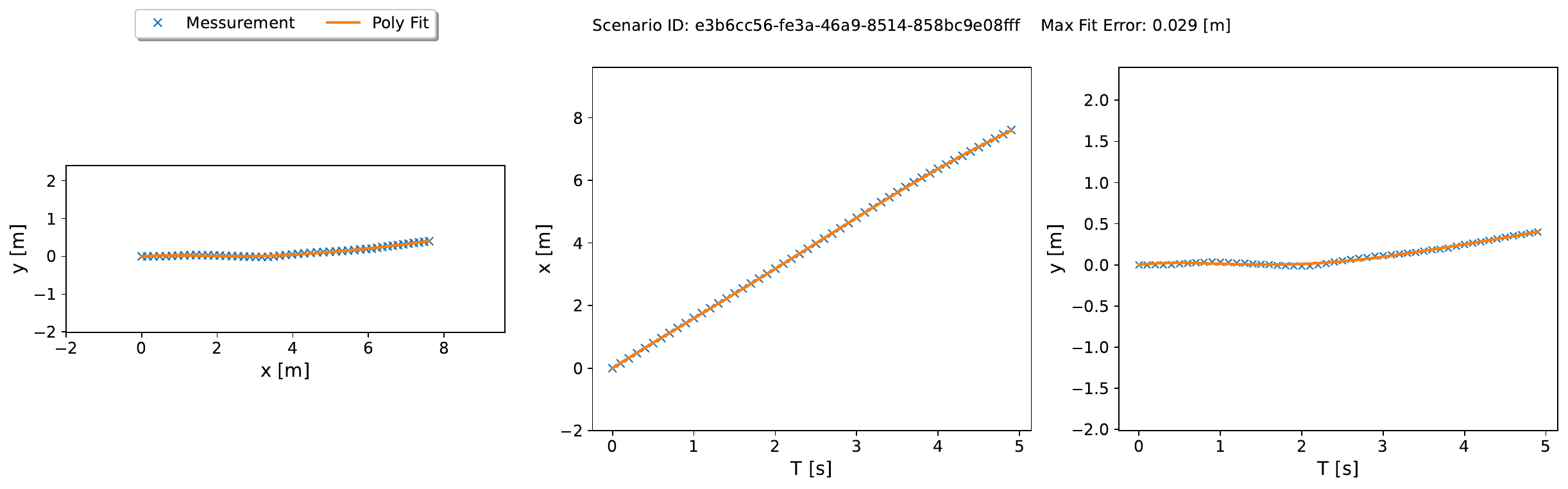} 

\section{10 5-seconds vehicle trajectories with highest fit error in WO (Fitted with $\hat{n} = 5$)}
\centering
\includegraphics[width=5.5in, height=1.8in]{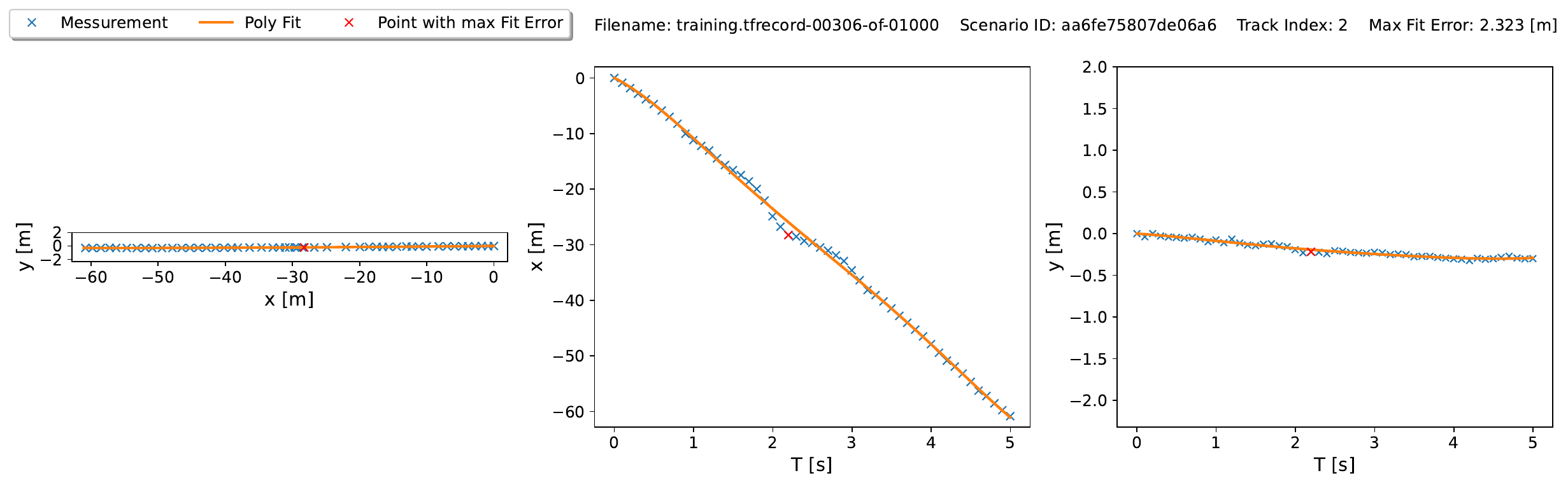} 

\includegraphics[width=5.5in, height=1.8in]{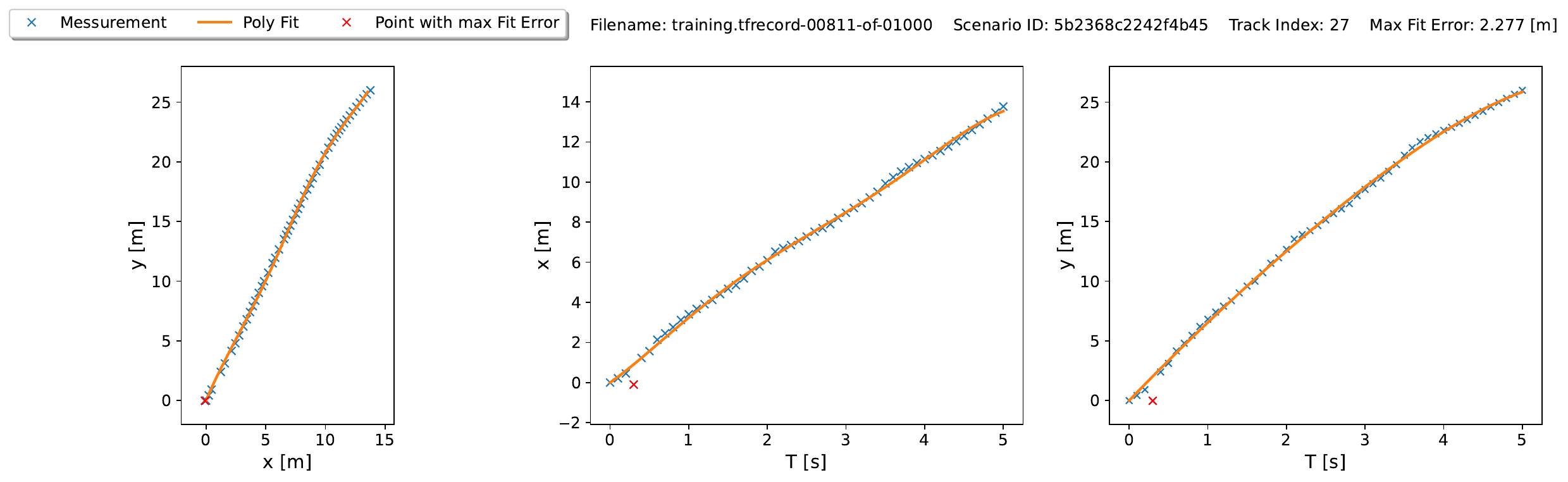} 

\includegraphics[width=5.5in, height=1.8in]{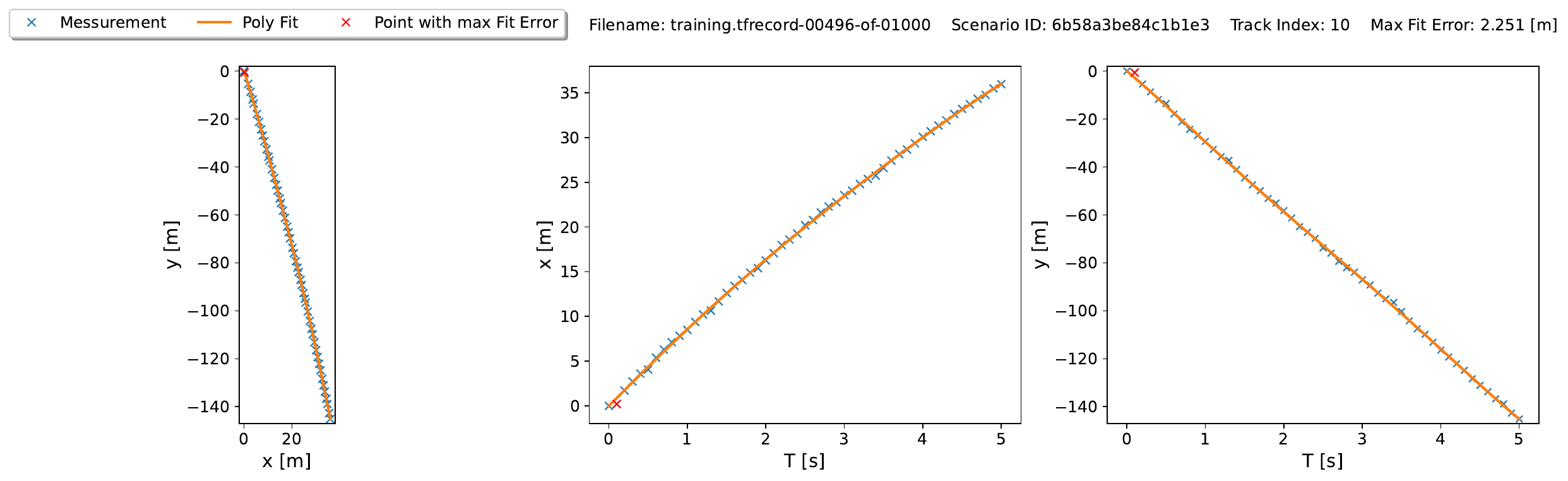} 

\includegraphics[width=5.5in, height=1.8in]{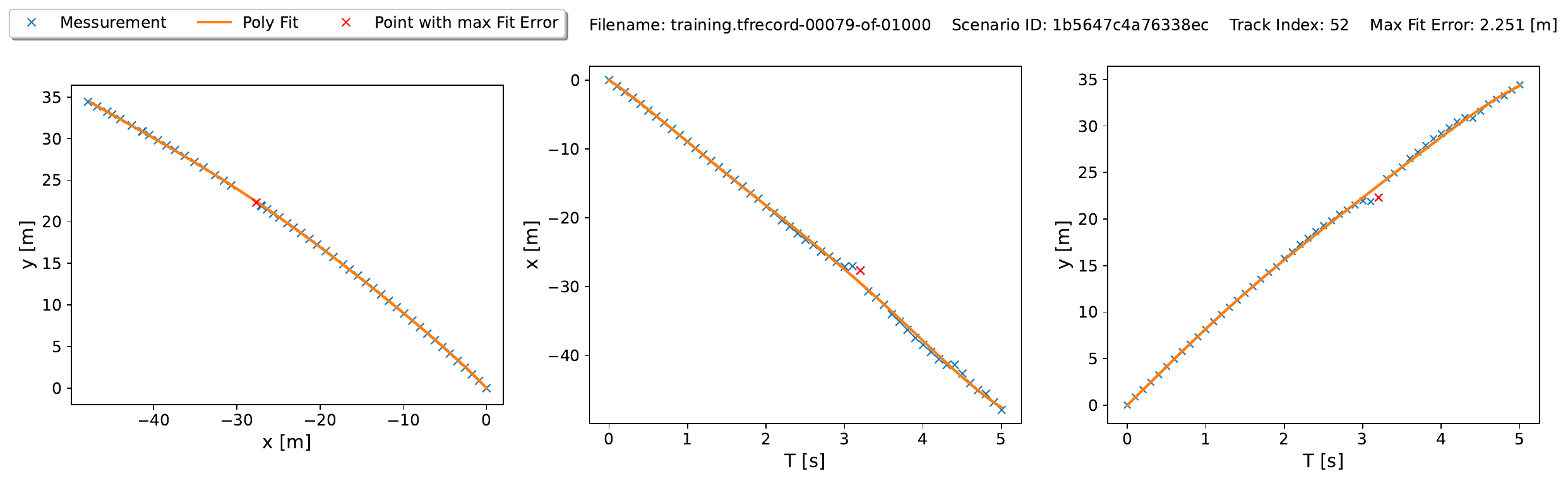} 

\includegraphics[width=5.5in, height=1.7in]{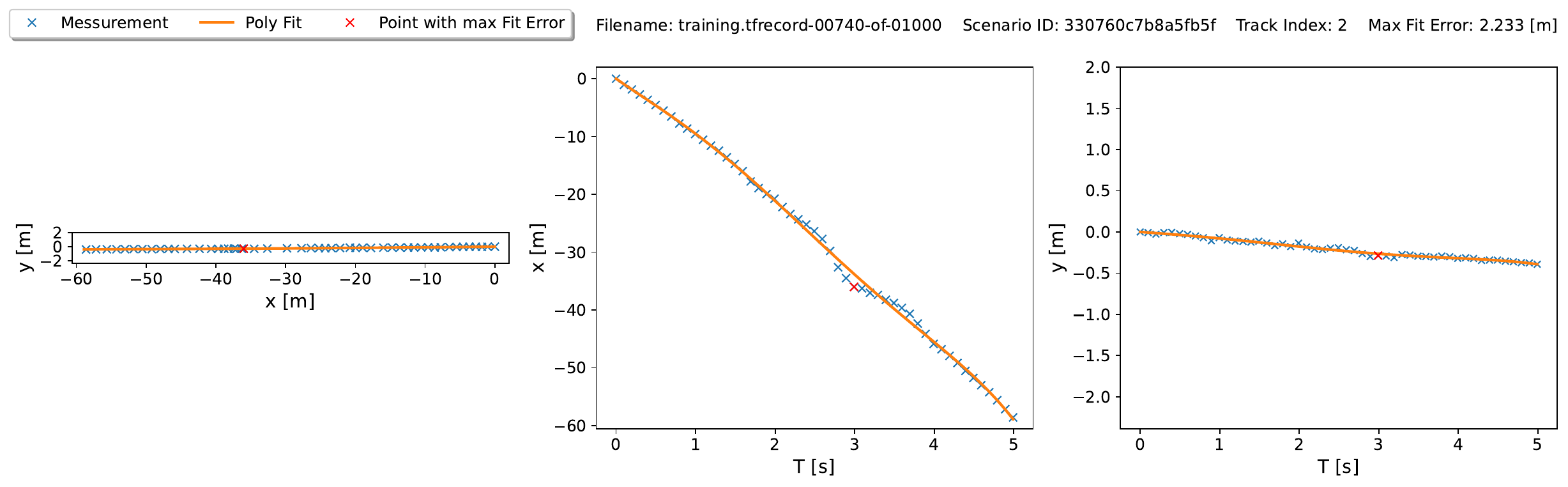} 

\includegraphics[width=5.5in, height=1.8in]{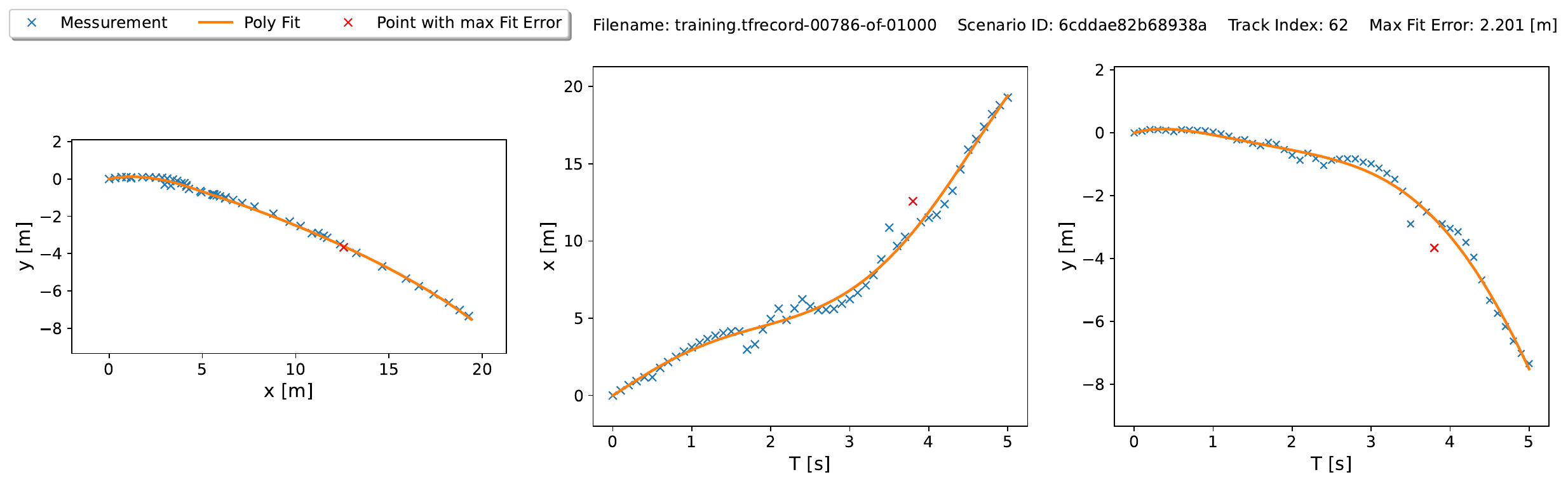} 

\includegraphics[width=5.5in, height=1.8in]{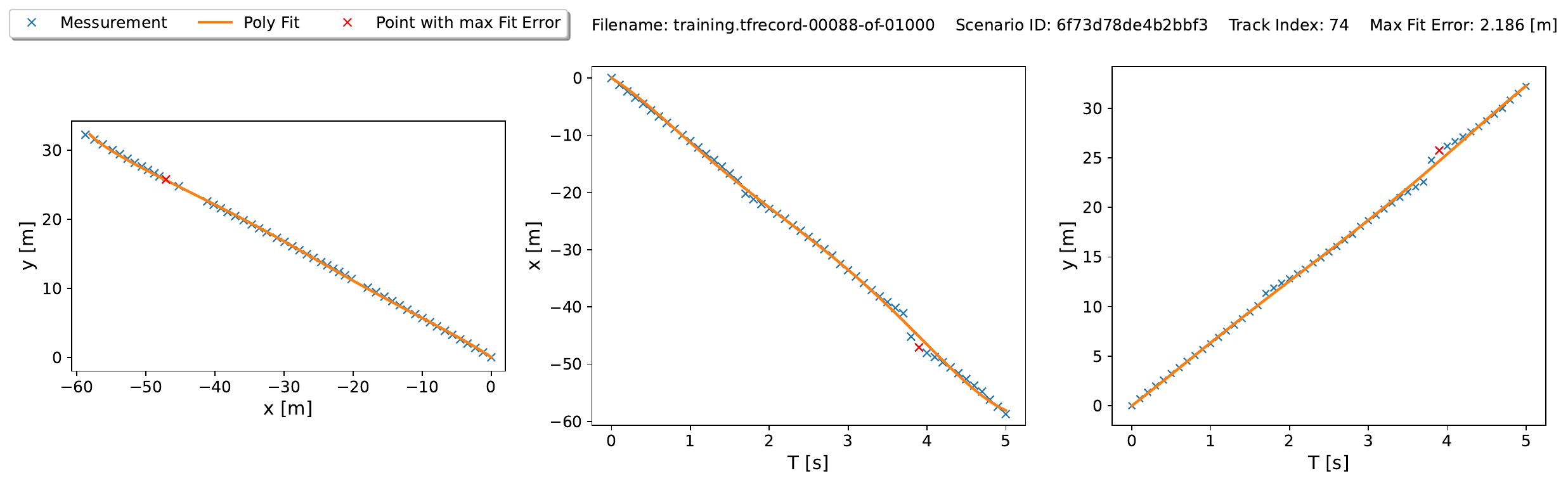} 

\includegraphics[width=5.5in, height=1.8in]{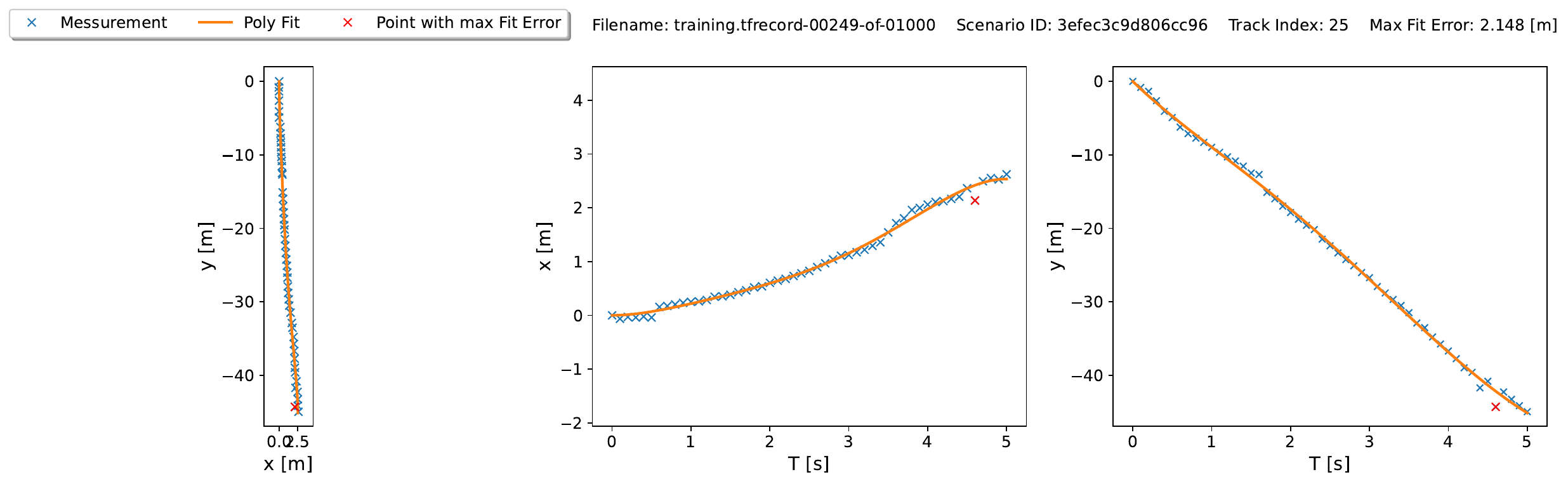} 

\includegraphics[width=5.5in, height=1.8in]{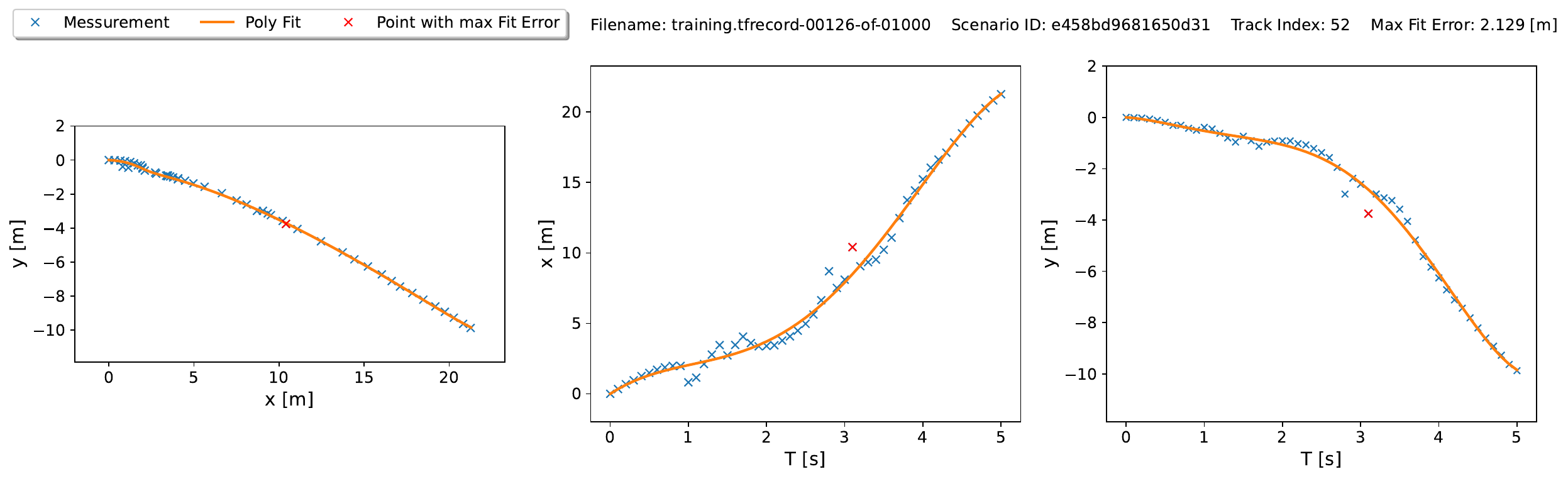} 

\includegraphics[width=5.5in, height=1.7in]{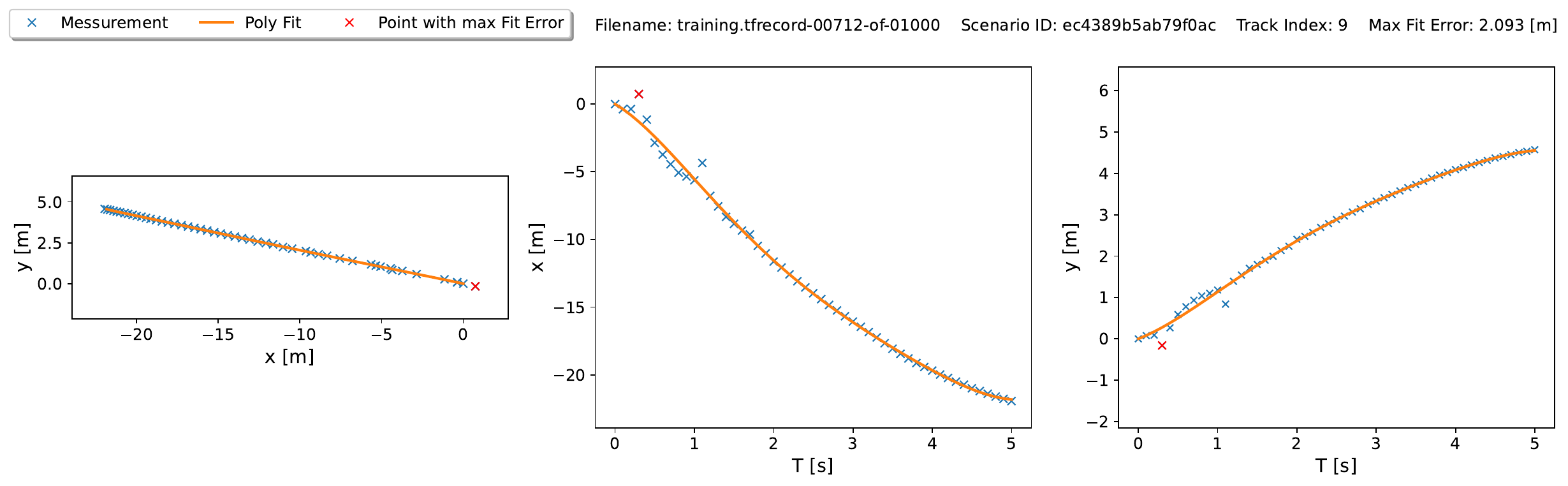} 

\section{10 random 5-seconds vehicle trajectories in WO (Fitted with $\hat{n} = 5$)}
\centering
\includegraphics[width=5.5in, height=1.8in]{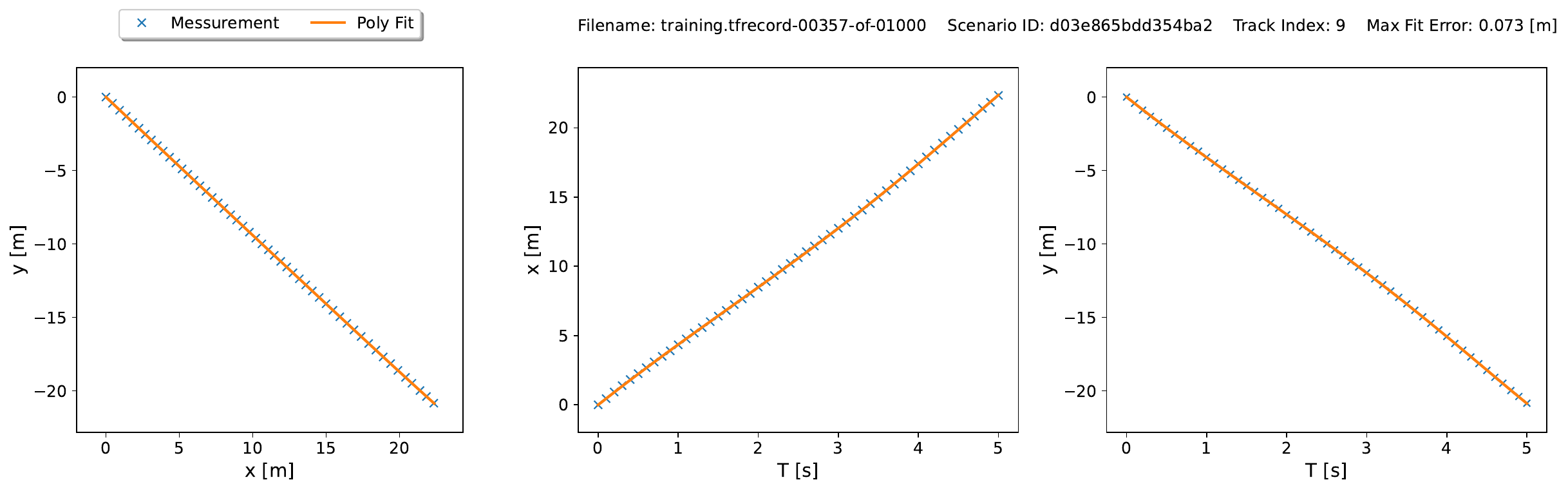} 

\includegraphics[width=5.5in, height=1.8in]{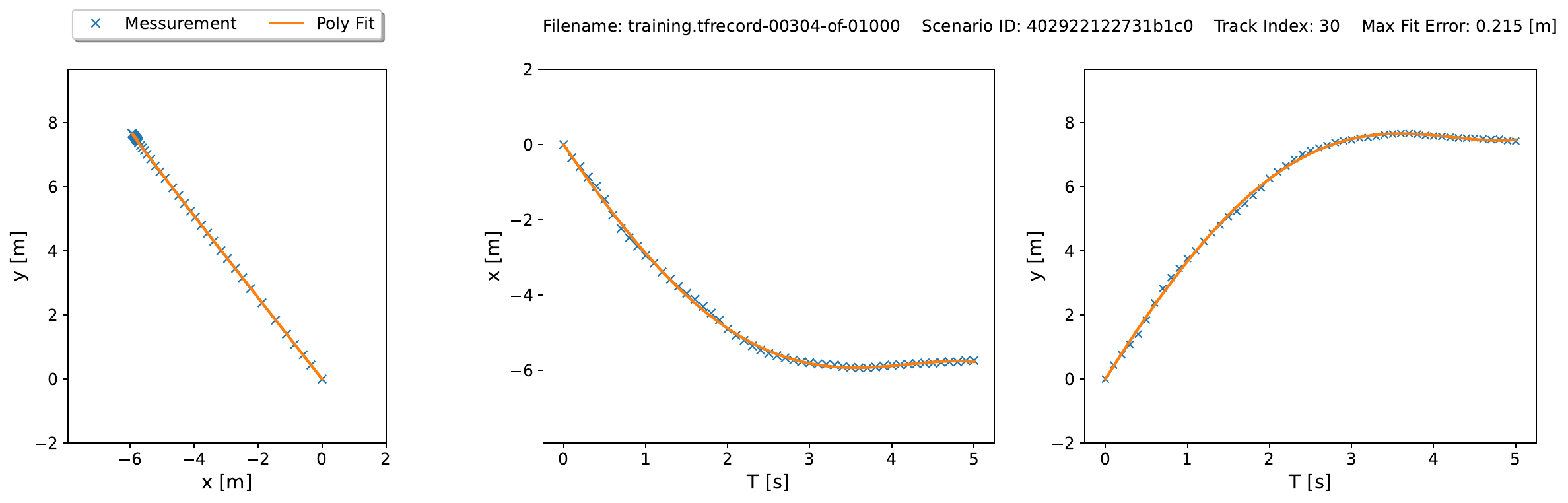} 

\includegraphics[width=5.5in, height=1.8in]{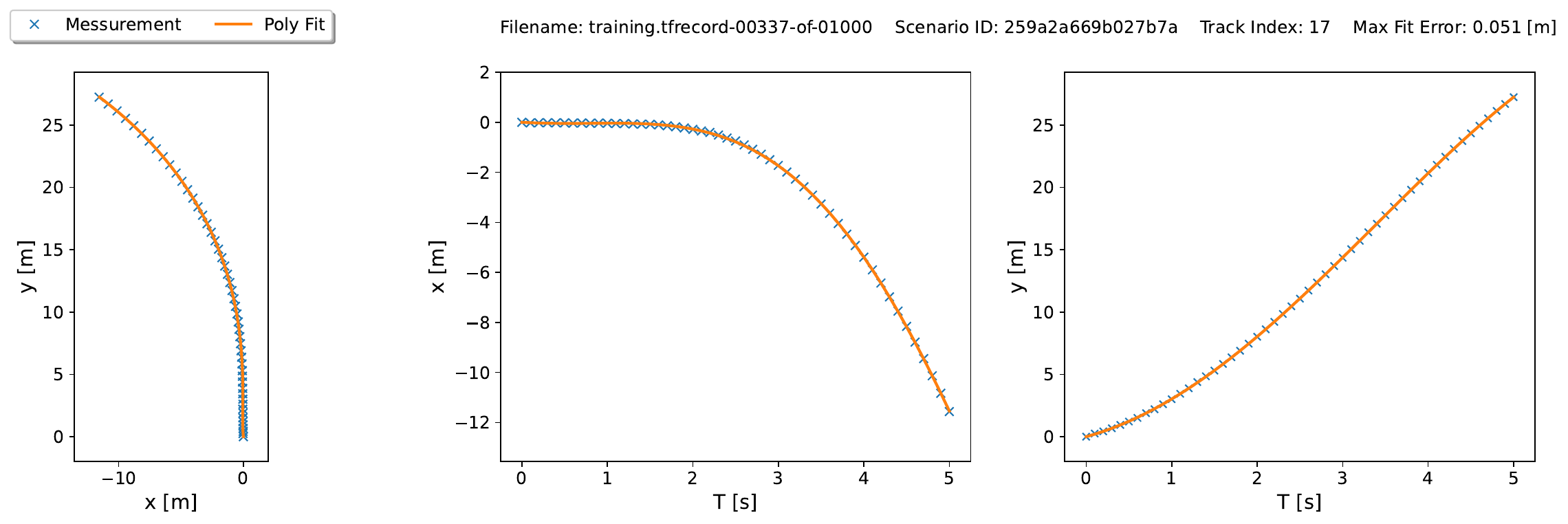} 

\includegraphics[width=5.5in, height=1.8in]{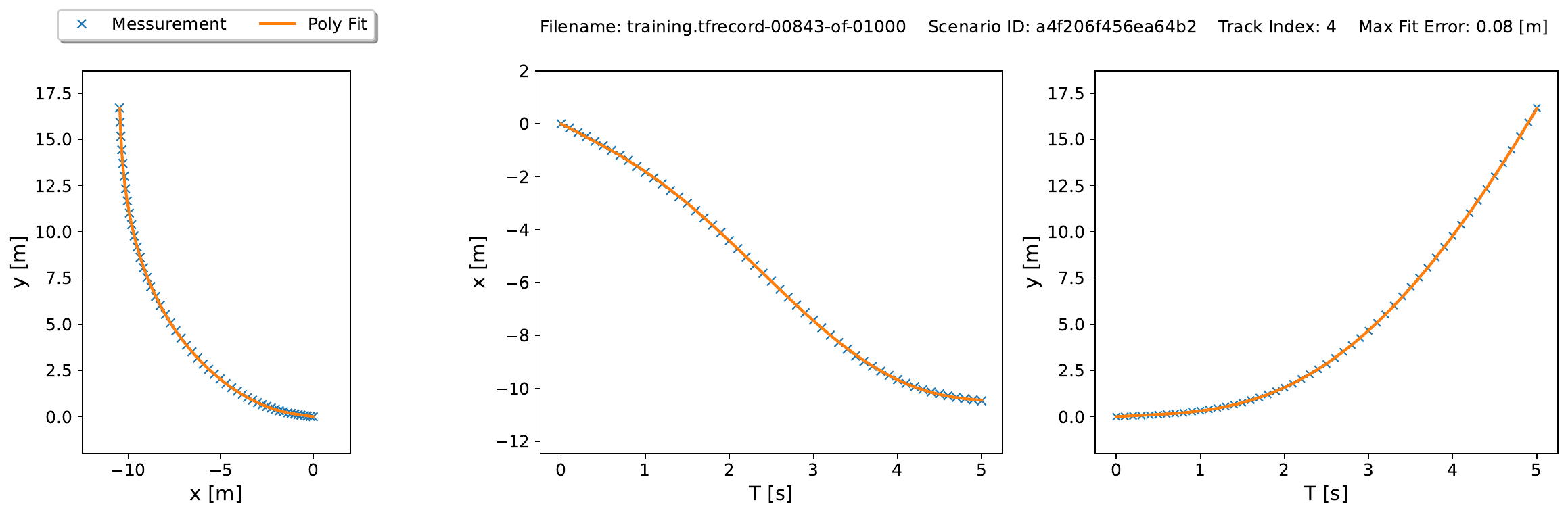} 

\includegraphics[width=5.5in, height=1.7in]{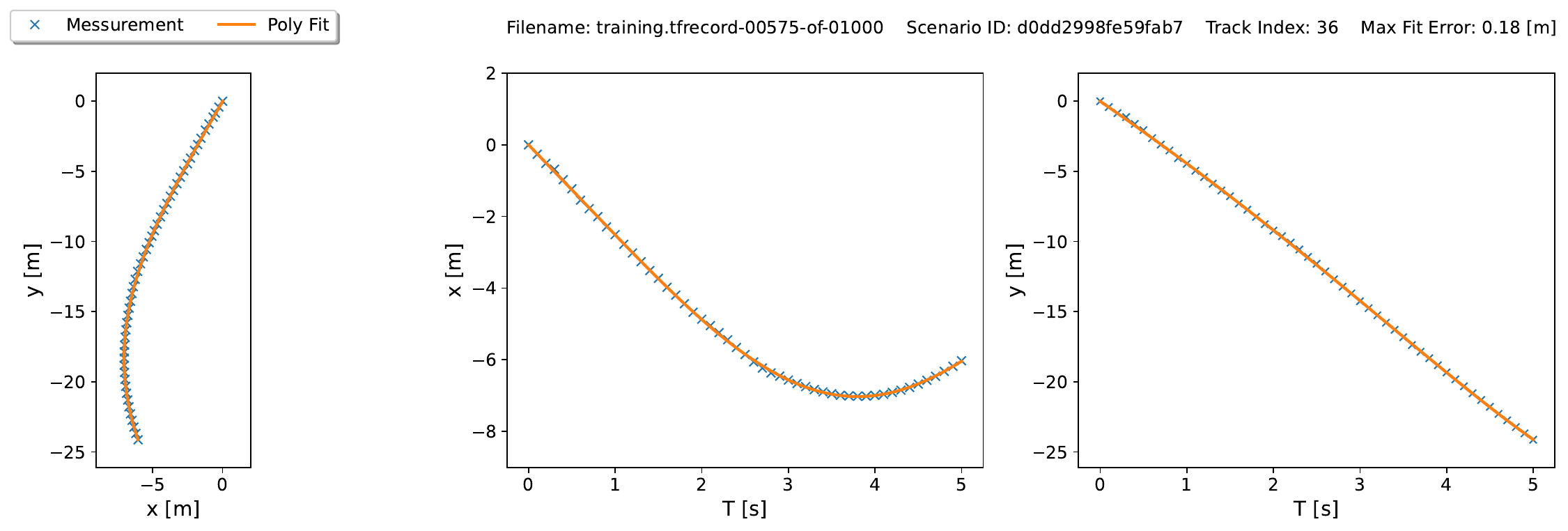} 

\includegraphics[width=5.5in, height=1.8in]{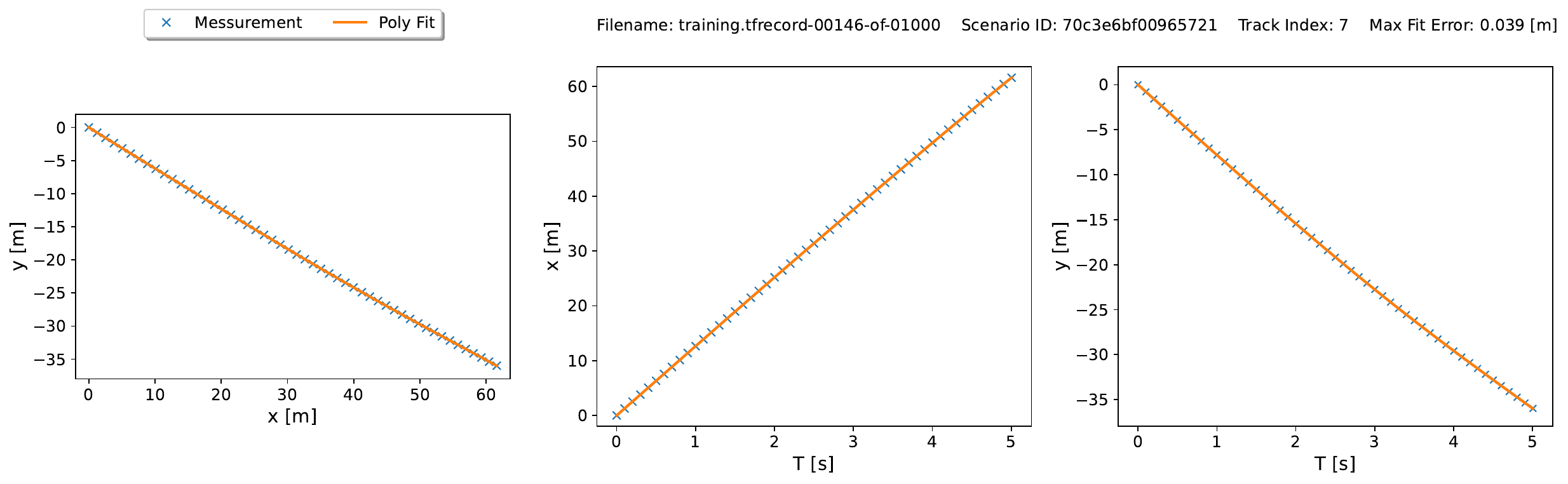} 

\includegraphics[width=5.5in, height=1.8in]{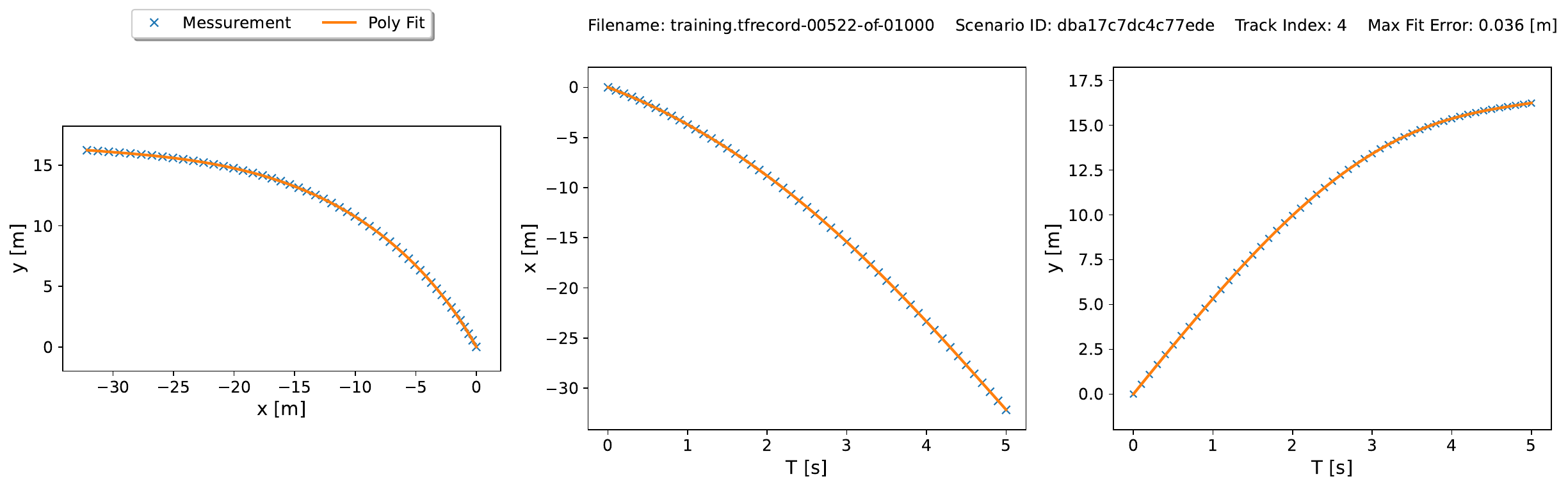} 

\includegraphics[width=5.5in, height=1.8in]{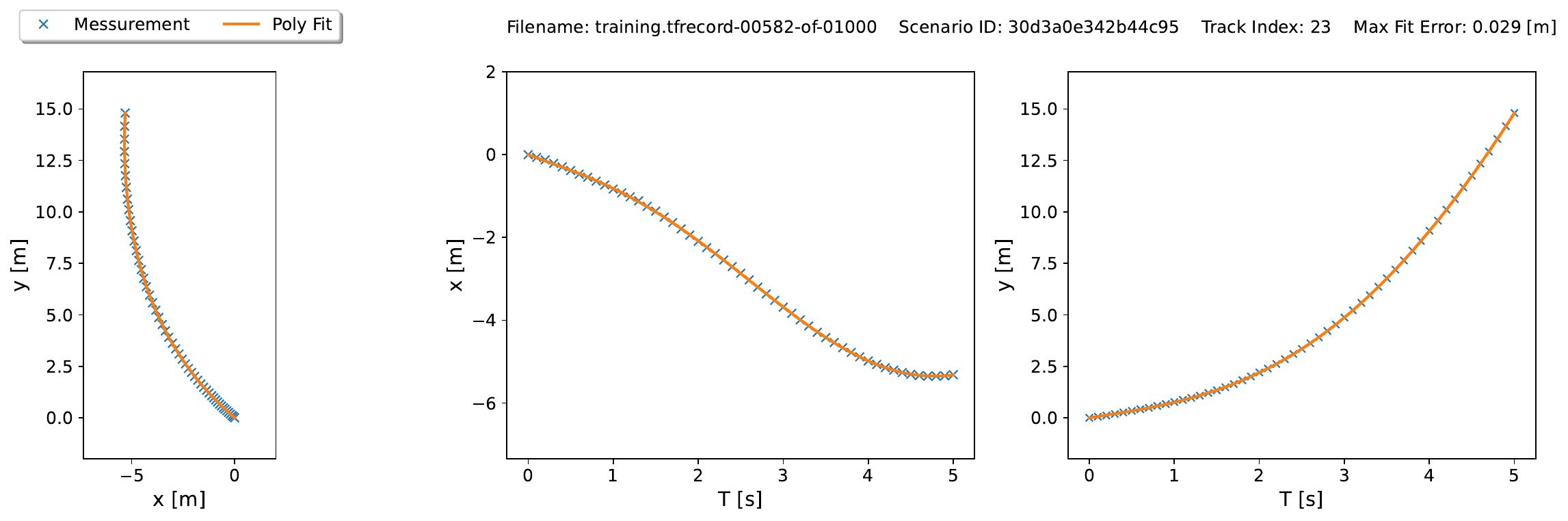} 

\includegraphics[width=5.5in, height=1.8in]{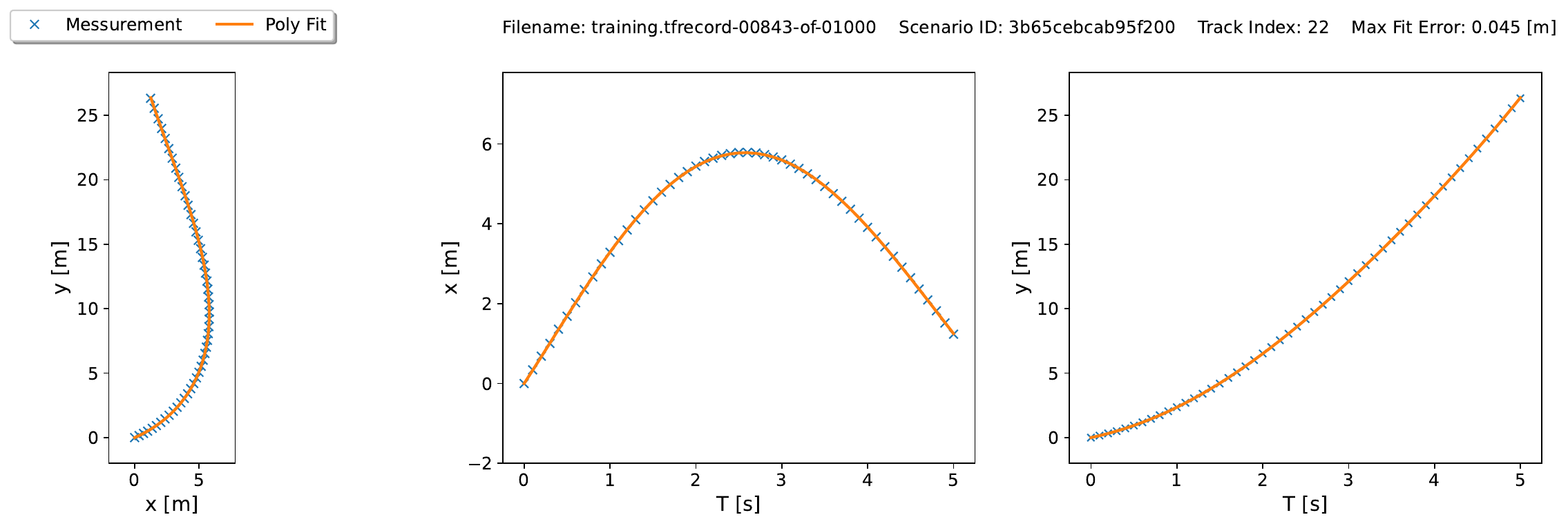} 

\includegraphics[width=5.5in, height=1.7in]{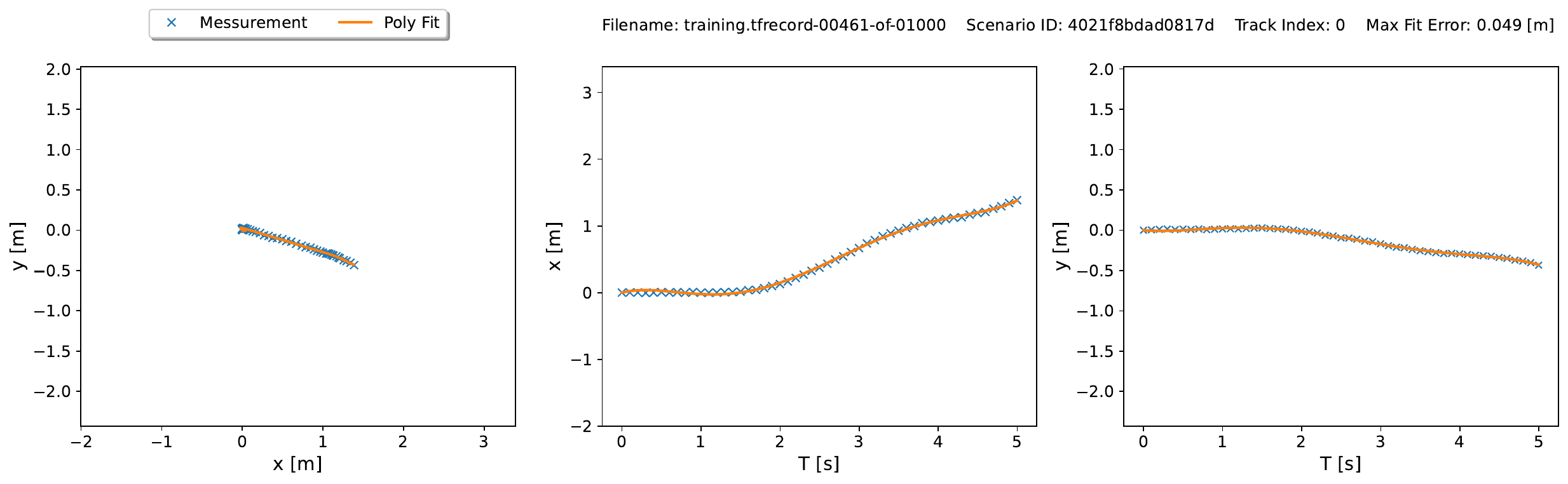} 

\section{10 5-seconds cyclist trajectories with highest fit error in WO (Fitted with $\hat{n} = 5$)}
\centering
\includegraphics[width=5.5in, height=1.8in]{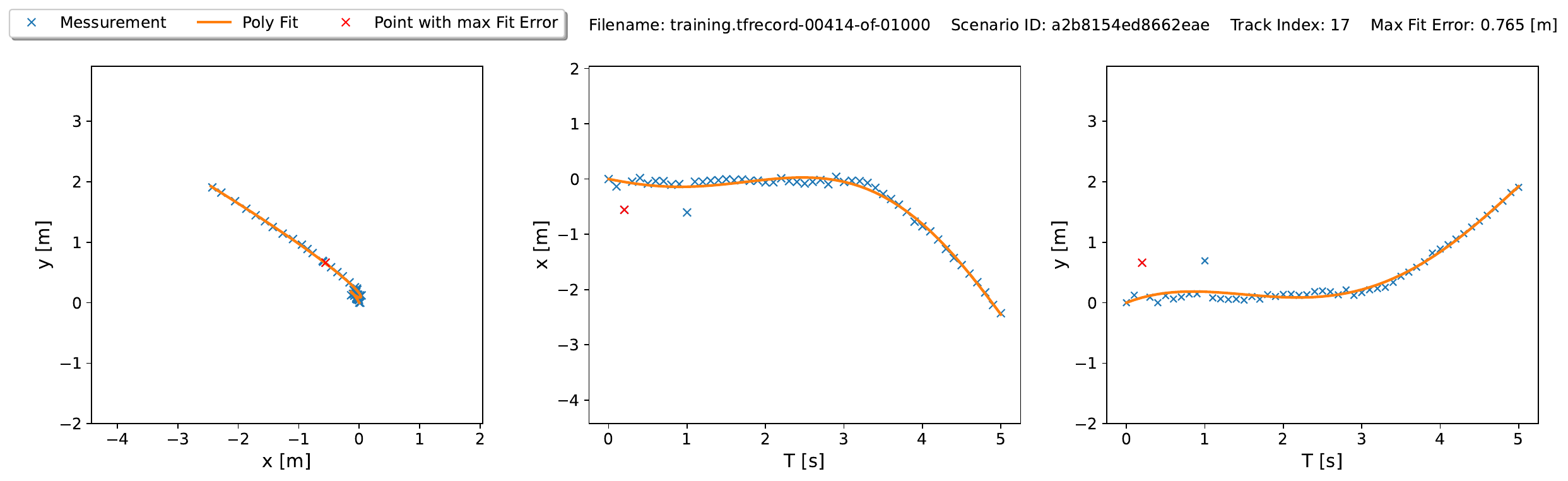} 

\includegraphics[width=5.5in, height=1.8in]{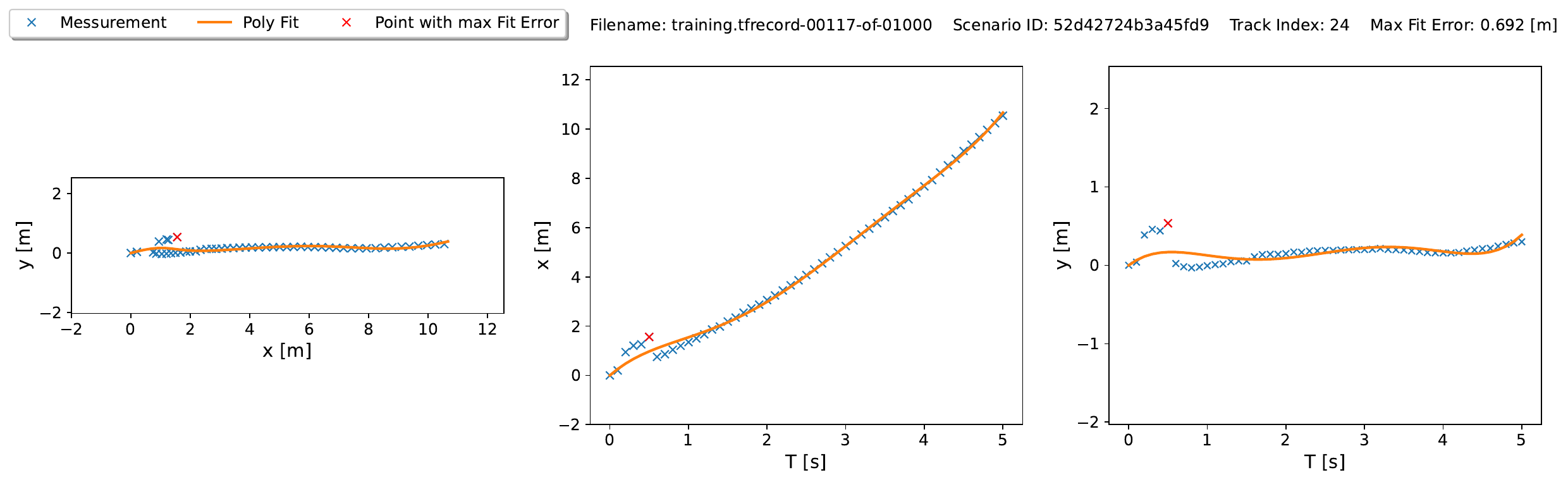} 

\includegraphics[width=5.5in, height=1.8in]{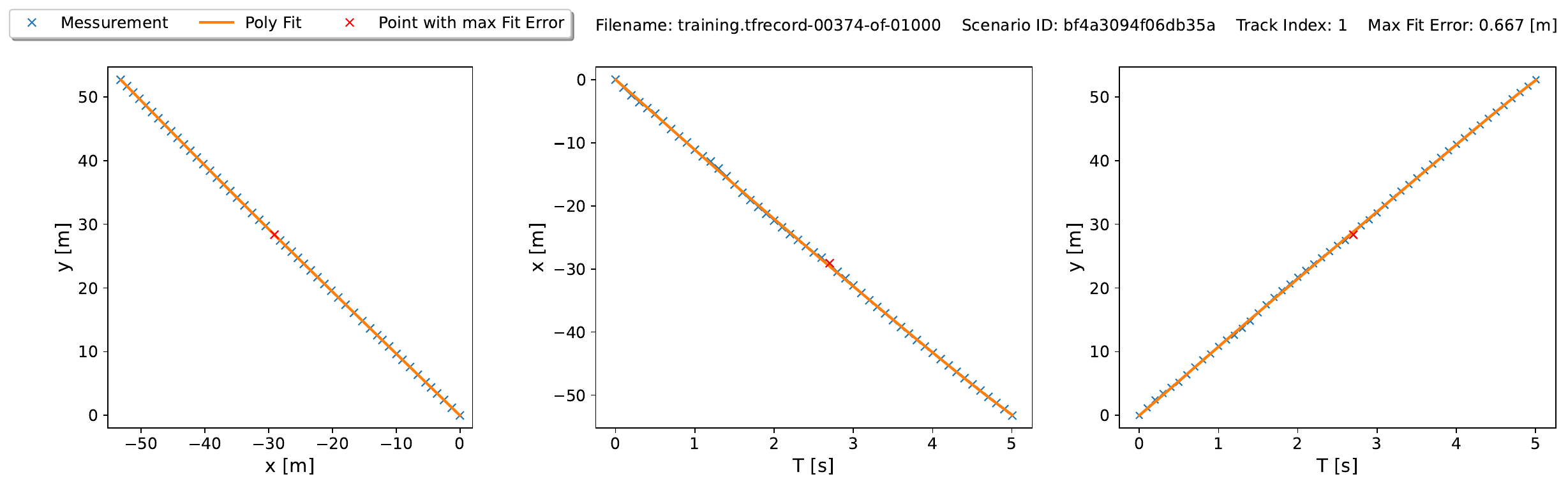} 

\includegraphics[width=5.5in, height=1.8in]{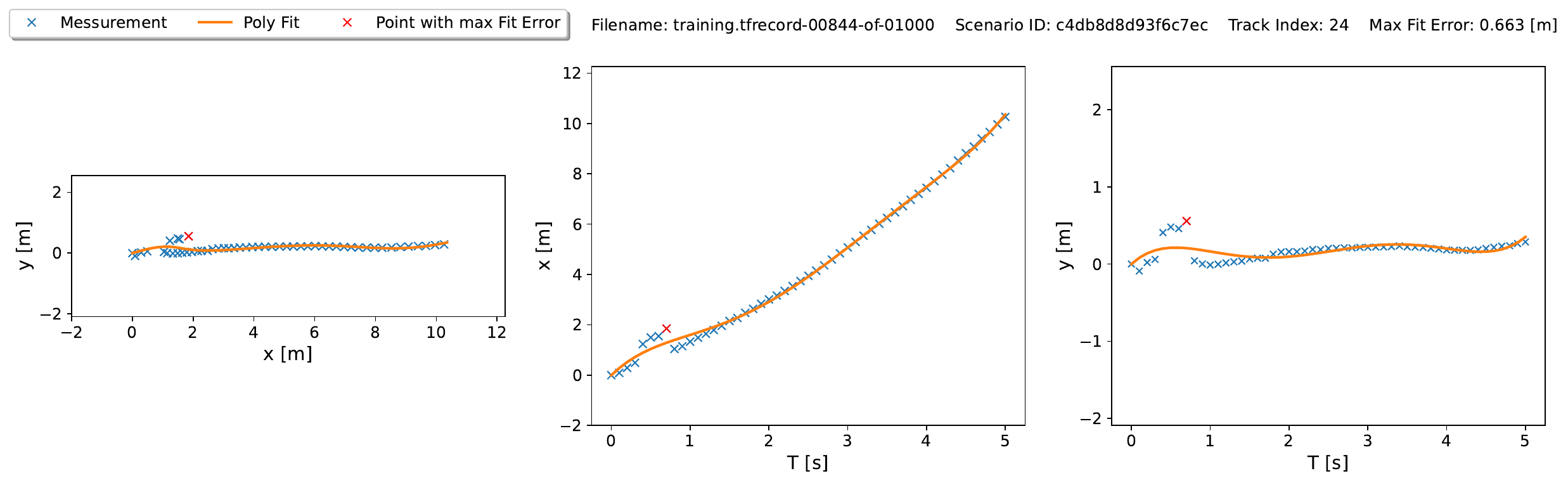} 

\includegraphics[width=5.5in, height=1.7in]{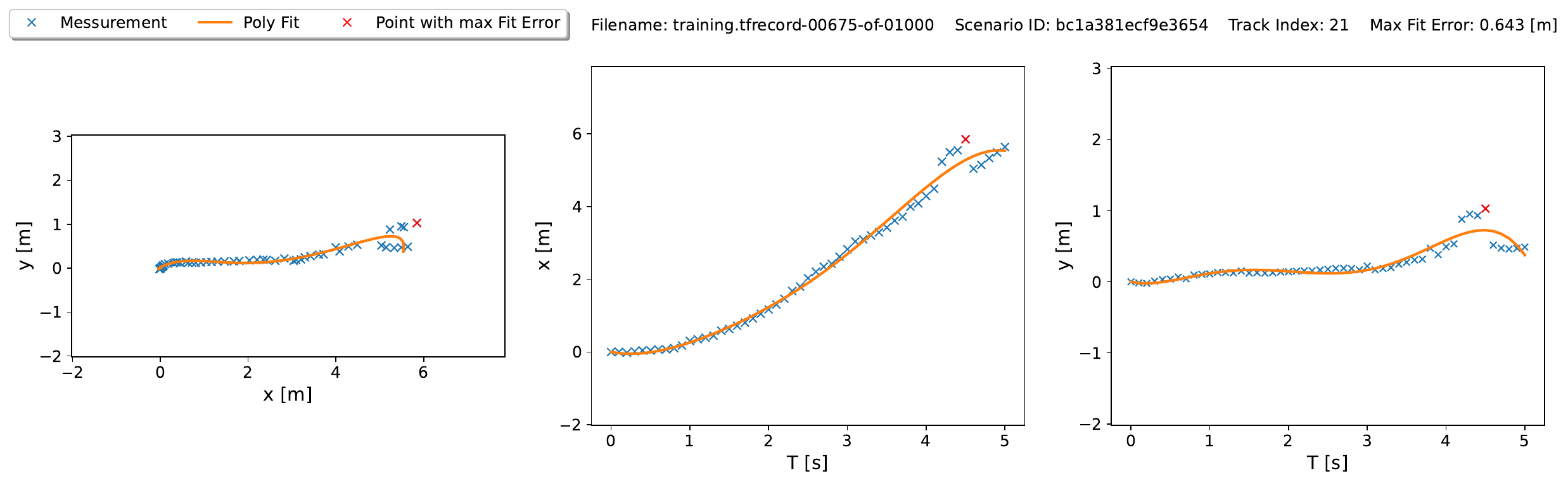} 

\includegraphics[width=5.5in, height=1.8in]{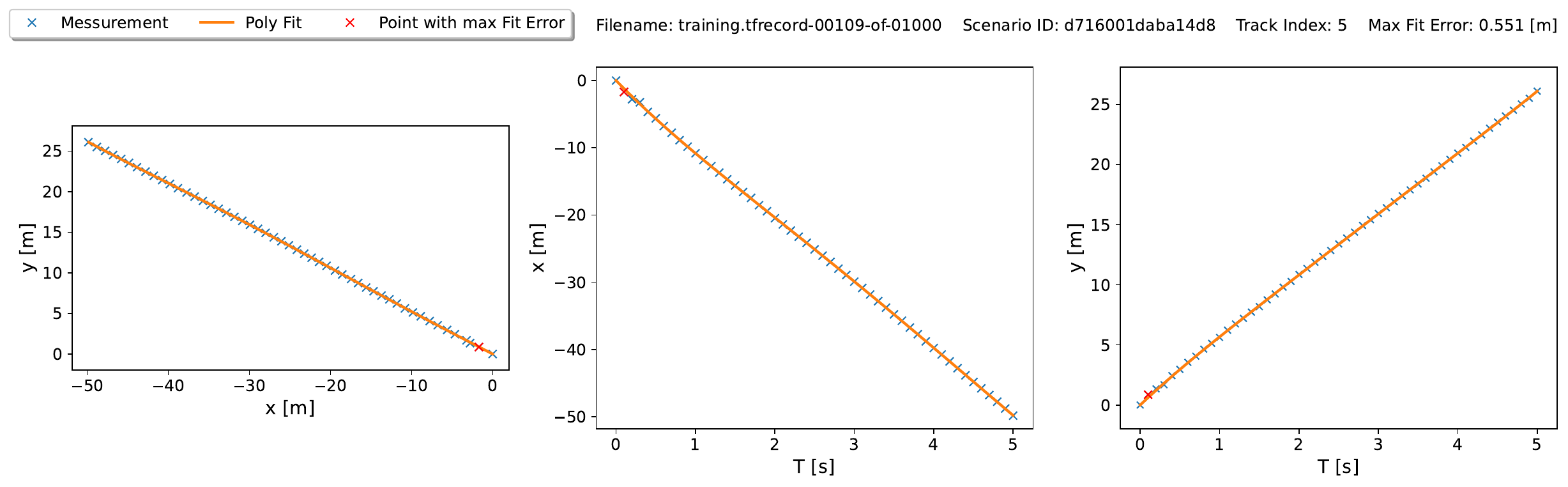} 

\includegraphics[width=5.5in, height=1.8in]{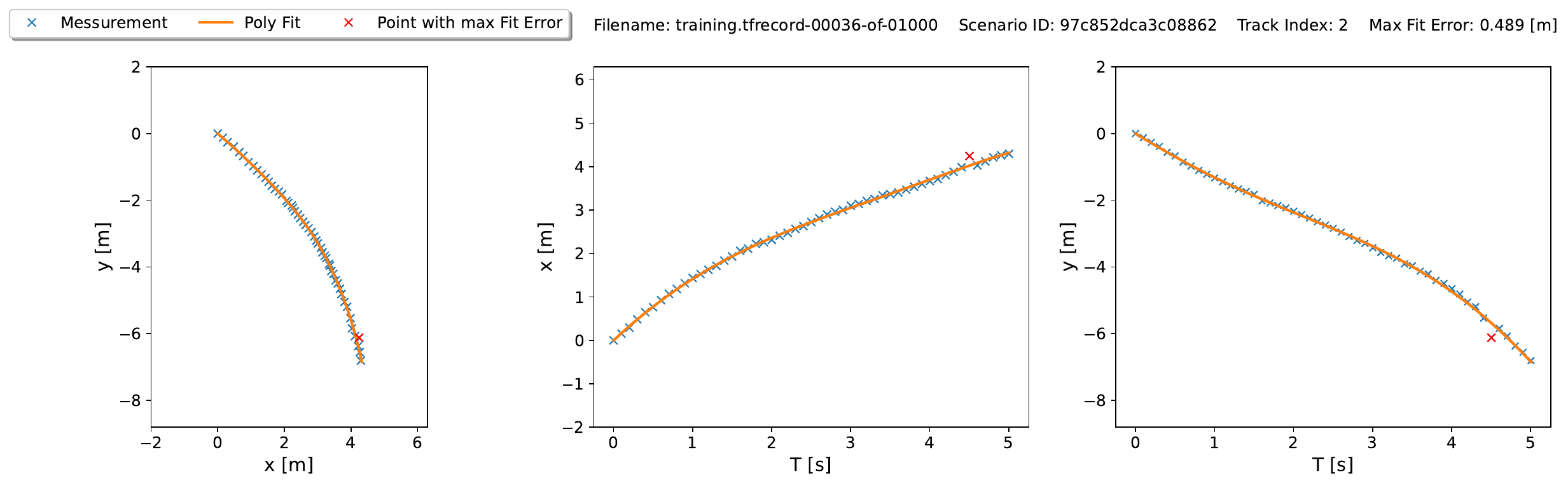} 

\includegraphics[width=5.5in, height=1.8in]{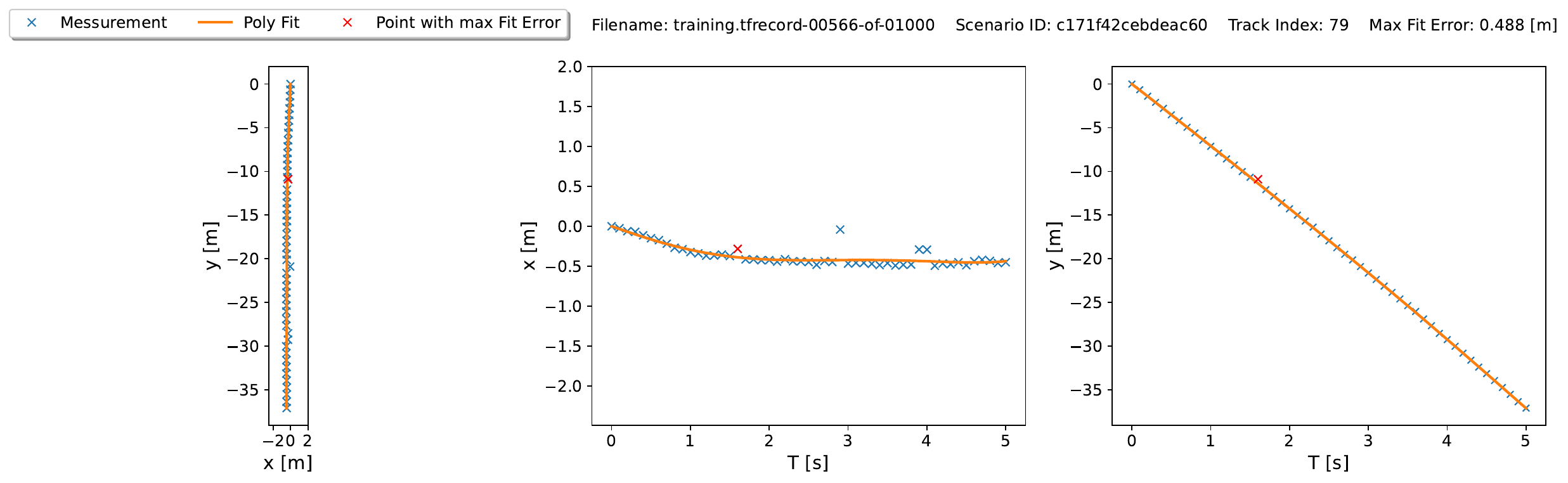} 

\includegraphics[width=5.5in, height=1.8in]{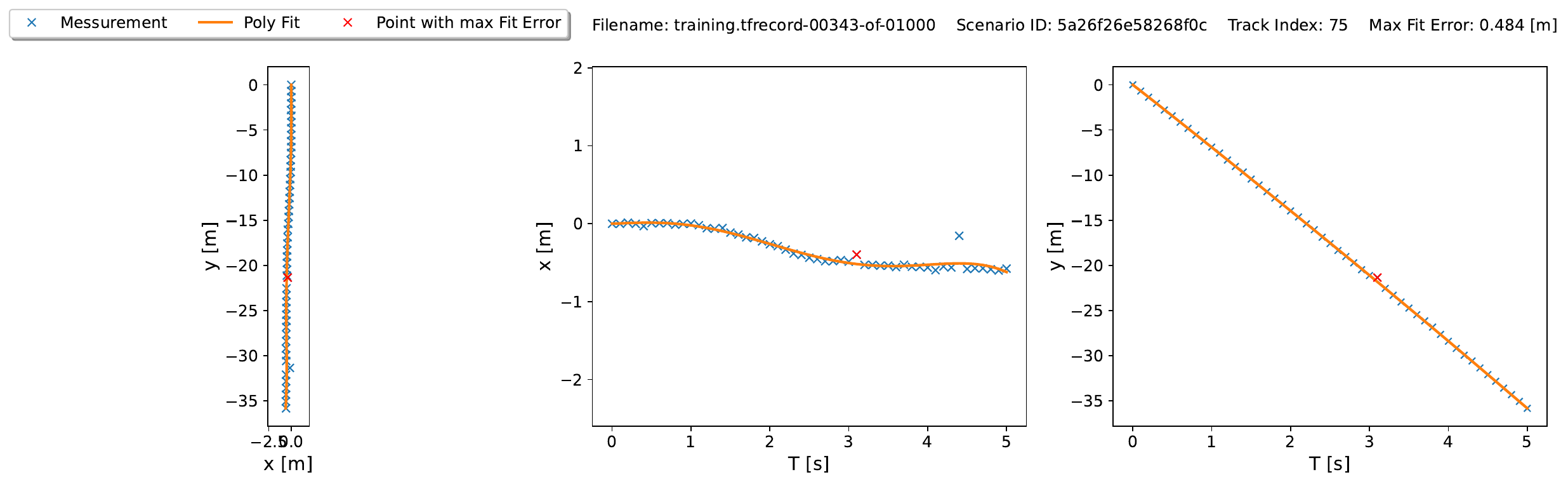} 

\includegraphics[width=5.5in, height=1.7in]{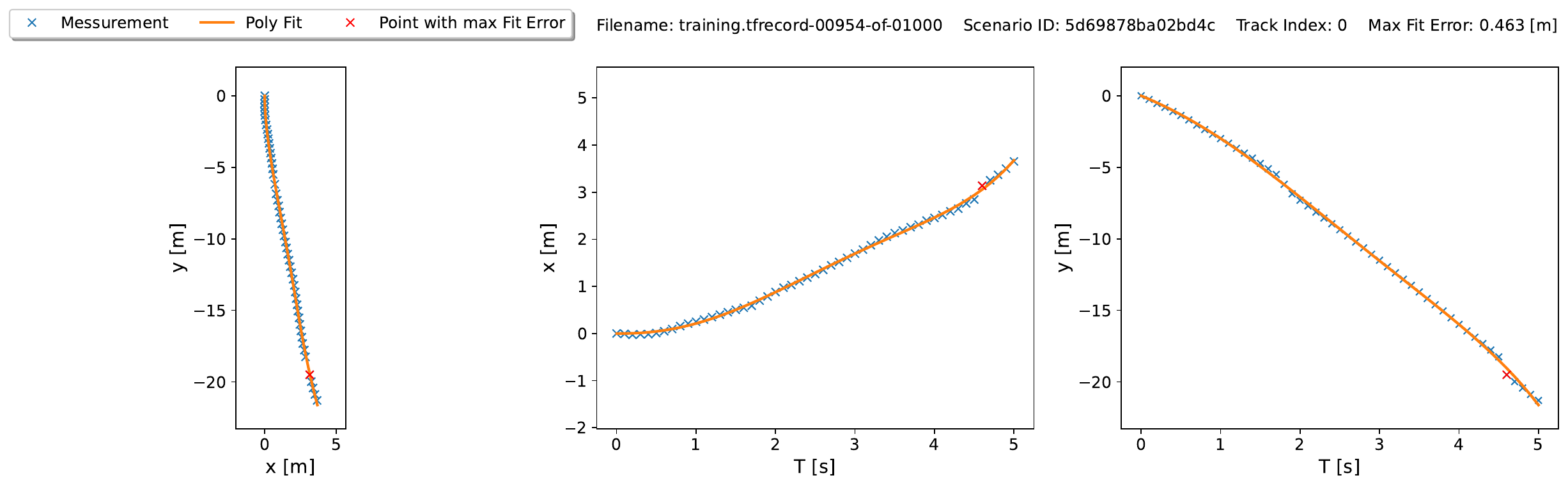} 

\section{10 random 5-seconds cyclist trajectories in WO (Fitted with $\hat{n} = 5$)}
\centering
\includegraphics[width=5.5in, height=1.8in]{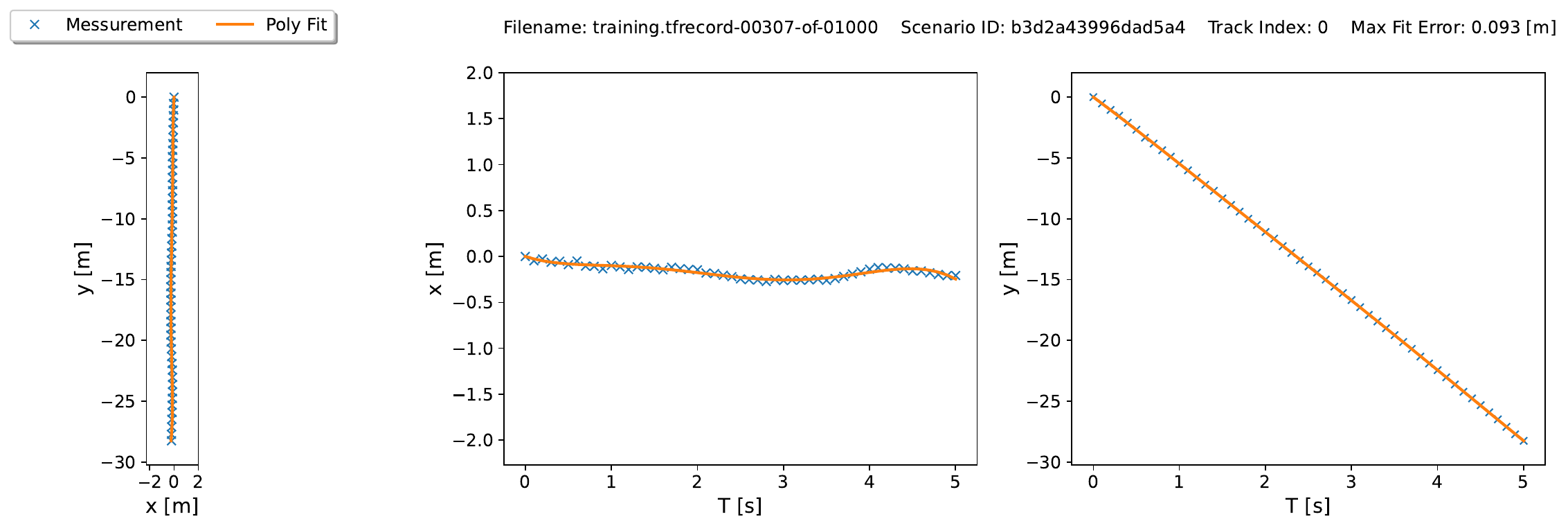} 

\includegraphics[width=5.5in, height=1.8in]{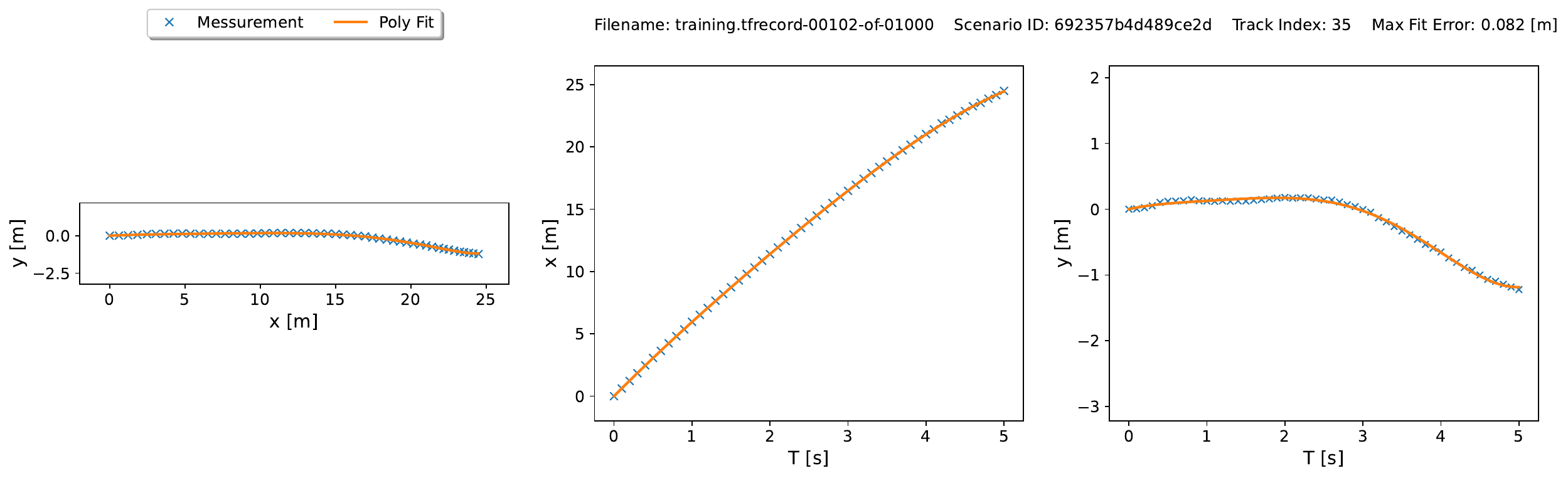} 

\includegraphics[width=5.5in, height=1.8in]{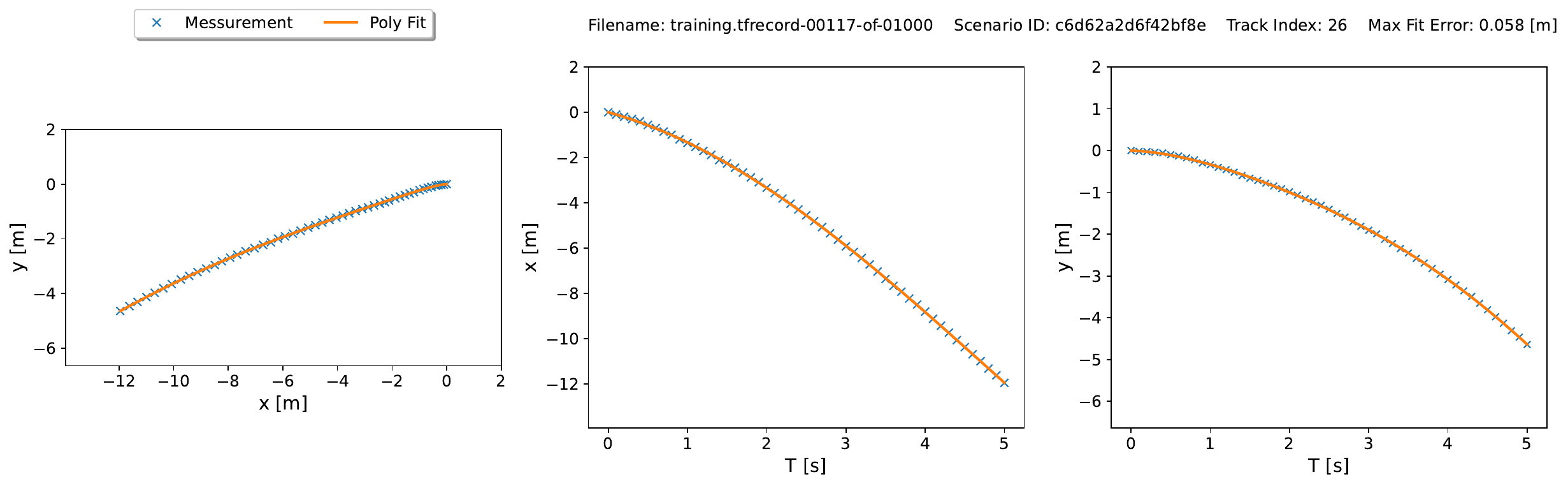} 

\includegraphics[width=5.5in, height=1.8in]{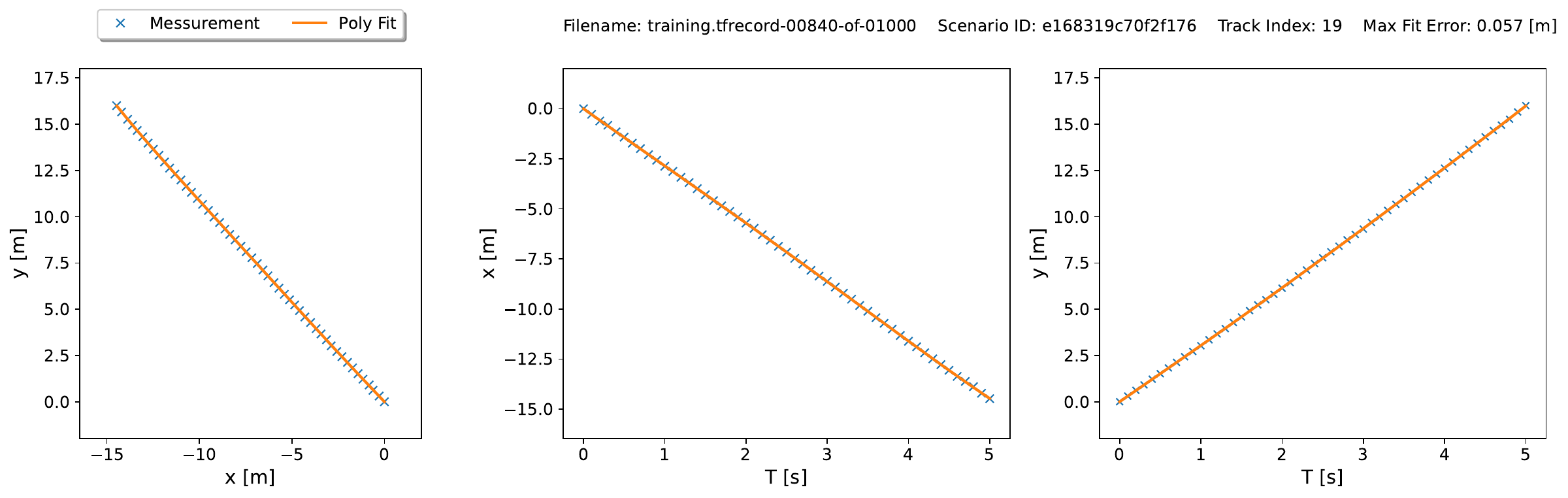} 

\includegraphics[width=5.5in, height=1.7in]{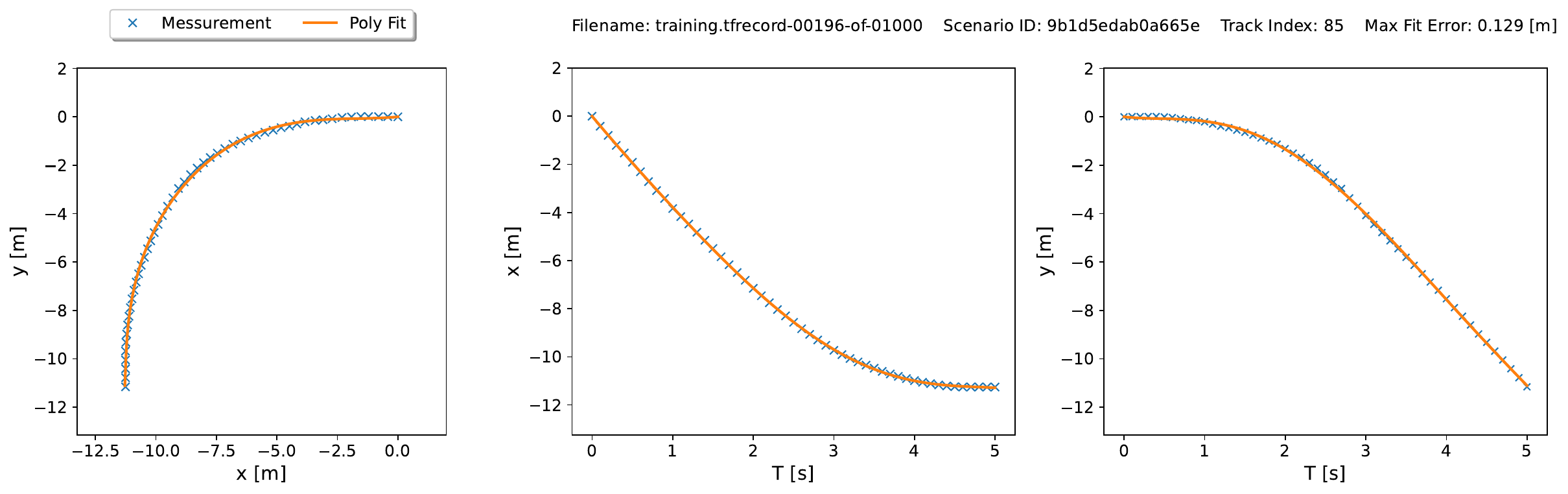} 

\includegraphics[width=5.5in, height=1.8in]{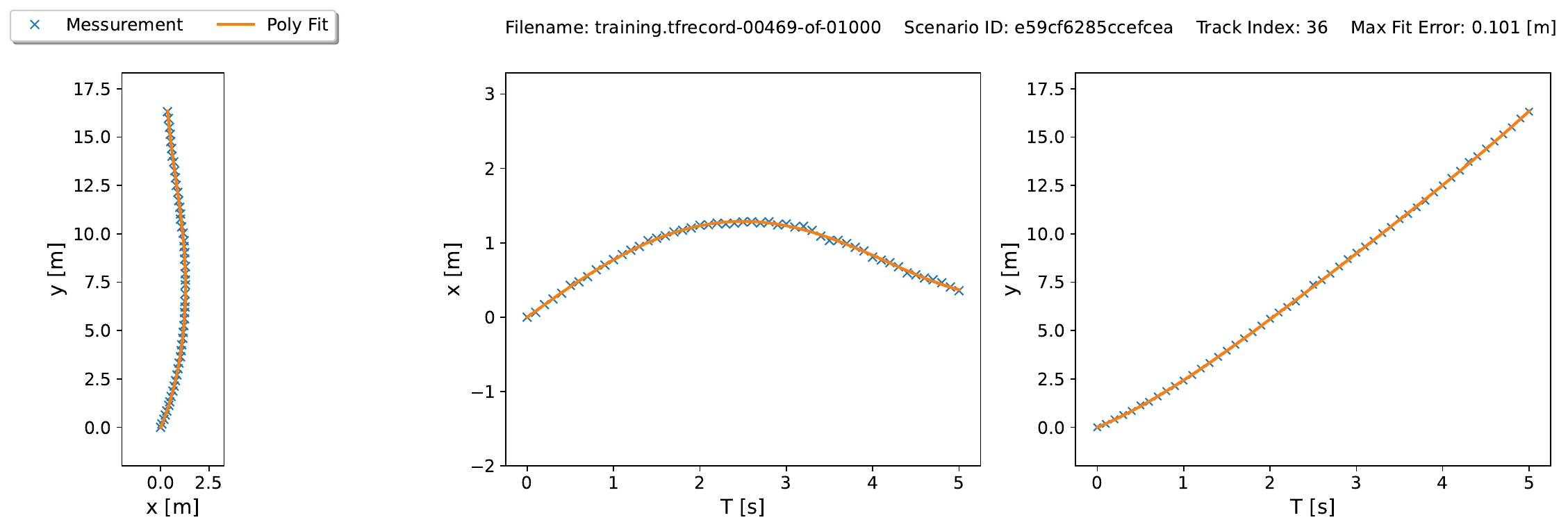} 

\includegraphics[width=5.5in, height=1.8in]{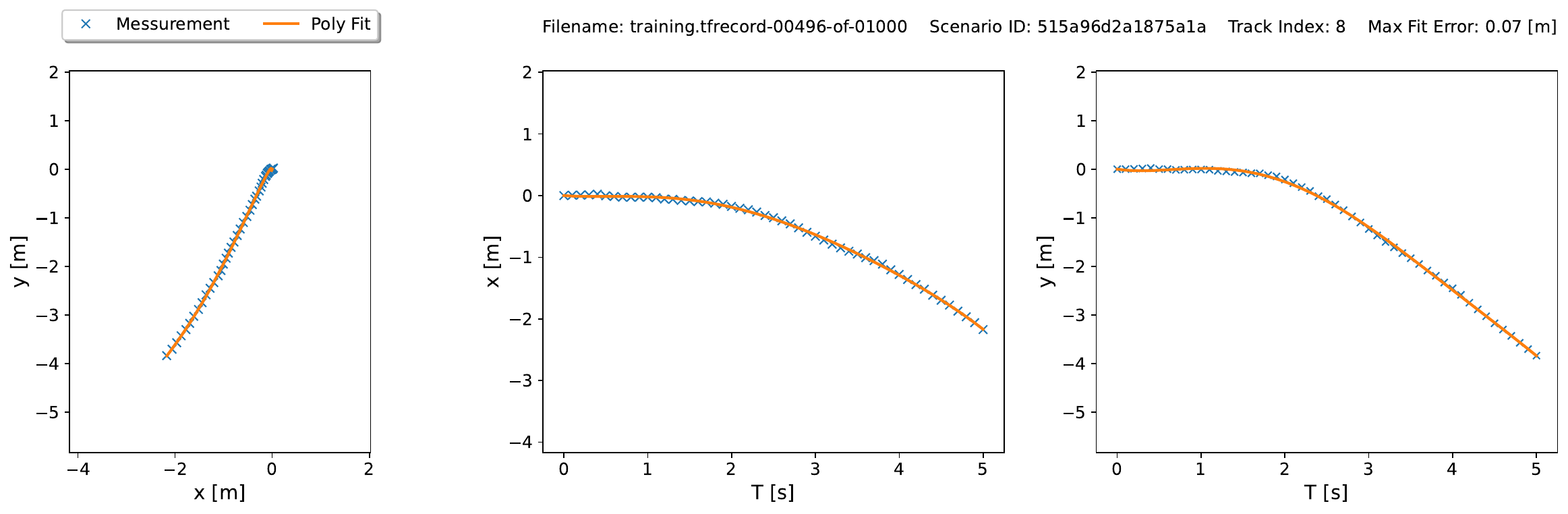} 

\includegraphics[width=5.5in, height=1.8in]{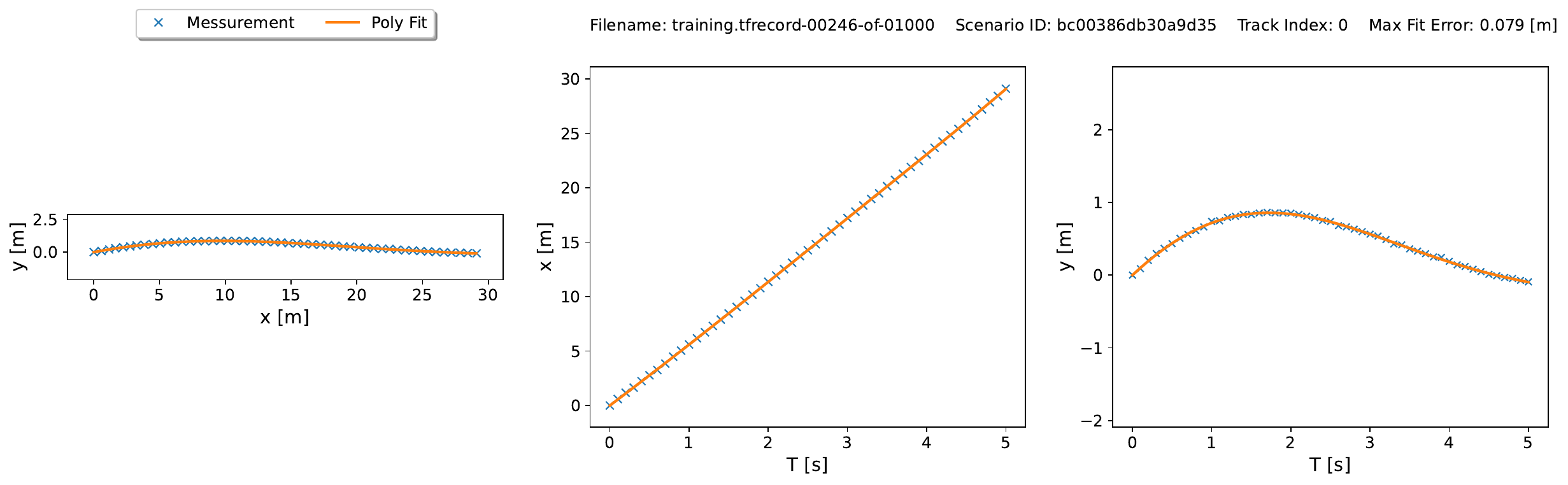} 

\includegraphics[width=5.5in, height=1.8in]{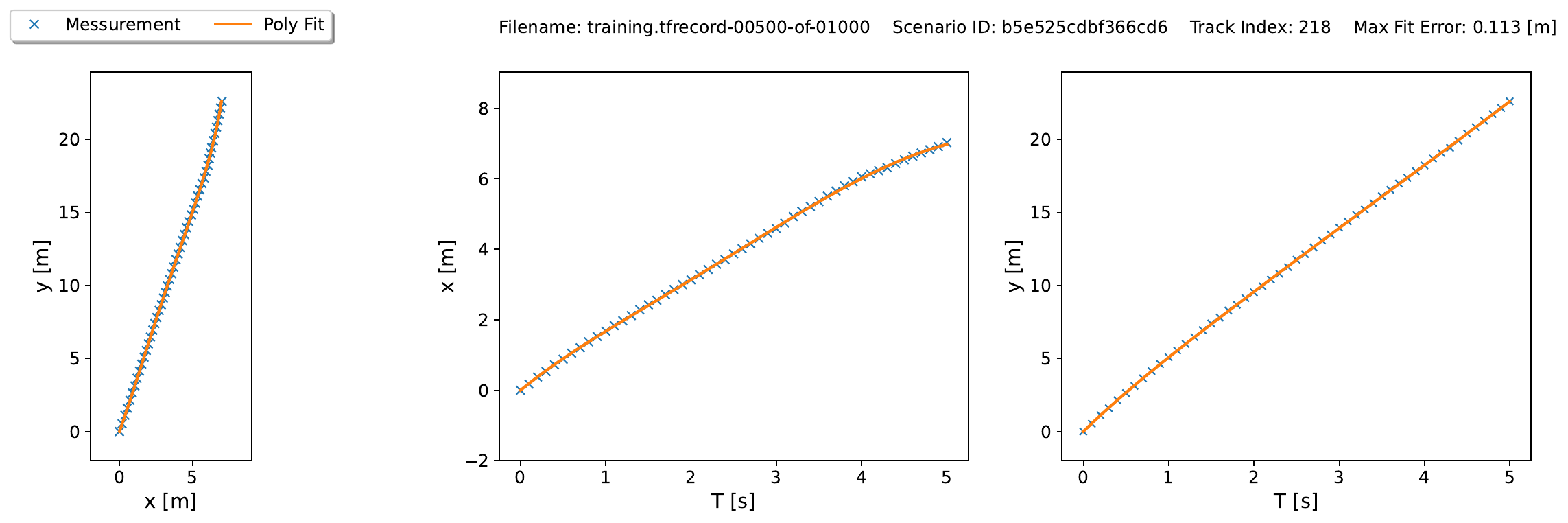} 

\includegraphics[width=5.5in, height=1.7in]{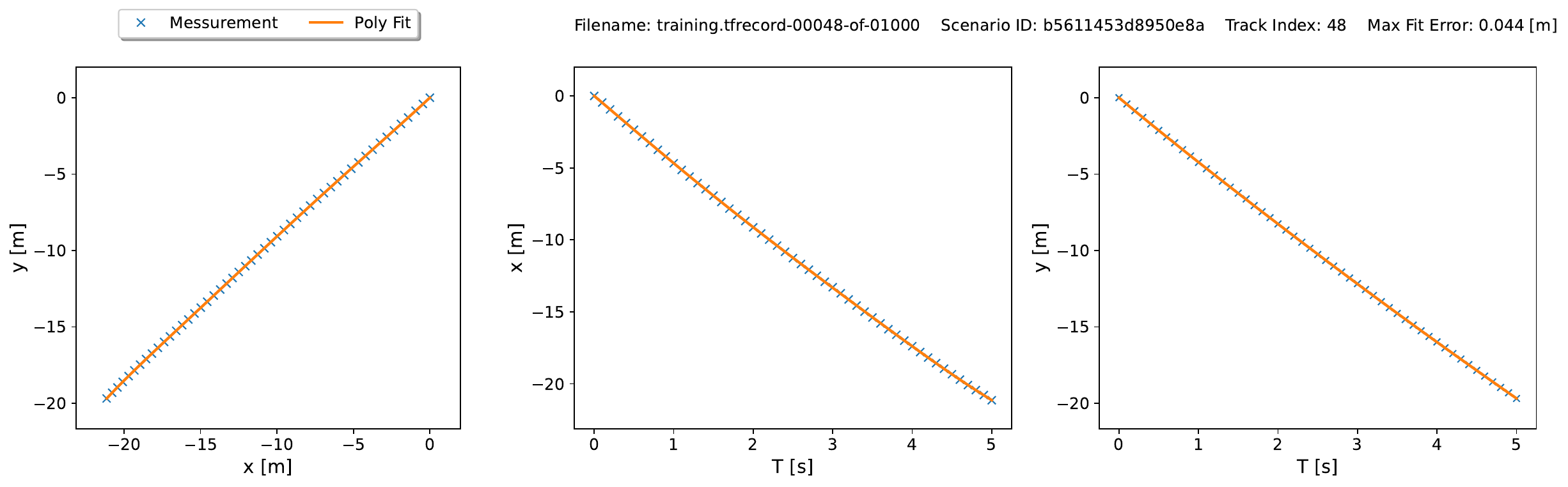} 

\section{10 5-seconds pedestrian trajectories with highest fit error in WO (Fitted with $\hat{n} = 5$)}
\centering
\includegraphics[width=5.5in, height=1.8in]{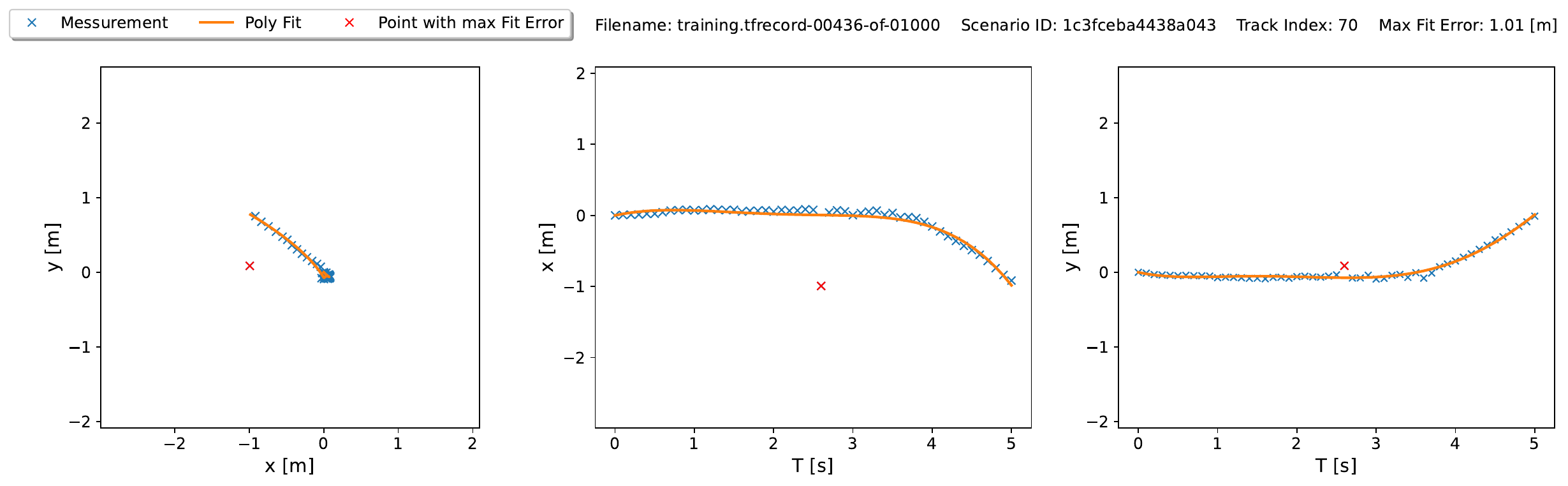} 

\includegraphics[width=5.5in, height=1.8in]{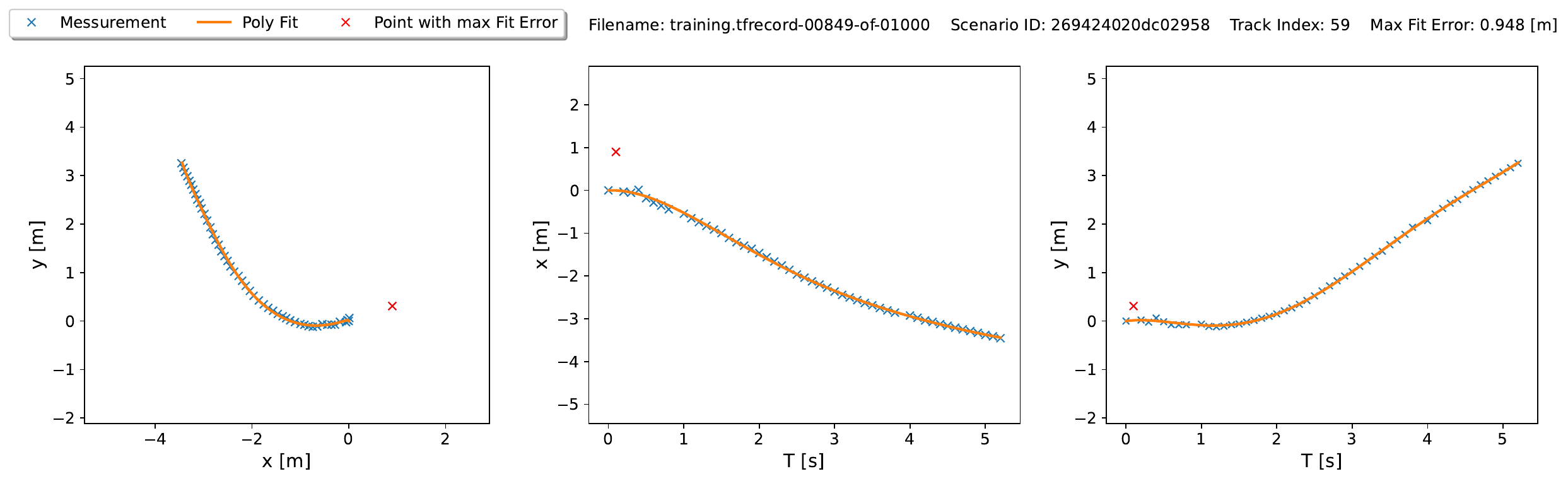} 

\includegraphics[width=5.5in, height=1.8in]{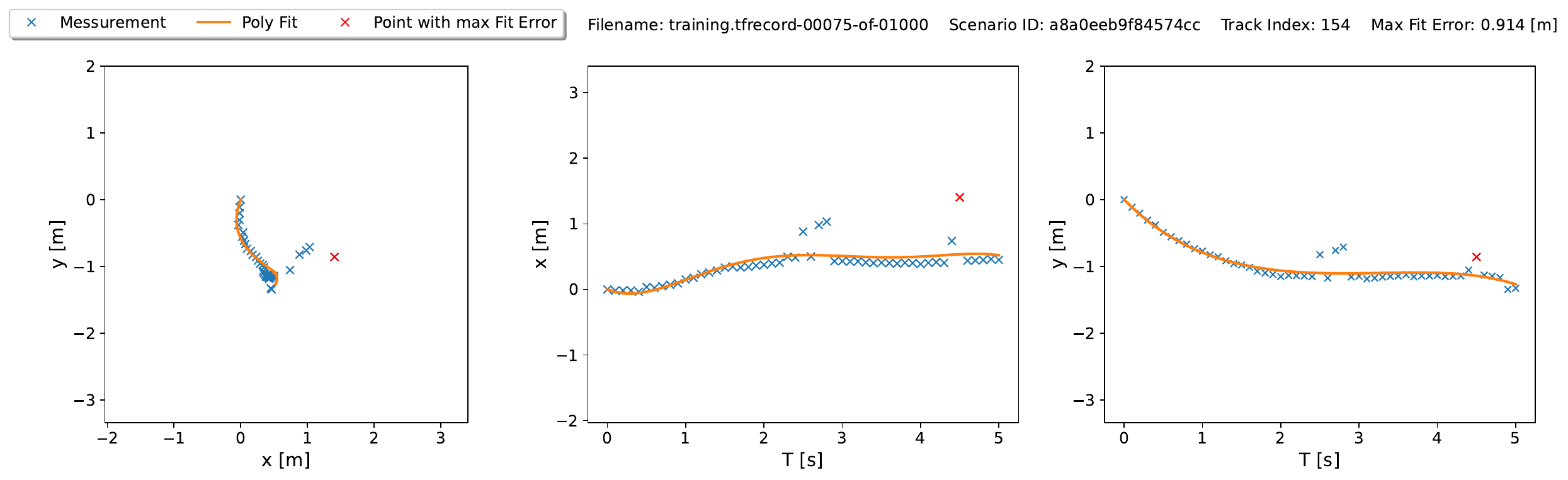} 

\includegraphics[width=5.5in, height=1.8in]{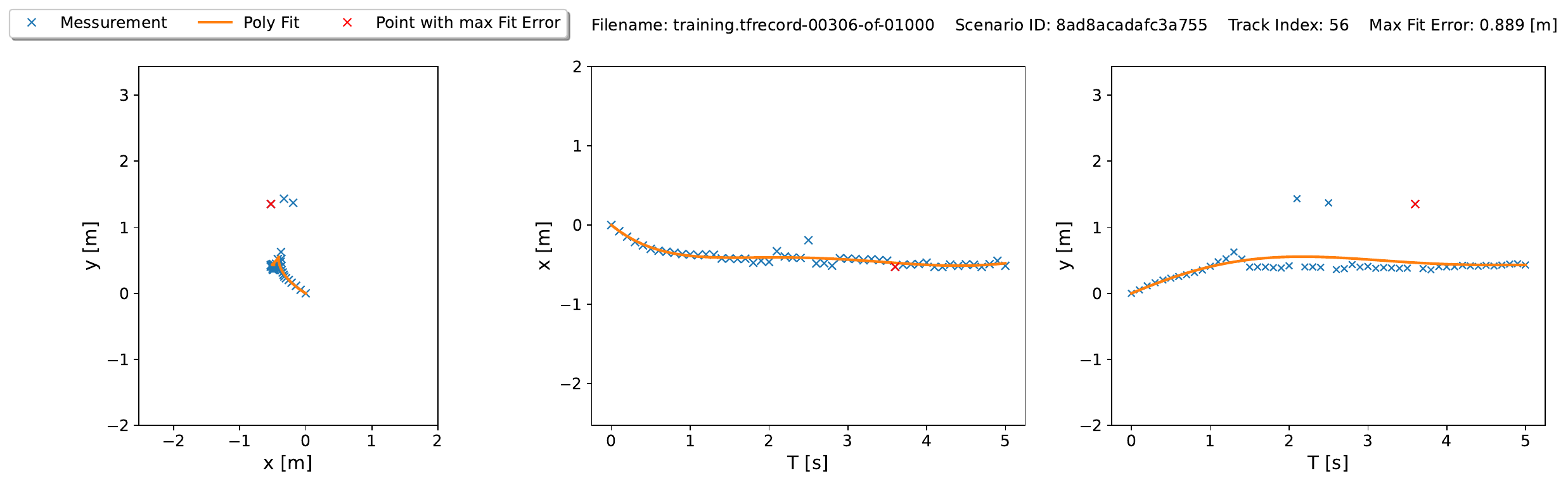} 

\includegraphics[width=5.5in, height=1.7in]{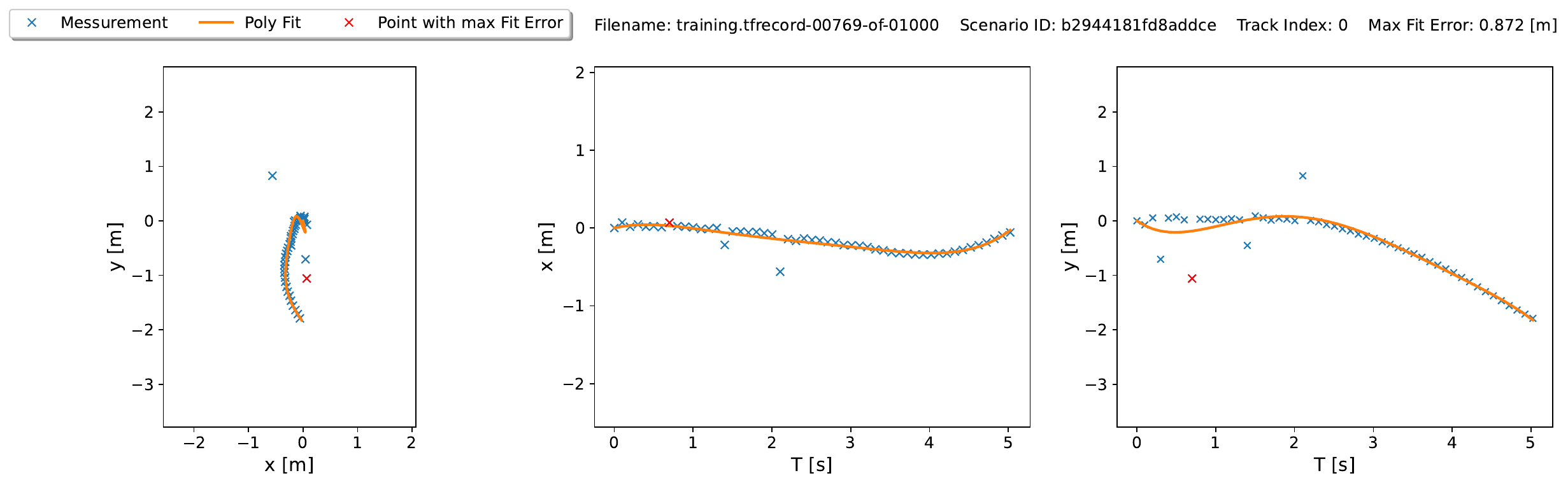} 

\includegraphics[width=5.5in, height=1.8in]{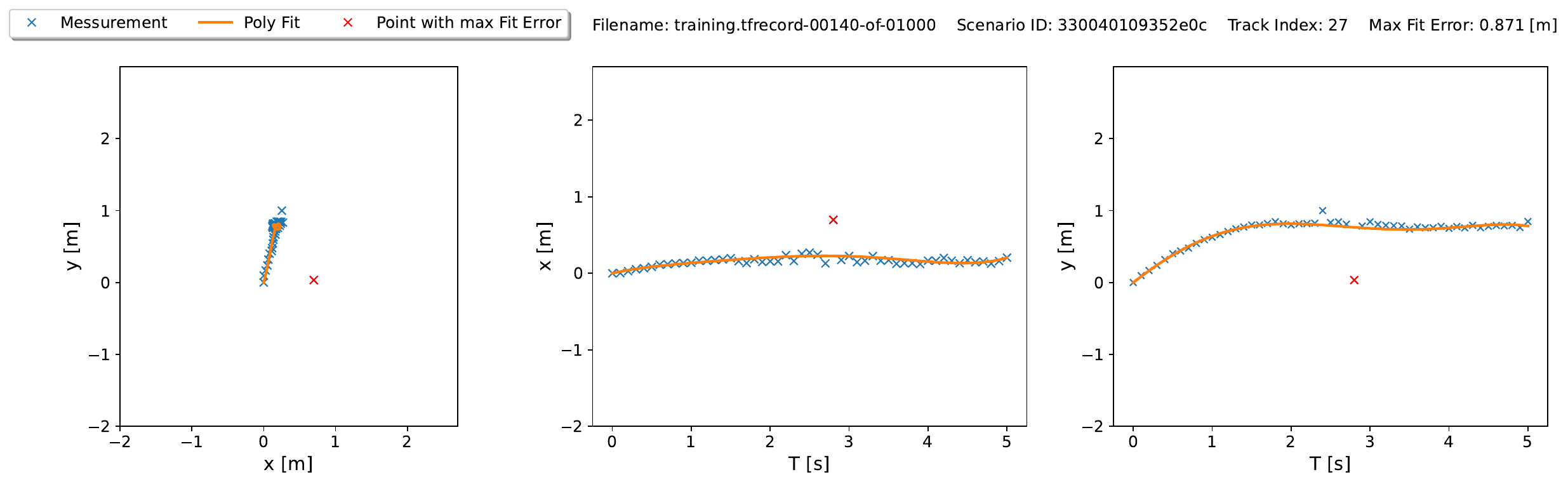} 

\includegraphics[width=5.5in, height=1.8in]{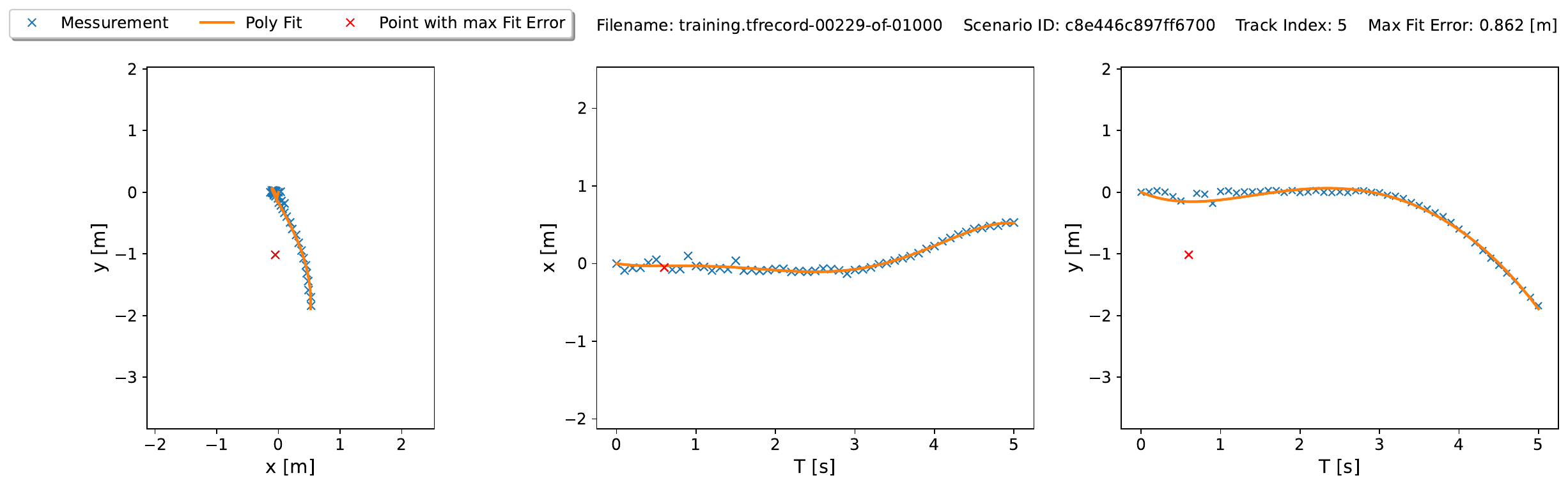} 

\includegraphics[width=5.5in, height=1.8in]{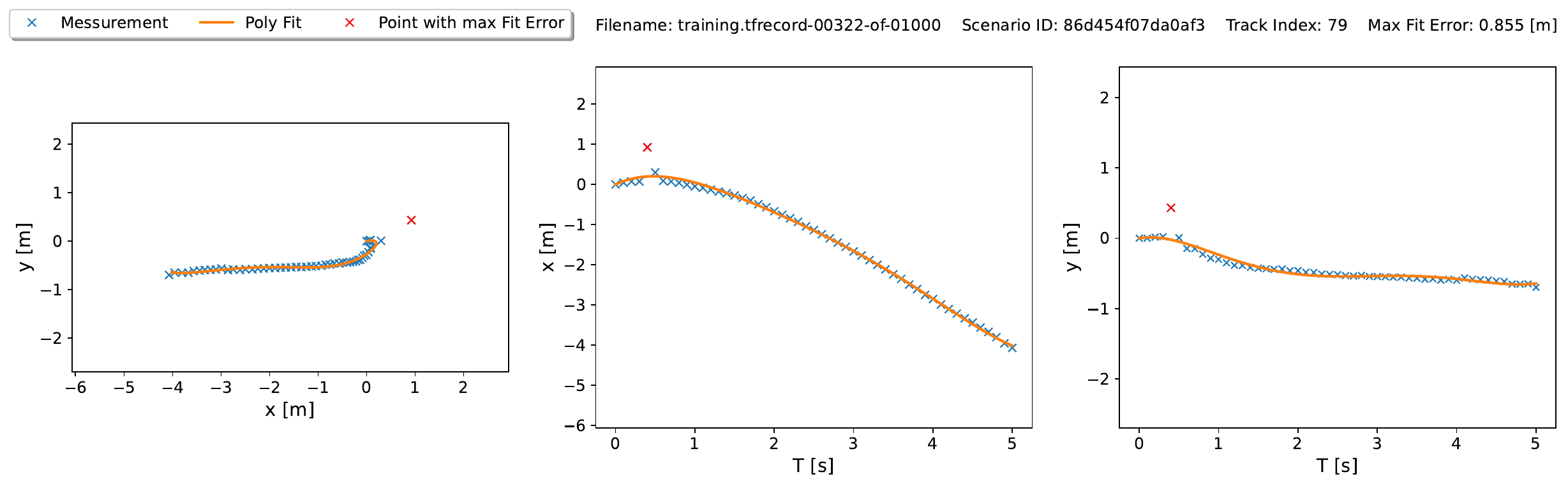} 

\includegraphics[width=5.5in, height=1.8in]{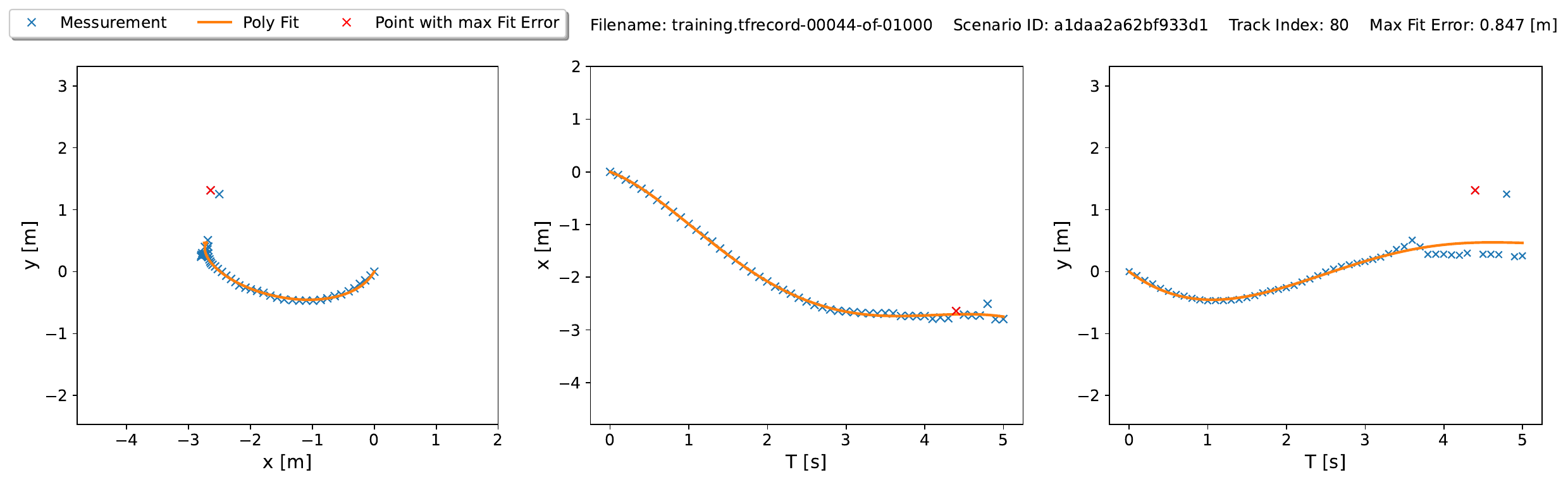} 

\includegraphics[width=5.5in, height=1.7in]{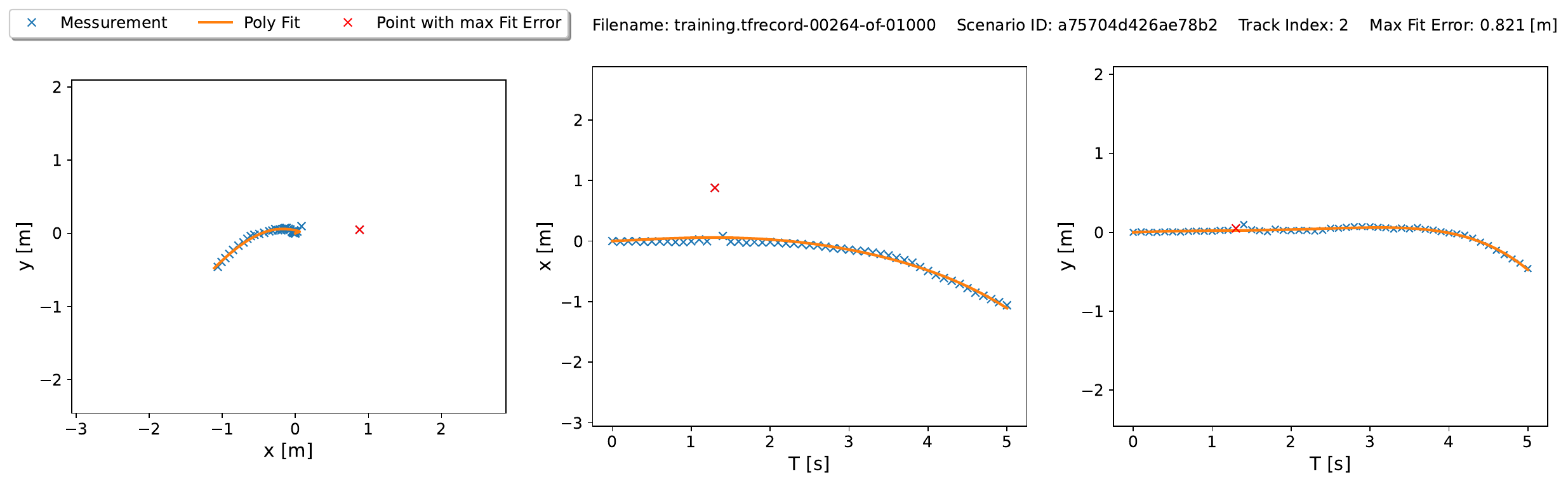} 

\section{10 random 5-seconds pedestrian trajectories in WO (Fitted with $\hat{n} = 5$)}
\centering
\includegraphics[width=5.5in, height=1.8in]{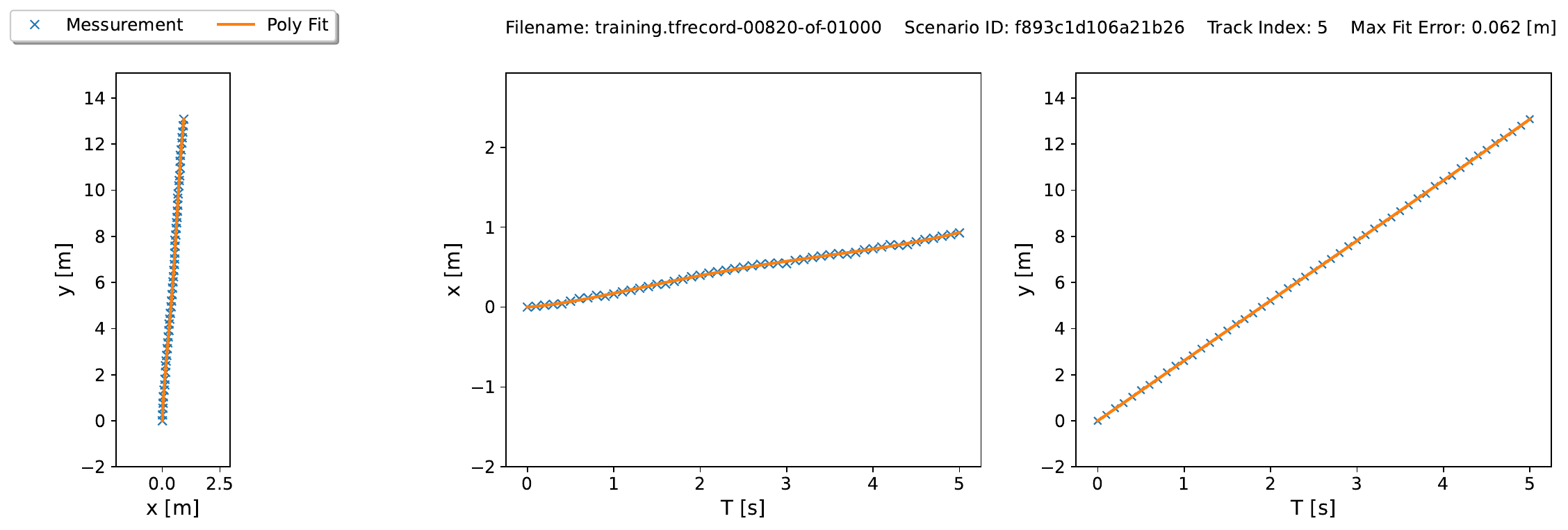} 

\includegraphics[width=5.5in, height=1.8in]{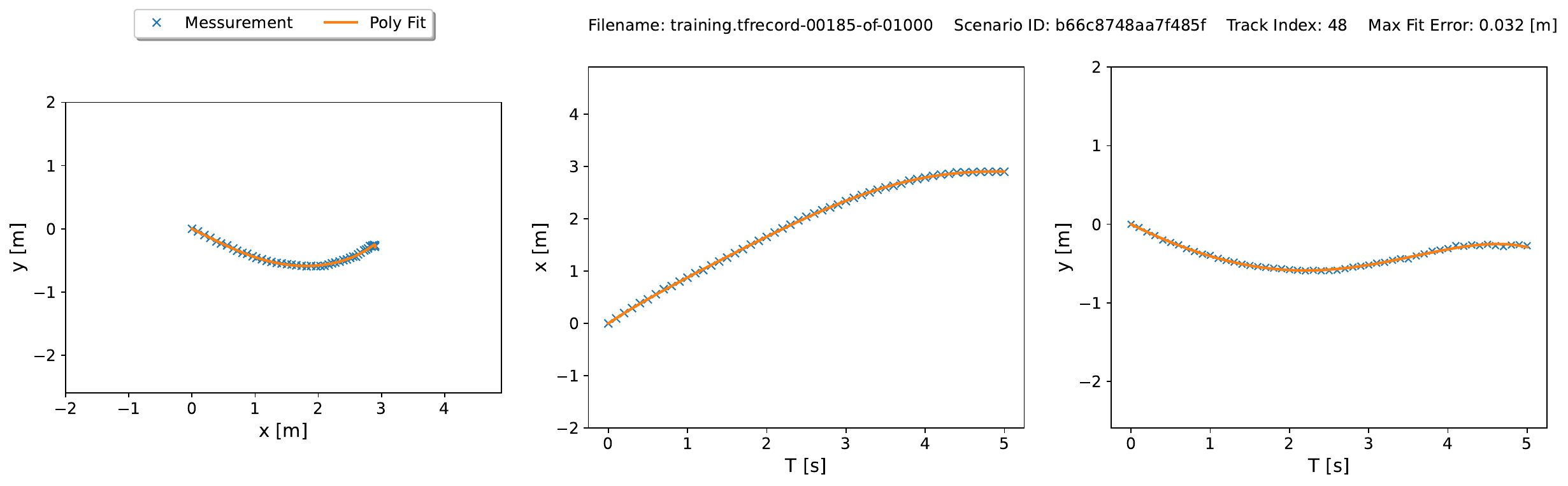} 

\includegraphics[width=5.5in, height=1.8in]{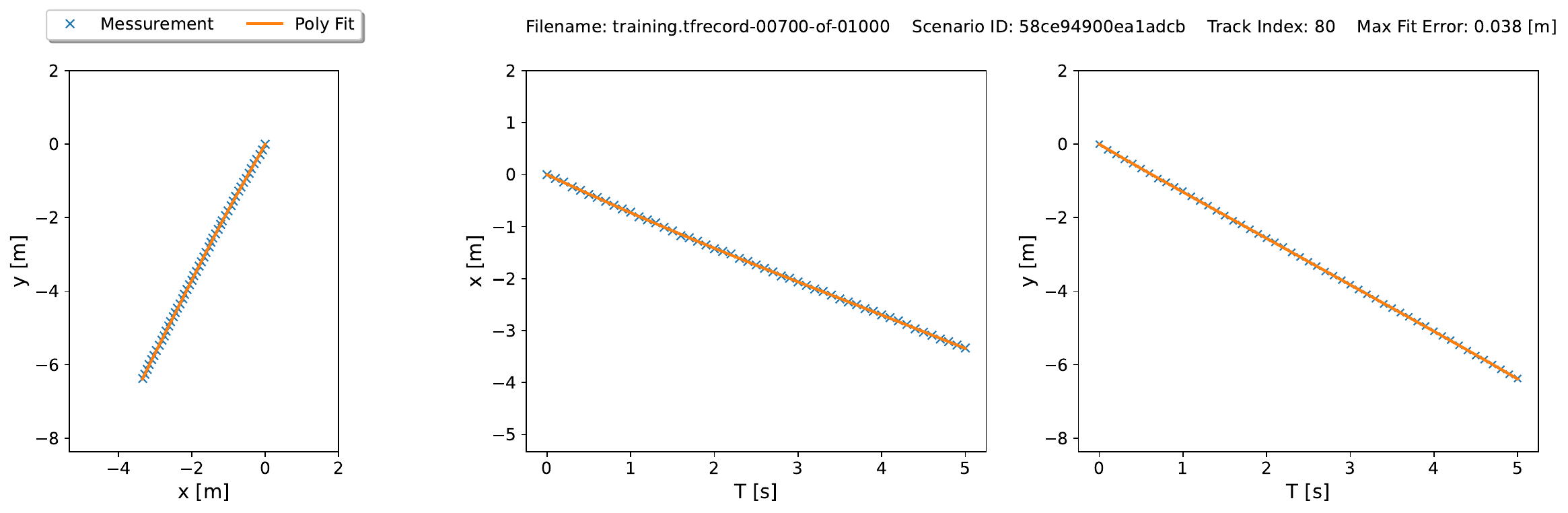} 

\includegraphics[width=5.5in, height=1.8in]{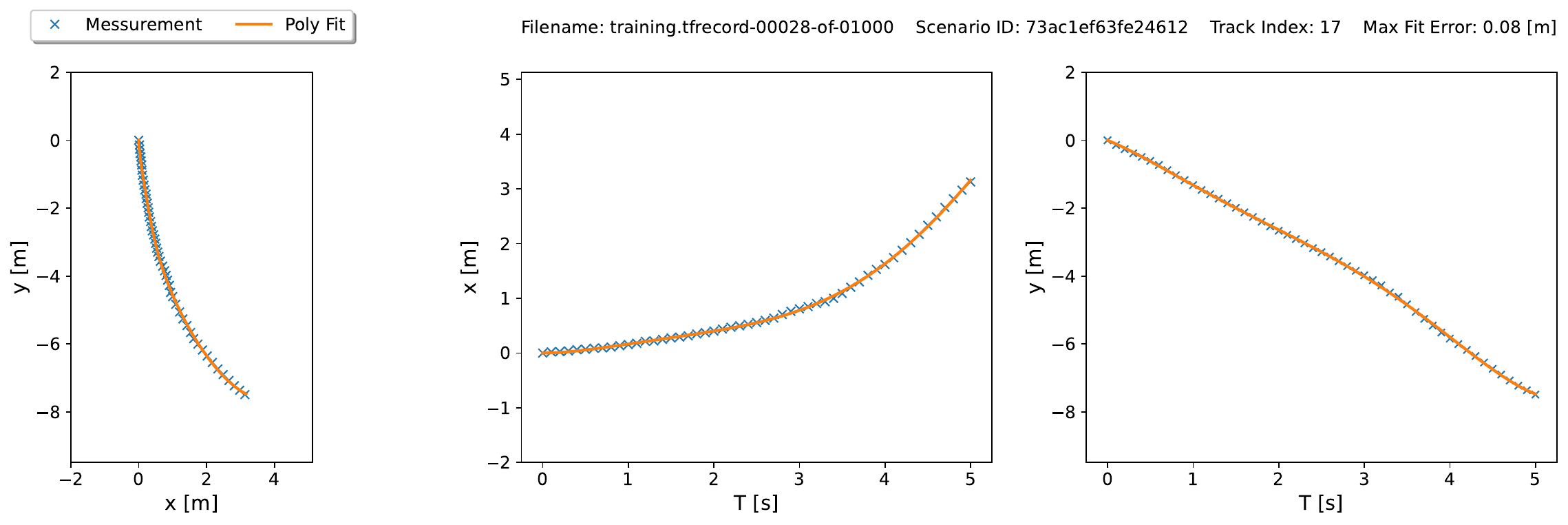} 

\includegraphics[width=5.5in, height=1.7in]{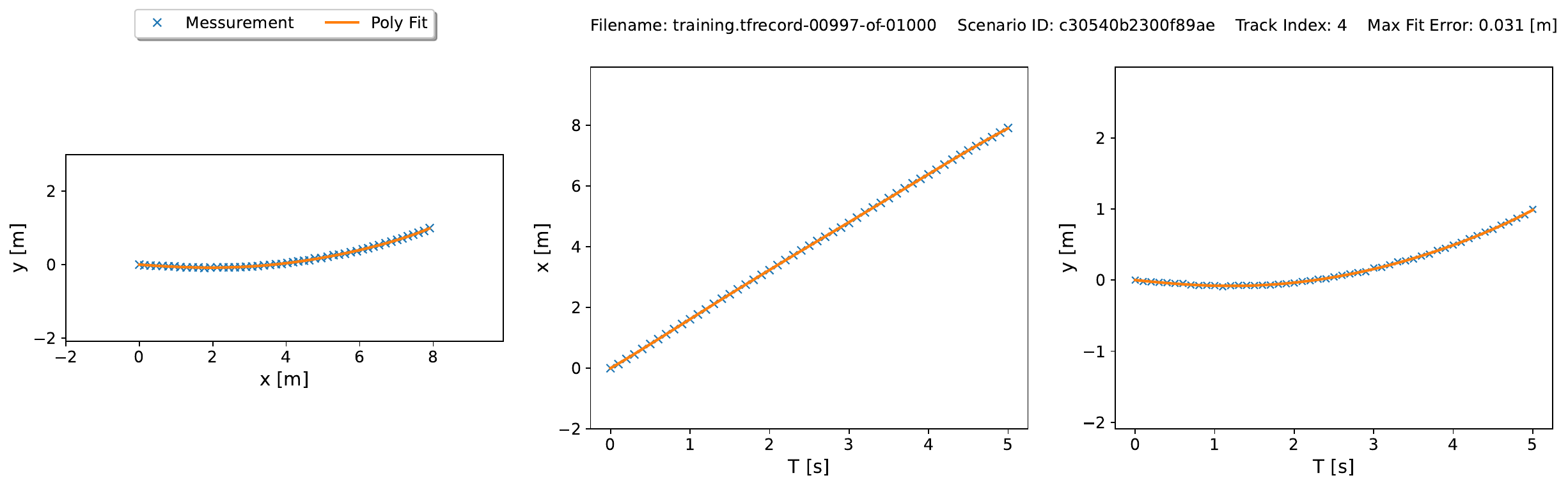} 

\includegraphics[width=5.5in, height=1.8in]{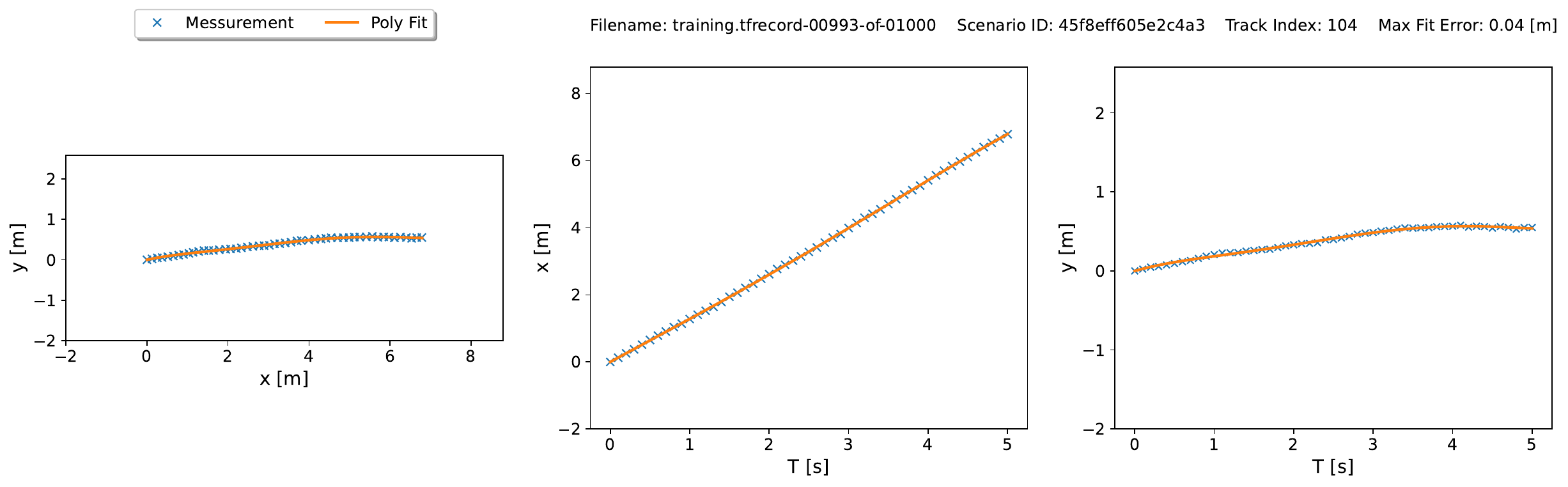} 

\includegraphics[width=5.5in, height=1.8in]{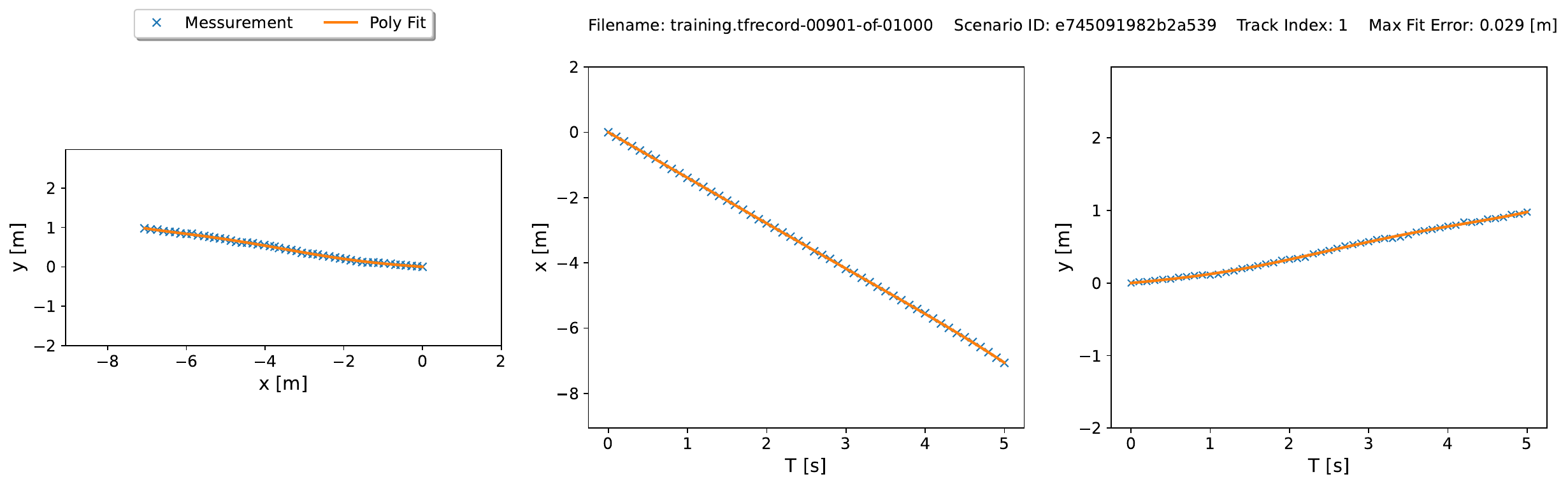} 

\includegraphics[width=5.5in, height=1.8in]{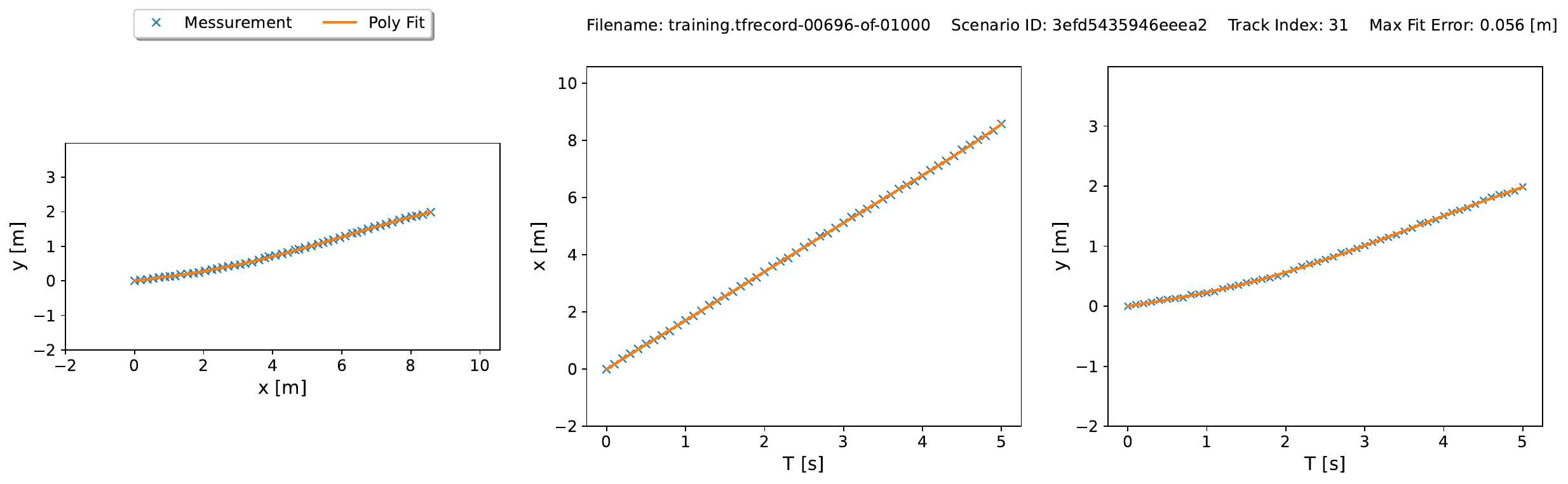} 

\includegraphics[width=5.5in, height=1.8in]{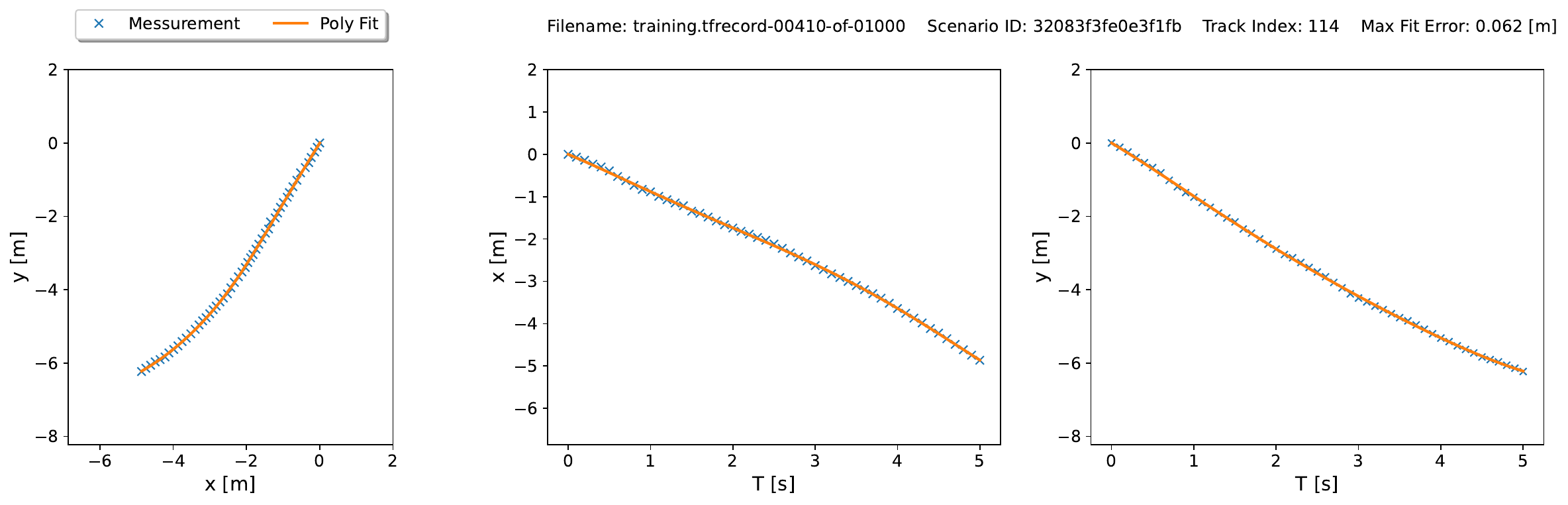} 

\includegraphics[width=5.5in, height=1.7in]{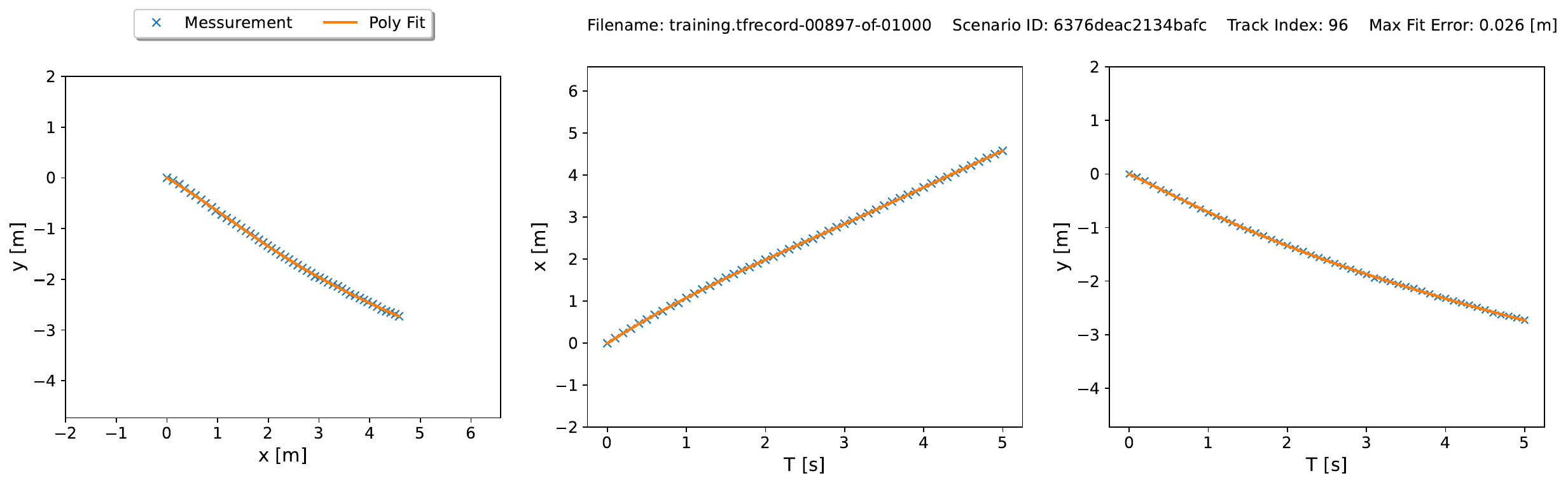} 

\end{appendices}

\end{document}